\documentclass[twoside]{article}

\usepackage[arxiv]{arxivstyle}
\usepackage[round]{natbib}
\usepackage{graphicx}
\renewcommand{\bibname}{References}
\renewcommand{\bibsection}{\subsubsection*{\bibname}}

\usepackage[hidelinks]{hyperref}
\usepackage{url}
\usepackage{amsmath}
\usepackage{amssymb}
\usepackage{xcolor}
\usepackage{booktabs}
\usepackage{subfig}

\renewcommand{\vec}[1]{\mathbf{#1}}
\newcommand{\vecgreek}[1]{\boldsymbol{#1}}
\newcommand{\mtx}[1]{\mathbf{#1}}

\DeclareMathOperator{\Tr}{Tr}
\newcommand{\mtxtrace}[1]{\Tr\left\lbrace{#1}\right\rbrace}
\newcommand{\expectation}[1]{\mathbb{E}\left[{{#1}}\right]}

\newcommand{\Ltwonormsquared}[1]{\left\Vert{{#1}}\right\Vert_2^2} 
\newcommand{\Ltwonorm}[1]{\left\Vert{{#1}}\right\Vert_2} 
\newcommand{\Frobnormsquared}[1]{\left\Vert{{#1}}\right\Vert_F^2} 
\newcommand{\Frobnorm}[1]{\left\Vert{{#1}}\right\Vert_F}

\begin{document}

%

%

\twocolumn[
    \aistatstitle{TL-PCA: Transfer Learning of Principal Component Analysis}

    \aistatsauthor{ Sharon Hendy \And  Yehuda Dar }
    
    \aistatsaddress{ 
    Computer Science Department\\Ben-Gurion University\\hendys@post.bgu.ac.il
    \And
    Computer Science Department\\Ben-Gurion University\\ydar@bgu.ac.il
    }
]

\begin{abstract}
Principal component analysis (PCA) can be significantly limited when there is too few examples of the target data of interest. We propose a transfer learning approach to PCA (TL-PCA) where knowledge from a related source task is used in addition to the scarce data of a target task. Our TL-PCA has two versions, one that uses a pretrained PCA solution of the source task, and another that uses the source data. Our proposed approach extends the PCA optimization objective with a penalty on the proximity of the target subspace and the source subspace as given by the pretrained source model or the source data. This optimization is solved by eigendecomposition for which the number of \textit{data-dependent} eigenvectors (i.e., principal directions of TL-PCA) is not limited to the number of target data examples, which is a root cause that limits the standard PCA performance. Accordingly, our results for image datasets show that the representation of test data is improved by TL-PCA for dimensionality reduction where the learned subspace dimension is lower or higher than the number of target data examples.

\end{abstract}

\section{Introduction}

Transfer learning is a common practice for mitigating the high data and computational demands of training deep neural networks (DNNs) \citep{pan2009survey}. In the prevalent transfer learning approach, a DNN is trained for a target task using a DNN that has been pretrained for a related source task. There are various ways to use the pretrained source DNN, for example, setting the first layers of the target DNN fixed on the corresponding layers from the pretrained source DNN, or initializing the target training process with the entire pretrained source DNN. 

More recently, there is an increasing research of transfer learning of linear models. One motivation is to provide analytical theory for transfer learning processes \citep{dar2020double,dar2024common}. Another motivation comes from applications where data is too scarce to achieve the best performance that linear models can obtain \citep{craig2024pretraining}. 
The various linear models examined in the recent transfer learning literature include the ordinary least squares \citep{obst2021transfer}, overparameterized least squares that interpolates the target dataset \citep{dar2020double}, linear regression with a ridge-like transfer learning penalty \citep{dar2024common}, and lasso regression \citep{craig2024pretraining}. Examples for other transfer learning studies of simple models that are not purely linear include the perceptron \citep{dhifallah2021phase} and two-layer neural networks \citep{gerace2021probing}.

In this paper, we study transfer learning for principal component analysis (PCA). Data analysis and dimensionality reduction using PCA are prevalent for many applications, including problems with limited availability of application-specific data. This motivates us to study the questions of \textit{how to design transfer learning procedures for PCA, and when they are expected to be beneficial.}

We propose the transfer learning for PCA (TL-PCA) approach where the usual optimization objective of standard PCA is augmented with a penalty to make the subspace learned (for the target task) sufficiently close to the subspace of a source task. We formulate two versions of TL-PCA: 
\begin{itemize}
    \item \textbf{TL-PCA-P:} Using a pretrained PCA solution of the source task. The penalty term corresponds to the principal angles between the learned target subspace and the (complete or part of the) pretrained source subspace. Using the pretrained source PCA is particularly useful when the source data is unavailable.
    \item \textbf{TL-PCA-D:} Using the source data directly to define the penalty term as the reconstruction error of the source data from projection onto the learned target subspace. This approach can be more beneficial than using the pretrained source model when source data is available. 
\end{itemize}
We show that TL-PCA is solved by eigendecomposition of a matrix formed by a weighted average of the target sample covariance with a matrix corresponding to the source subspace projection (in TL-PCA-P) or the source sample covariance (in TL-PCA-D). 

Our TL-PCA allows us to obtain more data-dependent eigenvectors (i.e., principal directions of TL-PCA) compared to the standard PCA on the target task. This is because the  principal directions of  standard PCA are limited to the number of target data examples, beyond which the additional principal directions are arbitrarily selected under orthogonality constraints to the already-selected principal directions. Such arbitrary selection of principal directions induces poor performance in standard PCA, a problem that our TL-PCA approach excellently solves. 

Our experiments on image datasets examine the generalization of the learned dimensionality reduction operators to test data of the target task. Our results show that the proposed TL-PCA approach performs better than standard PCA for the target data, at a wide range of learned subspace dimension (including both smaller and larger than the number of target examples). An additional comparison shows that, unless the learned subspace dimension is excessively larger than the number of target examples, TL-PCA is better than the direct usage of the pretrained source model on target test data. 
The TL-PCA-D version, which uses the source data, is shown to achieve better performance than the TL-PCA-P version, which uses the pretrained source model. 
Either way, both of the TL-PCA versions usually outperform the standard PCA options. Hence, this paper shows that transfer learning is beneficial for PCA in data-limited applications. 

This paper is organized as follows. In Section \ref{sec:Related Work}, we discuss related literature. In Section \ref{sec:The Target Task}, we define the target task and its standard PCA solution. In Section \ref{sec:Transfer Learning of PCA}, we define the source task (Section \ref{subsec:Transfer Learning of PCA - The Source Task}), the proposed TL-PCA-P method (Section \ref{subsec:TL-PCA-P: Transfer Learning using the Source Pretrained Model}), and the proposed TL-PCA-D method (Section \ref{subsec:TL-PCA-D: Transfer Learning using the Source Data}). In Section  \ref{sec:Experimental Results}, we describe the experiments and compare the reconstruction errors of the examined PCA and TL-PCA methods, showing the performance gains of TL-PCA. 
In Section \ref{sec:The TL-PCA Benefits as Reflected by Principal Angles}, we analyze principal angles to show that TL-PCA learns a subspace that is closer to a subspace learned by much more target data.
In the appendices, we provide mathematical proofs, additional experimental details and results. 

\section{Related Work}
\label{sec:Related Work}

As mentioned above, transfer learning for various linear models, especially linear regression, was studied by \citet{obst2021transfer,dar2020double,dar2024common,craig2024pretraining}. One prevalent approach, which follows the deep learning practice, is to use the pretrained source model to implement transfer learning for a target task \citep{obst2021transfer,dar2020double,dar2024common,craig2024pretraining}. Another approach is to use the source data to implement the transfer learning \citep{song2024generalization}; some may prefer to call this approach co-training due to the usage of source data rather than a pretrained source model.  
Importantly, the existing literature does not consider transfer learning for PCA -- a learning task that significantly differs from the well-studied regression and classification models. 
In this paper, we address this significant gap in the literature by studying transfer learning for PCA in two versions that use the source pretrained model and source data. 

From a technical perspective, we augment the optimization objective of standard PCA with an additional penalty on the proximity of the learned target subspace to the source subspace (using the source pretrained model or source data). This approach to transfer learning for linear models by adding a penalty in the optimization objective is conceptually similar to the proposed approach by \citet{dar2024common} for linear regression that includes a penalty on the distance of the learned parameters from the pretrained source parameters. In this paper, we study transfer learning for PCA, which is conceptually and technically different than the linear regression of \citet{dar2024common}. 

In our TL-PCA-P version, which uses the pretrained PCA model of the source task, the optimization objective includes a Frobenius-norm penalty on the distance between the orthogonal projection operators of the to-be-learned and the pretrained subspaces; this penalty can be equivalently formulated as constraining the principal angles \citep{bjorck1973numerical} between the learned target subspace and the source subspace. A technically-similar penalty term appeared in the method by \citet{panagopoulos2016constrained} for learning a constrained subspace classifier for binary classification tasks. Their method jointly learns two subspaces (one for each class) whose principal angles are constrained to be smaller or larger than in separate PCA for each of the classes. Their method alternates between intermediate optimizations that learn each subspace given the other one fixed. Their \textit{intermediate} optimization has some technical relation to our TL-PCA-P version that uses the source pretrained model (but not to our TL-PCA-D version that uses the source data). However, their goal and overall discussion are completely different as they jointly learn two subspaces for classifying data according to the closest subspace, in contrast to our transfer learning from source to target tasks where only the target subspace is learned for dimensionality reduction and representation of the target data.

\section{The Target Task: Definition and the Standard PCA Solution}
\label{sec:The Target Task}

We are given a dataset $\mathcal{D} = \{ \vec{x}_i \}_{i=1}^n \subset \mathbb{R}^d$, consisting of $n$ examples that are independent and identically distributed (i.i.d.)~drawn from an unknown probability distribution $P_{\vec{x}}$. The vectors in $\mathcal{D}$ are centered with respect to their sample mean $\vecgreek{\mu}_{\mathcal{D}}=\frac{1}{n}\sum_{i=1}^{n}\vec{x}_i$, i.e., the $i^{\rm th}$ centered example is $\bar{\vec{x}}_i\triangleq\vec{x}_i - \vecgreek{\mu}_{\mathcal{D}}$.
The centered examples of $\mathcal{D}$ are organized as the columns of the $d \times n$ data matrix $\mtx{X}\triangleq [\bar{\vec{x}}_1,\dots,\bar{\vec{x}}_n]$. The sample covariance matrix of $\mathcal{D}$ is $\widehat{\mtx{C}}_\mtx{X}\triangleq \frac{1}{n}\mtx{X}\mtx{X}^T$. The sample covariance empirically approximates the (unknown) true covariance matrix $\mtx{C}_{\vec{x}}\triangleq \expectation{(\vec{x}-\expectation{\vec{x}})(\vec{x}-\expectation{\vec{x}})^T}$ where the expectation is with respect to the distribution $P_{\vec{x}}$.

\subsection{Standard PCA: Formulation and Solution}
\label{subsec:Standard PCA: Formulations and Solution}
The goal is to learn a linear operator $\mtx{U}_k \in\mathbb{R}^{d\times k}$ for dimensionality reduction from the $d$-dimensional space of the given data in $\mathcal{D}$ to a $k$-dimensional subspace, where $k<d$ and the columns of $\mtx{U}_k$ are orthonormal. The dimensionality reduction aims to represent the data in $\mathcal{D}$ in a way that maximizes the preserved information (hence, minimizes the error) despite the reduced dimension. While only $n$ examples are given in $\mathcal{D}$, the learned dimensionality reduction operator should \textit{generalize} well to new test data from the unknown distribution that the given examples were drawn from, namely, $\mtx{U}_k$ should operate well on $\vec{x}\sim P_{\vec{x}}$ even though $\vec{x}\notin\mathcal{D}$.

A data vector $\vec{x} \in\mathbb{R}^d$ has a $k$-dimensional representation vector $\vec{z}\triangleq \mtx{U}_k^T \vec{x}$ and a $d$-dimensional reconstructed vector $\widehat{\vec{x}}\triangleq \mtx{U}_k \vec{z} = \mtx{U}_k\mtx{U}_k^T \vec{x}$. The squared error between the reconstruction and the original vectors is $\mathcal{E}(\vec{x}, \widehat{\vec{x}} ; \mtx{U}_k)\triangleq \Ltwonormsquared{\vec{x} - \widehat{\vec{x}}}$. The subscript $i$ in  the representation vector $\vec{z}_i$ and reconstruction vector $\widehat{\vec{x}}_i$ denotes their correspondence to the $i^{\rm th}$ example in $\mathcal{D}$.

For a given $k$, standard PCA learns the operator $\mtx{U}_k \in\mathbb{R}^{d\times k}$ (with orthonormal columns) that minimizes the mean squared error of the reconstruction of the examples in $\mathcal{D}$: 
\begin{equation}
\label{eq: Target PCA - error minimization (matrix) formulation}
    {\mtx{U}}_k = \underset{\mtx{W} \in \mathbb{R}^{d \times k}, \mtx{W}^T \mtx{W} = \mtx{I}_k}{\arg\min} \frac{1}{n} \Frobnormsquared{ (\mtx{I}_d - \mtx{W}\mtx{W}^T) \mtx{X} }
\end{equation}
where the optimization objective is the matrix formulation of $\frac{1}{n}\sum_{i=1}^{n}\mathcal{E}(\vec{x}_i, \widehat{\vec{x}}_i ; \mtx{W})=\frac{1}{n}\sum_{i=1}^{n}\Ltwonormsquared{\vec{x}_i - \mtx{W}\mtx{W}^T \vec{x}_i}$. 
It is well-known that the error minimization formulation in (\ref{eq: Target PCA - error minimization (matrix) formulation}) is equivalent to the variance maximization formulation:
\begin{equation}
\label{eq: Target PCA - variance maximization (matrix) formulation}
    \mtx{U}_k = \underset{\mtx{W} \in \mathbb{R}^{d \times k}, \mtx{W}^T \mtx{W} = \mtx{I}_k}{\arg\max} \, \mtxtrace{\mtx{W}^T \mtx{X} \mtx{X}^T \mtx{W}}.
\end{equation}
This optimization is solved by $\mtx{U}_k = [ \vec{u}_1, \dots, \vec{u}_k ]$ where $\vec{u}_i \in \mathbb{R}^d$ is the eigenvector corresponding to the $i^{\rm th}$ largest eigenvalue $\lambda_i$ of the sample covariance matrix $\widehat{\mtx{C}}_\mtx{X} = \frac{1}{n} \mtx{X} \mtx{X}^T$ of the \textit{target} dataset $\mathcal{D}$. 
In PCA, $\vec{u}_i$ and $\lambda_i$ are called the $i^{\rm th}$ principal direction and principal component, respectively.

\subsection{Standard PCA is Limited by Scarce Data}
Standard PCA might perform poorly in case of insufficient data examples. There are two main reasons for such degraded performance: 
\begin{itemize}
    \item Too few examples in $\mathcal{D}$ induce a sample covariance matrix $\widehat{\mtx{C}}_\mtx{X}$ that may significantly deviate from the true covariance matrix $\mtx{C}_\vec{x}$ of $P_{\vec{x}}$. Thus, the principal directions and components obtained from  $\mathcal{D}$ might have significant inaccuracies for test data, i.e., generalize poorly.
    \item In a high-dimensional or overparameterized setting, the data dimension \( d \) exceeds the number of samples \( n \). Hence, the sample covariance matrix $\widehat{\mtx{C}}_\mtx{X}$ has a rank of at most \( n - 1 \) (where the $-1$ is due to the data centering). As a result, for a subspace dimension \( k \ge n \), the learned subspace is based on a rank-deficient $\widehat{\mtx{C}}_\mtx{X}$. \citet{dar2020subspace} referred to such subspace learning problem as \textit{rank-overparameterized}. The learned subspace in this case is inherently affected by the rank deficiency of the sample covariance matrix, which leaves at least $k-n+1$ degrees of freedom in choosing the basis vectors for the learned subspace. 
\end{itemize}
These situations call for new ideas to improve PCA in modern settings where data is high dimensional and/or have insufficient examples for the task of interest.

\section{Transfer Learning of PCA}
\label{sec:Transfer Learning of PCA}

To address the issues of standard PCA for high dimensional and/or scarce data, we propose transfer learning of PCA. We call our approach TL-PCA, and propose for it two versions that leverage knowledge from a related source task with sufficient data examples. Our TL-PCA aims to mitigate the inaccuracies and rank deficiency of the sample covariance matrix of the target dataset $\mathcal{D}$ by utilizing information from the source subspace.


\subsection{The Source Task}
\label{subsec:Transfer Learning of PCA - The Source Task}
The source task is a standard PCA that learns a $\widetilde{k}$-dimensional subspace in $\mathbb{R}^d$ for the dataset $\widetilde{\mathcal{D}} = \{ \widetilde{\vec{x}}_i \}_{i=1}^{\widetilde{n}}\subset \mathbb{R}^d$ that includes $\widetilde{n}$ examples i.i.d.~drawn from an unknown probability distribution $P_{\widetilde{\vec{x}}}$. 
The vectors in $\widetilde{\mathcal{D}}$ are centered with respect to their sample mean $\vecgreek{\mu}_{\widetilde{\mathcal{D}}}=\frac{1}{\widetilde{n}}\sum_{i=1}^{\widetilde{n}}\widetilde{\vec{x}}_i$.
The centered examples of $\widetilde{\mathcal{D}}$ are organized as the columns of the $d \times \widetilde{n}$ data matrix $\widetilde{\mtx{X}}$. The sample covariance matrix of $\widetilde{\mathcal{D}}$ is $\widehat{\mtx{C}}_{\widetilde{\mtx{X}}}\triangleq \frac{1}{\widetilde{n}}\widetilde{\mtx{X}}\widetilde{\mtx{X}}^T$. 

Following the detailed explanation on the standard PCA given in Section \ref{subsec:Standard PCA: Formulations and Solution} for the target task, we write here more briefly the standard PCA formulations for the source task: 
\begin{align}
\label{eq: Source PCA - error minimization (matrix) formulation}
    \widetilde{\mtx{U}}_{\widetilde{k}} &= \underset{\mtx{W} \in \mathbb{R}^{d \times \widetilde{k}}, \mtx{W}^T \mtx{W} = \mtx{I}_{\widetilde{k}}}{\arg\min} \frac{1}{\widetilde{n}} \Frobnormsquared{ (\mtx{I}_d - \mtx{W}\mtx{W}^T) \widetilde{\mtx{X}} }
    \\
    \label{eq: Source PCA - variance maximization (matrix) formulation}
&= \underset{\mtx{W} \in \mathbb{R}^{d \times \widetilde{k}}, \mtx{W}^T \mtx{W} = \mtx{I}_{\widetilde{k}}}{\arg\max} \, \mtxtrace{\mtx{W}^T \widetilde{\mtx{X}} \widetilde{\mtx{X}}^T \mtx{W}}.    
\end{align}
This optimization is solved by $\widetilde{\mtx{U}}_{\widetilde{k}} = [ \widetilde{\vec{u}}_1, \dots, \widetilde{\vec{u}}_{\widetilde{k}} ]$ where $\widetilde{\vec{u}}_i \in \mathbb{R}^d$ is the eigenvector corresponding to the $i^{\rm th}$ largest eigenvalue $\widetilde{\lambda}_i$ of the sample covariance matrix $\widehat{\mtx{C}}_{\widetilde{\mtx{X}}} = \frac{1}{\widetilde{n}} \widetilde{\mtx{X}} \widetilde{\mtx{X}}^T$ of the \textit{source} dataset $\widetilde{\mathcal{D}}$. 

\subsection{TL-PCA-P: Transfer Learning using the Source Pretrained Model}
\label{subsec:TL-PCA-P: Transfer Learning using the Source Pretrained Model}
In this approach, we augment the standard PCA formulation for the target task (\ref{eq: Target PCA - error minimization (matrix) formulation}) with a constraint on the distance between the learned target subspace and the pretrained source subspace from (\ref{eq: Source PCA - error minimization (matrix) formulation}).  The idea is to mitigate the inaccuracies and possible rank-deficiency of the target sample covariance for solving (\ref{eq: Target PCA - error minimization (matrix) formulation}) by the additional information from the source subspace that was pretrained using more data examples.

To enforce proximity between the target and source subspaces, we define the penalty function 
\begin{equation}
    \label{eq:TL-PCA-P - penalty function - projection F norm}
    h(\mtx{W};\widetilde{\mtx{U}}_m) \triangleq \Frobnormsquared{ \mtx{W} \mtx{W}^T - \widetilde{\mtx{U}}_m \widetilde{\mtx{U}}_m^T }
\end{equation}
where $\mtx{W}\in\mathbb{R}^{d\times k}$ is the candidate solution (with orthonormal columns) to the TL-PCA-P and $\mtx{W} \mtx{W}^T$ is its orthogonal projection matrix to the subspace spanned by $\mtx{W}$ columns. Here, $m\in\{1,\dots,\widetilde{k}\}$ is the number of principal directions that are transferred from the pretrained source model $\widetilde{\mtx{U}}_{\widetilde{k}}$; the matrix $\widetilde{\mtx{U}}_m\triangleq [ \widetilde{\vec{u}}_1, \dots, \widetilde{\vec{u}}_{m} ]$ is the dimensionality reduction operator to the corresponding $m$-dimensional subspace of the source task, and $\widetilde{\mtx{U}}_m \widetilde{\mtx{U}}_m^T$ is the corresponding orthogonal projection matrix. 

Transferring only $m<k$ source principal directions implies that the $k$-dimensional learned target subspace has $k-m$ degrees of freedom that are not directly affected by the transfer penalty (\ref{eq:TL-PCA-P - penalty function - projection F norm}). Specifically, the transferred principal directions from the source are the most significant ones, which correspond to the more general features in the source data -- and are expected to be more similar to the general features of the target data. This recalls the transfer learning practice in deep learning where the first layers of a source DNN are often transferred to the target DNN learning --- under the assumption that the general, low-level features are more beneficial to transfer than the deeper layers that correspond to fine details and need more degrees of freedom in learning the target data.

The expression in (\ref{eq:TL-PCA-P - penalty function - projection F norm}) can be interpreted as the orientation between the two examined subspaces in terms of their  principal angles. More details and references to related literature are provided in Section \ref{subsec:Principal Angles: Definition and Formulation}. 

We propose to formulate the TL-PCA-P as 
\begin{equation}
\label{eq:TL-PCA-P - optimization - minimization problem}
    \small{\mtx{U}_k = \underset{\substack{\mtx{W} \in \mathbb{R}^{d \times k}\\\mtx{W}^T \mtx{W} = \mtx{I}_k}}{\arg\min} \left\lbrace \frac{1}{n} \Frobnormsquared{ (\mtx{I}_d - \mtx{W}\mtx{W}^T) \mtx{X} } + \frac{\alpha}{2} h(\mtx{W};\widetilde{\mtx{U}}_m) \right\rbrace}
\end{equation}
where $\mtx{X}$ is the $d\times n$ data matrix of the target task, and $\alpha>0$ is a hyperparameter that determines the strength of the penalty (i.e., the transfer learning) in the overall solution. 

The TL-PCA-P minimization problem in (\ref{eq:TL-PCA-P - optimization - minimization problem}) can be developed into the maximization problem of 
\begin{equation}
    \label{eq:TL-PCA-P - optimization - maximization problem}
    \mtx{U}_k = \underset{\substack{\mtx{W} \in \mathbb{R}^{d \times k}\\\mtx{W}^T \mtx{W} = \mtx{I}_k}}{\arg\max} \left\lbrace  \mtxtrace{ \mtx{W}^T \left( \frac{1}{n}\mtx{X} \mtx{X}^T + \alpha \widetilde{\mtx{U}}_m \widetilde{\mtx{U}}_m^T \right) \mtx{W} } \right\rbrace
\end{equation}
The proof of the equivalence between (\ref{eq:TL-PCA-P - penalty function - projection F norm})-(\ref{eq:TL-PCA-P - optimization - minimization problem}) to (\ref{eq:TL-PCA-P - optimization - maximization problem}) is provided in Appendix \ref{proof_of_target_max_problem}.

The TL-PCA-P optimization in (\ref{eq:TL-PCA-P - optimization - maximization problem}) can be solved by eigendecomposition of the matrix 
\begin{equation}
\label{eq:matrix M definition for TL-PCA-P}
\mtx{M}_{\rm P} \triangleq \frac{1}{n}\mtx{X} \mtx{X}^T + \alpha \widetilde{\mtx{U}}_m \widetilde{\mtx{U}}_m^T.    
\end{equation}
Specifically, the solution of (\ref{eq:TL-PCA-P - optimization - maximization problem}) is $\mtx{U}_k=[\vec{u}_1,\dots,\vec{u}_k]$ where $\vec{u}_i$ is the eigenvector that corresponds to the $i^{\rm th}$ largest eigenvalue of $\mtx{M}_{\rm P}$. We refer to $\vec{u}_i$ as the $i^{\rm th}$ principal direction of the TL-PCA-P.


\subsubsection{Efficient Computation}
\label{subsubsec:TL-PCA-P - Computationally Efficient Implementation}
Recall that the computational complexity of standard eigendecomposition of a $d\times d$ matrix is $O(d^3)$. Therefore, for a high dimension $d$, the computational complexity of the  eigendecomposition of $\mtx{M}_{\rm P}$ can be very high. Yet, if $n + m \ll d$, we can reduce the computational complexity as follows. Define the $d\times(n + m)$ matrix $\mtx{R}=[\frac{1}{\sqrt{n}}\mtx{X}, \sqrt{\alpha} \widetilde{\mtx{U}}_m]$, for which $\mtx{R}\mtx{R}^T = \mtx{M}_{\rm P}$. Define the $(n + m)\times (n + m)$ matrix $\mtx{Q}=\mtx{R}^T \mtx{R}$. The eigenvector $\vecgreek{\rho}\in\mathbb{R}^{n + m}$ and eigenvalue $\lambda\in\mathbb{R}$ of $\mtx{Q}$ correspond to $\mtx{Q}\vecgreek{\rho}=\lambda\vecgreek{\rho}$. Due to the construction of $\mtx{Q}$ from $\mtx{R}$, we get that $\mtx{R}\vecgreek{\rho}$ and $\lambda$ are an eigenvector-eigenvalue pair of $\mtx{M}_{\rm P}$; this is proved by   
\begin{equation}
  \mtx{M}_{\rm P}\left(\mtx{R}\vecgreek{\rho}\right) = \mtx{R}\mtx{R}^T \mtx{R}\vecgreek{\rho}= \mtx{R}\mtx{Q}\vecgreek{\rho} = \mtx{R}\lambda\vecgreek{\rho}= \lambda\mtx{R}\vecgreek{\rho}.  
\end{equation}
This shows that, for $n + m \ll d$, we can reduce the overall computational cost by eigendecomposition of $\mtx{Q}$ with computational complexity of $O\left((n + m)^3\right)$ instead of directly eigendecomposing $ \mtx{M}_{\rm P}$ at complexity of $O(d^3)$. 

\subsubsection{Hyperparameters}
\label{subsubsec:TL-PCA-P - Hyperparameters}
For a target subspace dimension $k$, our TL-PCA-P has two hyperparameters: the penalty strength $\alpha>0$ and the transferred subspace dimension $m\in\{1,\dots,\widetilde{k}\}$. In practice, the values of $\alpha$ and $m$ can be selected from a predetermined grid using cross validation. We exemplify this in Section \ref{sec:Experimental Results}.

\subsection{TL-PCA-D: Transfer Learning using the Source Data}
\label{subsec:TL-PCA-D: Transfer Learning using the Source Data}
In this version of TL-PCA, we directly use the source data to augment the standard PCA of the target task. Directly using the source data, if available to the target task solver, might have benefits over using the source model that was independently pretrained on the same source data.

For learning the target subspace, we use here the $d\times \widetilde{n}$ source data matrix $\widetilde{\mtx{X}}$ to define a penalty function that measures the reconstruction error of representing the \textit{source} data in the learned target subspace: 
\begin{equation}
    \label{eq:TL-PCA-D - penalty function}
    g(\mtx{W};\widetilde{\mtx{X}}) \triangleq \frac{1}{\widetilde{n}} \Frobnormsquared{ (\mtx{I}_d - \mtx{W}\mtx{W}^T) \widetilde{\mtx{X}}  }
\end{equation}
where $\mtx{W}\in\mathbb{R}^{d\times k}$ is the candidate solution (with orthonormal columns) to the TL-PCA-D.

We propose to formulate the TL-PCA-D as 
\begin{equation}
\label{eq:TL-PCA-D - optimization - minimization problem}
    \mtx{U}_k = \underset{\substack{\mtx{W} \in \mathbb{R}^{d \times k}\\\mtx{W}^T \mtx{W} = \mtx{I}_k}}{\arg\min} \left\lbrace \frac{1}{n} \Frobnormsquared{ (\mtx{I}_d - \mtx{W}\mtx{W}^T) \mtx{X} } + \alpha g(\mtx{W};\widetilde{\mtx{X}}) \right\rbrace
\end{equation}
where $\mtx{X}$ is the $d\times n$ data matrix of the target task, and $\alpha>0$ is a hyperparameter that determines the strength of the penalty (i.e., the strength of transfer learning) in the overall solution. The value of $\alpha$ can be set by cross validation, as we show in Section \ref{sec:Experimental Results}.

The TL-PCA-D minimization problem in (\ref{eq:TL-PCA-D - optimization - minimization problem}) can be developed into the maximization problem of 
\begin{equation}
    \label{eq:TL-PCA-D - optimization - maximization problem}
    \mtx{U}_k = \underset{\substack{\mtx{W} \in \mathbb{R}^{d \times k}\\\mtx{W}^T \mtx{W} = \mtx{I}_k}}{\arg\max} \left\lbrace  \mtxtrace{ \mtx{W}^T \left( \frac{1}{n}\mtx{X} \mtx{X}^T +  \frac{\alpha}{\widetilde{n}} \widetilde{\mtx{X}} \widetilde{\mtx{X}}^T \right) \mtx{W} } \right\rbrace
\end{equation}
The proof of the equivalence between (\ref{eq:TL-PCA-D - penalty function})-(\ref{eq:TL-PCA-D - optimization - minimization problem}) to (\ref{eq:TL-PCA-D - optimization - maximization problem}) is provided in Appendix \ref{appendix:sec:Proof of TL-PCA-D Maximization Form}.

The TL-PCA-D optimization in (\ref{eq:TL-PCA-D - optimization - maximization problem}) can be solved by eigendecomposition of the matrix 
\begin{align}
\label{eq:matrix M definition for TL-PCA-D}
\mtx{M}_{\rm D} &\triangleq \frac{1}{n}\mtx{X} \mtx{X}^T + \frac{\alpha}{\widetilde{n}} \widetilde{\mtx{X}} \widetilde{\mtx{X}}^T  = \widehat{\mtx{C}}_\mtx{X} + \alpha \widehat{\mtx{C}}_{\widetilde{\mtx{X}}}.
\end{align}
Specifically, the solution of (\ref{eq:TL-PCA-D - optimization - maximization problem}) is $\mtx{U}_k=[\vec{u}_1,\dots,\vec{u}_k]$ where $\vec{u}_i$ is the eigenvector that corresponds to the $i^{\rm th}$ largest eigenvalue of $\mtx{M}_{\rm D}$. We refer to $\vec{u}_i$ as the $i^{\rm th}$ principal direction of the TL-PCA-D. 

The computational complexity of TL-PCA-D can be reduced when $n+\widetilde{n} \ll d$. This is technically similar to the procedure explained for TL-PCA-P in Section \ref{subsubsec:TL-PCA-P - Computationally Efficient Implementation}, but here we use the $d\times(n + \widetilde{n})$ matrix $\mtx{R}=[\frac{1}{\sqrt{n}}\mtx{X}, \sqrt{\frac{\alpha}{\widetilde{n}}} \widetilde{\mtx{X}}]$ for which $\mtx{R}\mtx{R}^T = \mtx{M}_{\rm D}$.

\section{Experimental Results}
\label{sec:Experimental Results}

\begin{table*}[h]
\caption{Test error rates for Omniglot-to-MNIST with different methods at various values of $k$}
\label{table_mnist}
\begin{center}
\footnotesize
\begin{tabular}
{p{2.5cm}p{1.3cm}p{1.3cm}p{1.3cm}p{1.3cm}p{1.3cm}p{1.3cm}p{1.3cm}p{1.3cm}}
\toprule
& \multicolumn{1}{c}{$k=10$} & \multicolumn{1}{c}{$k=20$} & \multicolumn{1}{c}{$k=30$} & \multicolumn{1}{c}{$k=40$} & \multicolumn{1}{c}{$k=50$} & \multicolumn{1}{c}{$k=70$} & \multicolumn{1}{c}{$k=100$} & \multicolumn{1}{c}{$k=150$} \\
\midrule
TL-PCA-D (CV)& \textbf{59.34}\tiny{$\pm1.15$} & \textbf{45.66}\tiny{$\pm0.59$} & \textbf{36.73}\tiny{$\pm0.42$} & \textbf{30.29}\tiny{$\pm0.33$} & \textbf{25.67}\tiny{$\pm0.33$} & \textbf{19.61}\tiny{$\pm0.23$} & \textbf{14.11}\tiny{$\pm0.17$} & \textbf{9.22}\tiny{$\pm0.10$} \\
TL-PCA-P (CV)& \textbf{59.93}\tiny{$\pm0.93$} & \textbf{46.31}\tiny{$\pm0.60$} & 38.12\tiny{$\pm0.33$} & 31.83\tiny{$\pm0.49$} & 26.86\tiny{$\pm0.40$} & 20.60\tiny{$\pm0.32$} & 14.95\tiny{$\pm0.26$} & 9.83\tiny{$\pm0.15$} \\
\midrule 
Pretrained source \tiny{(w.~target mean)} &
71.31\tiny{$\pm0.81$} & 54.95\tiny{$\pm1.12$} & 43.19\tiny{$\pm0.80$} & 35.44\tiny{$\pm0.63$} & 30.03\tiny{$\pm0.51$} & 22.65\tiny{$\pm0.32$} & 16.20\tiny{$\pm0.23$} & 10.42\tiny{$\pm0.12$}  \\
Pretrained source \tiny{(w.~source mean)} &
69.34\tiny{$\pm3.33$} & 63.13\tiny{$\pm2.23$} & 57.94\tiny{$\pm1.58$} & 54.47\tiny{$\pm1.58$} & 51.88\tiny{$\pm1.44$} & 46.48\tiny{$\pm2.20$} & 40.32\tiny{$\pm2.33$} & 34.75\tiny{$\pm3.11$}  \\
Standard PCA & 
\textbf{60.83}\tiny{$\pm0.88$} & 49.07\tiny{$\pm1.18$} & 42.49\tiny{$\pm1.20$} & 38.29\tiny{$\pm0.93$} & 36.68\tiny{$\pm1.19$} & 34.73\tiny{$\pm1.04$} & 33.07\tiny{$\pm1.11$} & 29.85\tiny{$\pm1.37$}  \\

\hline
\end{tabular}
\end{center}
\end{table*}

\begin{table}[h]
\caption{Test error rates for CIFAR10-to-SVHN with different methods at various values of $k$ (see Table \ref{app:table_svhn})}
\label{table_svhn}
\begin{center}
\footnotesize
\begin{tabular}
{p{2.5cm}p{1.3cm}p{1.3cm}p{1.3cm}p{1.3cm}}
\toprule
 & \multicolumn{1}{c}{$k=30$} &  \multicolumn{1}{c}{$k=50$} & \multicolumn{1}{c}{$k=100$} \\
\midrule
TL-PCA-D (CV) &  \textbf{8.92}\tiny{$\pm0.23$} & \textbf{5.77}\tiny{$\pm0.12$} &  \textbf{3.05}\tiny{$\pm0.07$} \\ 
TL-PCA-P (CV) &  9.65\tiny{$\pm0.29$} &  6.36\tiny{$\pm0.18$} &  3.33\tiny{$\pm0.07$} \\
\midrule
Pretrained source \tiny{(w.~target mean)}  & 10.39\tiny{$\pm0.17$} &  6.93\tiny{$\pm0.14$} &  3.87\tiny{$\pm0.09$} \\
Pretrained source \tiny{(w.~source mean)}  & 10.12\tiny{$\pm0.12$} &  6.75\tiny{$\pm0.10$} & 3.77\tiny{$\pm0.07$} \\
Standard PCA &  11.48\tiny{$\pm0.38$} &  8.79\tiny{$\pm0.23$} &  8.46\tiny{$\pm0.23$} \\ 
\hline
\end{tabular}
\end{center}
\end{table}

\begin{table*}[h]
\caption{Test error rates for  Tiny ImageNet-to-CelebA with different methods at various values of $k$}
\label{table_celeba}
\begin{center}
\footnotesize
\begin{tabular}
{p{2.5cm}p{1.3cm}p{1.3cm}p{1.3cm}p{1.3cm}p{1.3cm}p{1.3cm}p{1.3cm}p{1.3cm}}
\toprule
& \multicolumn{1}{c}{$k=10$} & \multicolumn{1}{c}{$k=20$} & \multicolumn{1}{c}{$k=30$} & \multicolumn{1}{c}{$k=40$} & \multicolumn{1}{c}{$k=50$} & \multicolumn{1}{c}{$k=70$} & \multicolumn{1}{c}{$k=100$} & \multicolumn{1}{c}{$k=150$} \\
\midrule
TL-PCA-D (CV) & \textbf{33.50}\tiny{$\pm0.28$} & \textbf{25.42}\tiny{$\pm0.20$} & \textbf{21.63}\tiny{$\pm0.13$} & \textbf{19.23}\tiny{$\pm0.08$} & \textbf{17.47}\tiny{$\pm0.11$} & \textbf{15.12}\tiny{$\pm0.11$} & \textbf{13.07}\tiny{$\pm0.08$} & \textbf{11.08}\tiny{$\pm0.11$} \\ 
TL-PCA-P (CV) & \textbf{33.88}\tiny{$\pm0.26$} & 26.16\tiny{$\pm0.12$} & 22.38\tiny{$\pm0.18$} & 19.81\tiny{$\pm0.11$} & 17.98\tiny{$\pm0.09$} & 15.67\tiny{$\pm0.07$} & 13.57\tiny{$\pm0.09$} & 11.55\tiny{$\pm0.09$} \\
\midrule
Pretrained source (w.~target mean) & 37.15\tiny{$\pm0.39$} & 28.51\tiny{$\pm0.24$} & 23.84\tiny{$\pm0.22$} & 21.00\tiny{$\pm0.18$} & 19.10\tiny{$\pm0.18$} & 16.65\tiny{$\pm0.17$} & 14.54\tiny{$\pm0.14$} & 12.60\tiny{$\pm0.14$} \\
Pretrained source (w.~source mean) & 38.70\tiny{$\pm0.57$} & 29.27\tiny{$\pm0.18$} & 24.04\tiny{$\pm0.17$} & 21.04\tiny{$\pm0.21$} & 19.03\tiny{$\pm0.24$} & 16.48\tiny{$\pm0.14$} & 14.30\tiny{$\pm0.10$} & 12.32\tiny{$\pm0.10$} \\
Standard PCA & 34.87\tiny{$\pm0.28$} & 27.74\tiny{$\pm0.24$} & 24.63\tiny{$\pm0.22$} & 22.73\tiny{$\pm0.25$} & 21.38\tiny{$\pm0.25$} & 19.51\tiny{$\pm0.19$} & 17.96\tiny{$\pm0.12$} & 17.80\tiny{$\pm0.15$} \\ 
\hline
\end{tabular}
\end{center}
\end{table*}

\begin{figure}
    \centering
    \subfloat[Omniglot to MNIST]{\includegraphics[width=0.48\linewidth]{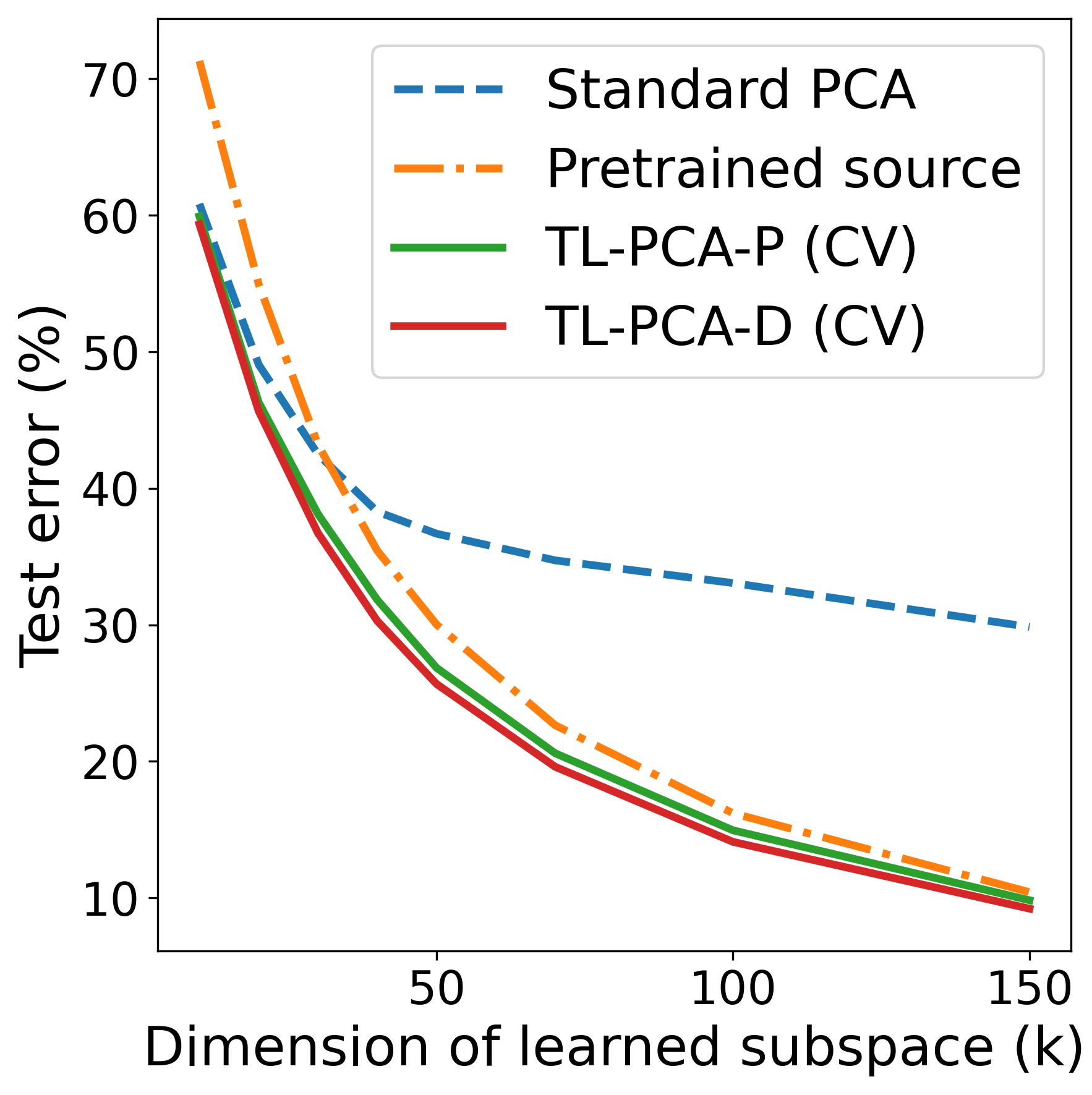}}
    \subfloat[CIFAR-10 to SVHN]{\includegraphics[width=0.48\linewidth]{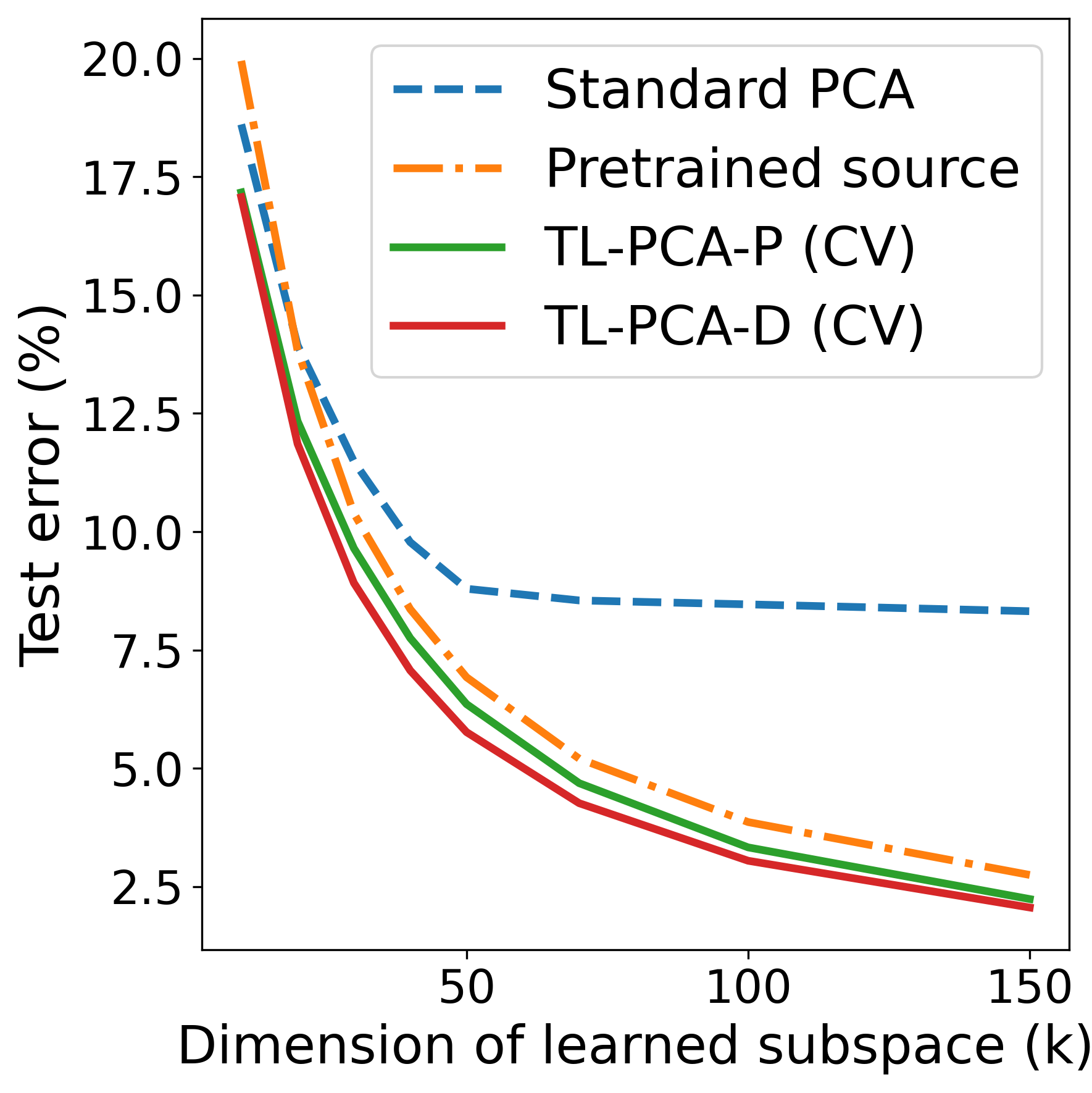}}
    \caption{Error comparison of PCA and TL-PCA. }
    \label{fig:error_graphs_main_paper}
\end{figure}

\begin{figure}[t]
    \centering
    \includegraphics[width=0.98\linewidth]{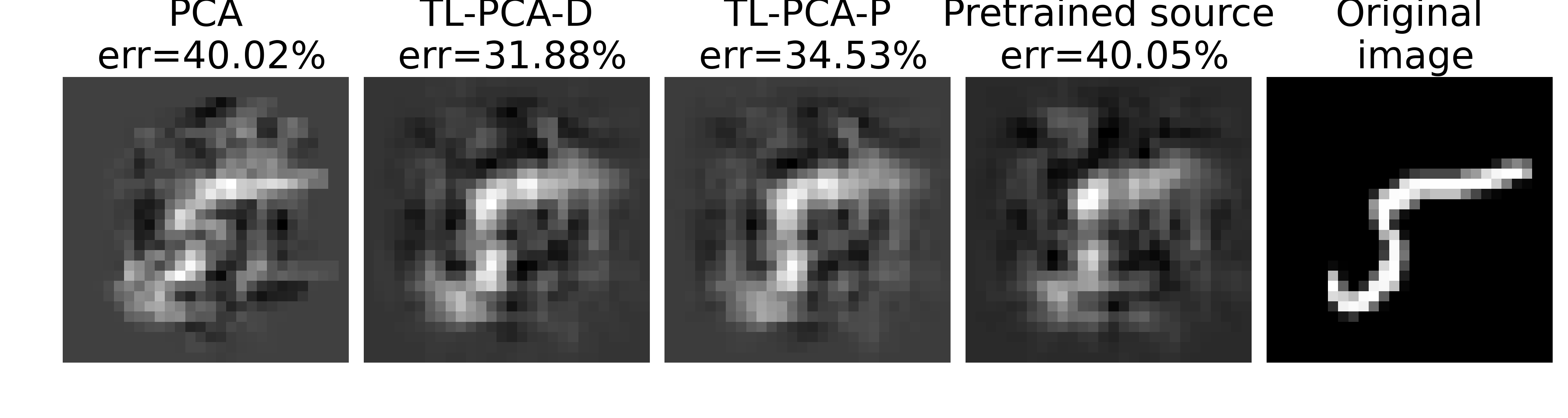}\\
    \caption{Comparison of the reconstructions of an MNIST test example ($k=40$).}
    \label{fig:MNIST_rec_8}
\end{figure}

\begin{figure}[t]
    \centering
    \includegraphics[width=0.98\linewidth]{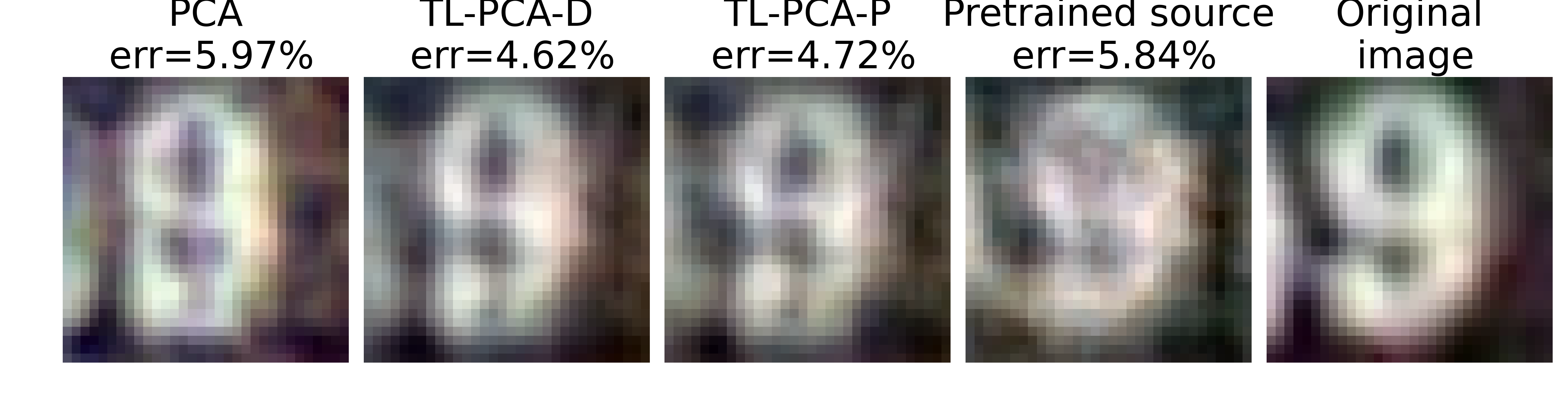}
    \caption{Comparison of the reconstructions of a SVHN test example ($k=40$).}
    \label{fig:SVHN_rec_1}
\end{figure}

\begin{figure}[t]
    \centering
    \includegraphics[width=0.98\linewidth]{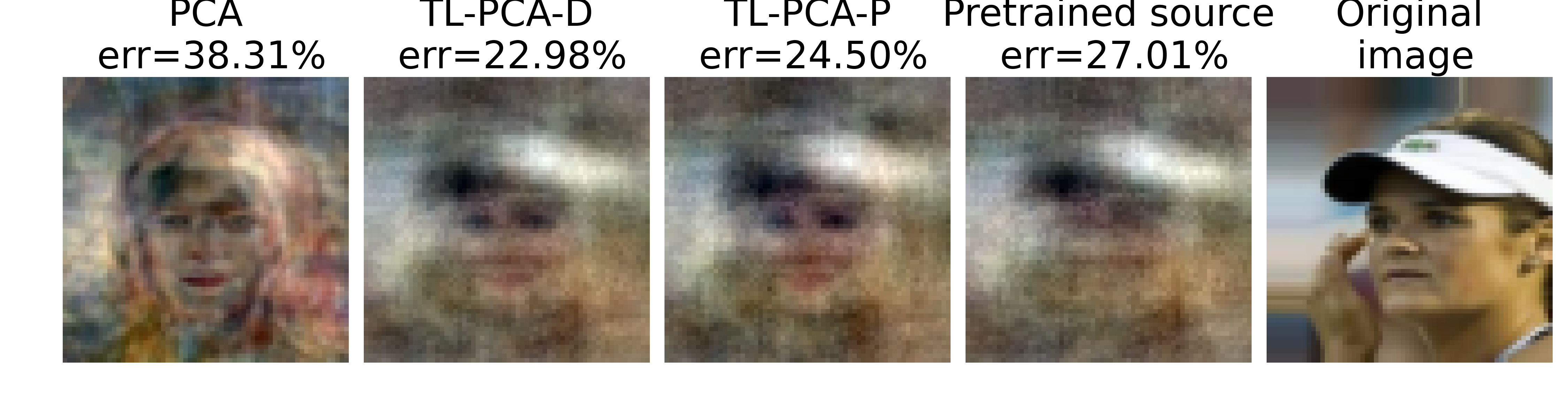}
    \caption{Comparison of the reconstructions of a CelebA test example ($k=80$).}
    \label{fig:CelebA_rec_1}
\end{figure}

\subsection{Experiment Setting}

\subsubsection{Datasets and Their Source-Target Pairing}
In our experiments, we evaluated the performance of our TL-PCA approach using three pairs of datasets. For each pair, we randomly sampled examples for the source and target train datasets (in sizes mentioned below for each experiment), along with 2000 test examples.
We repeated the experiments 8 times, each time with a different randomly-selected source and target train datasets; the average errors are reported in a normalized form (as explained below in Section \ref{subsubsec:Experiments - Error Evaluation}), in percentages, together with their standard deviation across the repeated experiments. 
The datasets and their source-target pairing\footnote{More details are available in Appendix \ref{appendix:subsec:Additional Experimental Details - Datasets}.}:
\begin{itemize}
    \item \textbf{Omniglot to MNIST:} Omniglot contains a diverse set of handwritten characters across 50 alphabets, serving as the source dataset. MNIST, with its grayscale images of handwritten digits (0–9), is used as the target dataset. We used $\widetilde{n}=500$ examples from Omniglot and $n=50$ examples from MNIST. 
    The images here are of 28x28 grayscale pixels, hence, the data dimension is $d=784$.
    \item \textbf{CIFAR-10 to SVHN:} CIFAR-10 is a dataset of images from 10 object categories such as animals and vehicles. SVHN consists of digit images extracted from street house numbers. We used $\widetilde{n}=500$ source examples from CIFAR-10 and $n=50$ target examples from SVHN. The images here are of 32x32 color pixels, hence, the data dimension is $d=3072$.
    \item \textbf{Tiny ImageNet to CelebA:} Tiny ImageNet contains 64x64 pixel color images across 200 object categories, whereas CelebA consists of face images. We used $\widetilde{n}=1000$  source examples from Tiny ImageNet and $n=100$ target examples from CelebA, downsampled to 64x64 pixel images. The data dimension here is $d=12,288$.    
\end{itemize}

\subsubsection{Cross Validation}
Unless otherwise specified, we use cross validation (CV) to choose the \( \alpha \) hyperparameter from the value grid of \(\{0.01, 0.1, 1, 5, 10, 50, 100, 1000\}\). For TL-PCA-P, we use the cross validation to jointly choose $\alpha$ (from the mentioned grid) and the hyperparameter $m$ that defines how many principal directions are transferred from the pretrained source model. For a given $k$, the value grid for $m$ is defined as \(\{ \left\lceil 0.2\cdot k \right\rceil, \left\lceil 0.5\cdot k \right\rceil, \left\lceil 0.8\cdot k \right\rceil, k\}\), i.e., the grid examines transfer of 20\%, 50\%, 80\%, 100\% of the $k$ principal directions of the pretrained source model; for Tiny ImageNet to CelebA, the 20\% option is excluded. We used 10-fold cross-validation for all dataset pairs.

\subsubsection{Data Centering}
Recall that the data used for PCA and TL-PCA computation is centered. The target (train) data is centered w.r.t. its mean $\vecgreek{\mu}_{\mathcal{D}}$, and the source (train) data is centered w.r.t. its mean $\vecgreek{\mu}_{\widetilde{\mathcal{D}}}$. 
For a given test data $\vec{x}$ for the target dataset, the mean is subtracted before the dimensionality reduction, and added back after the reconstruction -- to create $\widehat{\vec{x}}$. For standard PCA and our TL-PCA versions, the centering is naturally using the target train mean $\vecgreek{\mu}_{\mathcal{D}}$. For the baseline of using the pretrained source model for the target test data, we found that it is sometimes better to center using the source train mean $\vecgreek{\mu}_{\widetilde{\mathcal{D}}}$; accordingly, for this pretrained source usage we report in the error \textit{tables} results for each of the mean subtraction options. In the error \textit{graphs}, for visual clarity, the pretrained source is reported only with one mean subtraction option (it is reported with the source mean subtraction for all datasets, except for Omniglot-to-MNIST where it is with the target mean subtraction). 

\subsubsection{Error Evaluation}
\label{subsubsec:Experiments - Error Evaluation}
For performance comparison, we evaluate the squared errors of the reconstructions for each of the train and test datasets according to the following formulas. For a given dataset $\mathcal{D}$ of $n$ examples, the normalized squared error is 
\begin{equation}
\label{eq:empirical error formulation}
    \widehat{\mathcal{E}}_{\mathcal{D}}\triangleq \frac{\sum_{i=1}^{n}\Ltwonormsquared{\vec{x}_i - \widehat{\vec{x}}_i}}{\sum_{i=1}^{n}\Ltwonormsquared{\vec{x}_i}}
\end{equation}
where $\widehat{\vec{x}}_i$ is the reconstruction of the $i^{\rm th}$ example $\vec{x}_i$ from the dataset $\mathcal{D}$. This error formula is computed separately  for each dimensionality reduction operator of interest. The dataset  $\mathcal{D}$ and its size $n$ are according to whether the error is computed for a test or train dataset.
Note that (\ref{eq:empirical error formulation}) is the relative error w.r.t.~the overall variance of the original data; accordingly, this formula value is in the range $[0,1]$. For better readability of the errors and their standard deviations, the graphs and tables report the errors in percentages between 0\% and 100\%. We used (\ref{eq:empirical error formulation}) to report the errors in the tables and graphs (Tables \ref{table_mnist}-\ref{table_celeba}, \ref{app:table_svhn}, and Figures \ref{fig:error_graphs_main_paper}, \ref{appendix:fig:error_graphs_mnist_train_test_errors}, \ref{appendix:fig:error_graphs_svhn_train_test_errors}, \ref{appendix:fig:error_graphs_celeba_train_test_errors}).

In visual comparisons of specific test examples (Figures \ref{fig:MNIST_rec_8}, \ref{fig:SVHN_rec_1}, \ref{fig:CelebA_rec_1}, \ref{fig:mnist_rec_k40}, \ref{fig:svhn_rec_k40}, \ref{appendix:fig:celeba_rec_k80_5}), we report the numerical error of the specific test example normalized by its own norm, i.e., $\widehat{\mathcal{E}}_{\vec{x}}\triangleq \frac{\Ltwonormsquared{\vec{x} - \widehat{\vec{x}}}}{\Ltwonormsquared{\vec{x}}}$.

\subsection{Comparison of Methods}

In Tables \ref{table_mnist}-\ref{table_celeba}, \ref{app:table_svhn}, and Figures \ref{fig:error_graphs_main_paper}, \ref{appendix:fig:error_graphs_mnist_train_test_errors}, \ref{appendix:fig:error_graphs_svhn_train_test_errors}, \ref{appendix:fig:error_graphs_celeba_train_test_errors}, we compare the test error of various PCA methods for a varying dimension $k$ of the subspace dimension. These results demonstrate that our TL-PCA performs better than the standard PCA of the target task, and using directly the pretrained (standard) PCA of the source task. All the methods, including the pretrained (standard) PCA of the source task, are evaluted for their test performance on the target test dataset. 

Our results show that TL-PCA-P and TL-PCA-D usually perform better than the alternatives for the entire examined range of subspace dimension $k$, including both $k<n$ and $k>n$. As expected, the direct use of source data helps TL-PCA-D to perform better than TL-PCA-P. 

As for the alternative methods, for a small $k$, the standard target PCA has relatively enough examples to be better than directly using the source pretrained model for the target data. For a larger $k$, the source pretrained model\footnote{For the source pretrained model evaluation, in this discussion we refer to the better between the two options of subtracting the source train mean or subtracting the target train mean from the test example.} is usually better than the standard target PCA. The TL-PCA perform better than both of these alternatives, but it is noticeable that for $k$ sufficiently larger than $n$ the TL-PCA performance might converge to the performance of the source pretrained model. This is expected, as for $k\gg n$, the $n$ target (centered) examples span a $(n-1)$-dimensional subspace that has a relatively small effect on a much larger target subspace of dimension $k$.

In addition to the quantitative comparison of errors, the better performance of the TL-PCA methods is visually observed in Figures \ref{fig:MNIST_rec_8}, \ref{fig:SVHN_rec_1}, \ref{fig:CelebA_rec_1}, \ref{fig:mnist_rec_k40}, \ref{fig:svhn_rec_k40}, \ref{appendix:fig:celeba_rec_k80_5}.

\section{The TL-PCA Benefits as Reflected by Principal Angles}
\label{sec:The TL-PCA Benefits as Reflected by Principal Angles}

\subsection{Principal Angles: Definition and Formulation}
\label{subsec:Principal Angles: Definition and Formulation}
Consider two linear subspaces $\Omega$ and $\widetilde{\Omega}$ in $\mathbb{R}^d$, where ${\rm dim}(\Omega)=k$, ${\rm dim}(\widetilde{\Omega})=\widetilde{k}$, and $k \le \widetilde{k} < d $. The principal angles $0 \leq \theta_1 \leq \theta_2 \leq \dots \leq \theta_k \leq \pi/2$ between the subspaces $\Omega$ and $\widetilde{\Omega}$ are defined recursively for $j=1,\dots,k$ by
\begin{align}
\label{eq:principal angles - definition by optimization}
& \cos \theta_j = \max_{\vec{v}\in\Omega, \widetilde{\vec{v}}\in\widetilde{\Omega}} \vec{v}^T \widetilde{\vec{v}}
\\\nonumber
& \text{subject to}~\Ltwonorm{\vec{v}} = 1, \, \Ltwonorm{\widetilde{\vec{v}}} = 1 
\\\nonumber
&\quad\quad \vec{v}^T \vec{v}_s = 0, \, \widetilde{\vec{v}}^T \widetilde{\vec{v}}_s = 0 \, ~~\text{for}~~ \,  s=1,2,\dots,j-1
\end{align}
where $\vec{v}_j$ and $\widetilde{\vec{v}}_j$ are the principal vectors that solve the optimization for the $j^{\rm th}$ principal angle $\theta_j$. 
Intuitively, the first principal angle $\theta_1$ is the smallest angle between all pairs of unit vectors in the first and second subspaces, and the rest of the principal angles are defined similarly.

Consider a matrix $\mtx{U} \in \mathbb{R}^{d \times k}$ whose columns are orthonormal and span the subspace $\Omega$, and a matrix $\widetilde{\mtx{U}} \in \mathbb{R}^{d \times \widetilde{k}}$ whose columns are orthonormal and span the subspace $\widetilde{\Omega}$.
The singular values of 
\begin{equation}
\label{eq:matrix whose singular values are the cosines of principal angles}
    \mtx{B} = \mtx{U}^T \widetilde{\mtx{U}} \in \mathbb{R}^{k \times \widetilde{k}},
\end{equation} 
denoted as $\sigma_1 \geq \sigma_2 \geq \dots \geq \sigma_k$, are related to the principal angles by
\begin{equation}
\label{eq:principal angles - singular values relation}
\cos \theta_j = \sigma_j  \, ~~\text{for}~~ \, j=1, \dots, k.    
\end{equation}
The proof of (\ref{eq:principal angles - singular values relation}) was given by \citet{bjorck1973numerical}. 

The literature includes several definitions for the distance between two linear subspaces \citep{grassmann2008hamm}. Here we consider the definition of distance between the subspaces using their projection matrices $\mtx{U}\mtx{U}^T$ and $\widetilde{\mtx{U}}\widetilde{\mtx{U}}^T$: 
\begin{equation}
\label{eq: projection distance - definition}
{\rm dist}(\mtx{U}, \widetilde{\mtx{U}}) = \frac{1}{\sqrt{2}} \Frobnorm{ \mtx{U} \mtx{U}^T - \widetilde{\mtx{U}} \widetilde{\mtx{U}}^T }
\end{equation}
that relates to the principal angles by
\begin{equation}
\label{eq: projection distance - formulation using principal angles}
{\rm dist}(\mtx{U}, \widetilde{\mtx{U}}) = \sqrt{ \frac{\widetilde{k}-k}{2} + \sum_{j=1}^{k} \sin^2 \theta_j}.
\end{equation}
The proof of (\ref{eq: projection distance - formulation using principal angles}) is based on algebraically developing the Frobenius norm in (\ref{eq: projection distance - definition}) using the relation (\ref{eq:principal angles - singular values relation}) between the singular values of $\mtx{U}^T \widetilde{\mtx{U}}$ and the principal angles $\{\theta_j\}_{j=1}^{k}$.
Eq.~(\ref{eq: projection distance - formulation using principal angles}) shows that the distance between the two subspaces decreases together with their principal angles. This subspace distance definition appeared also in the works by \citet{panagopoulos2016constrained, grassmann2008hamm, edelman1998geometry}.

\begin{figure}[t]
    \centering
    \subfloat[$k=20$]{\includegraphics[width=0.48\linewidth]{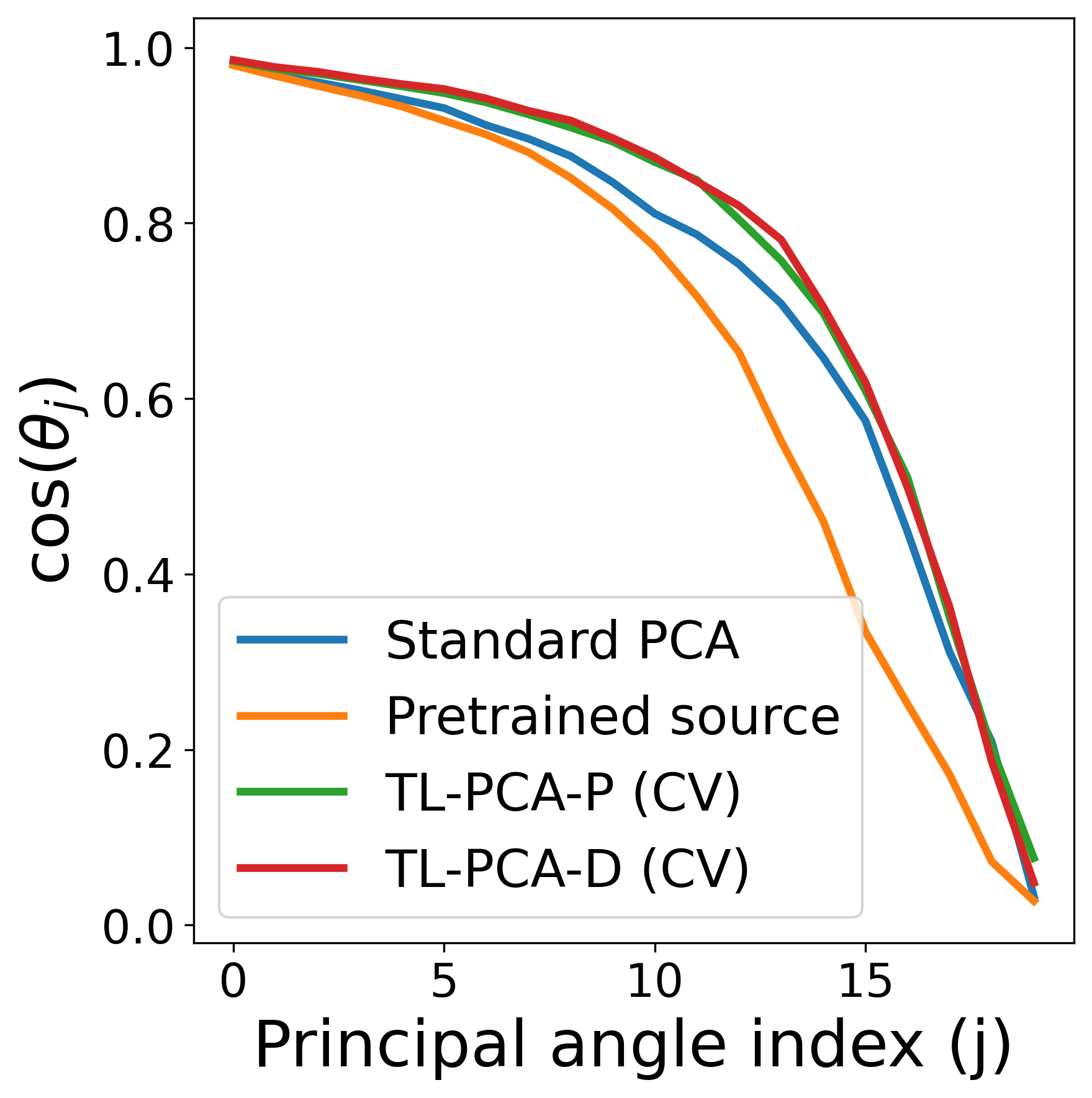}}
    \subfloat[$k=50$]{\includegraphics[width=0.48\linewidth]{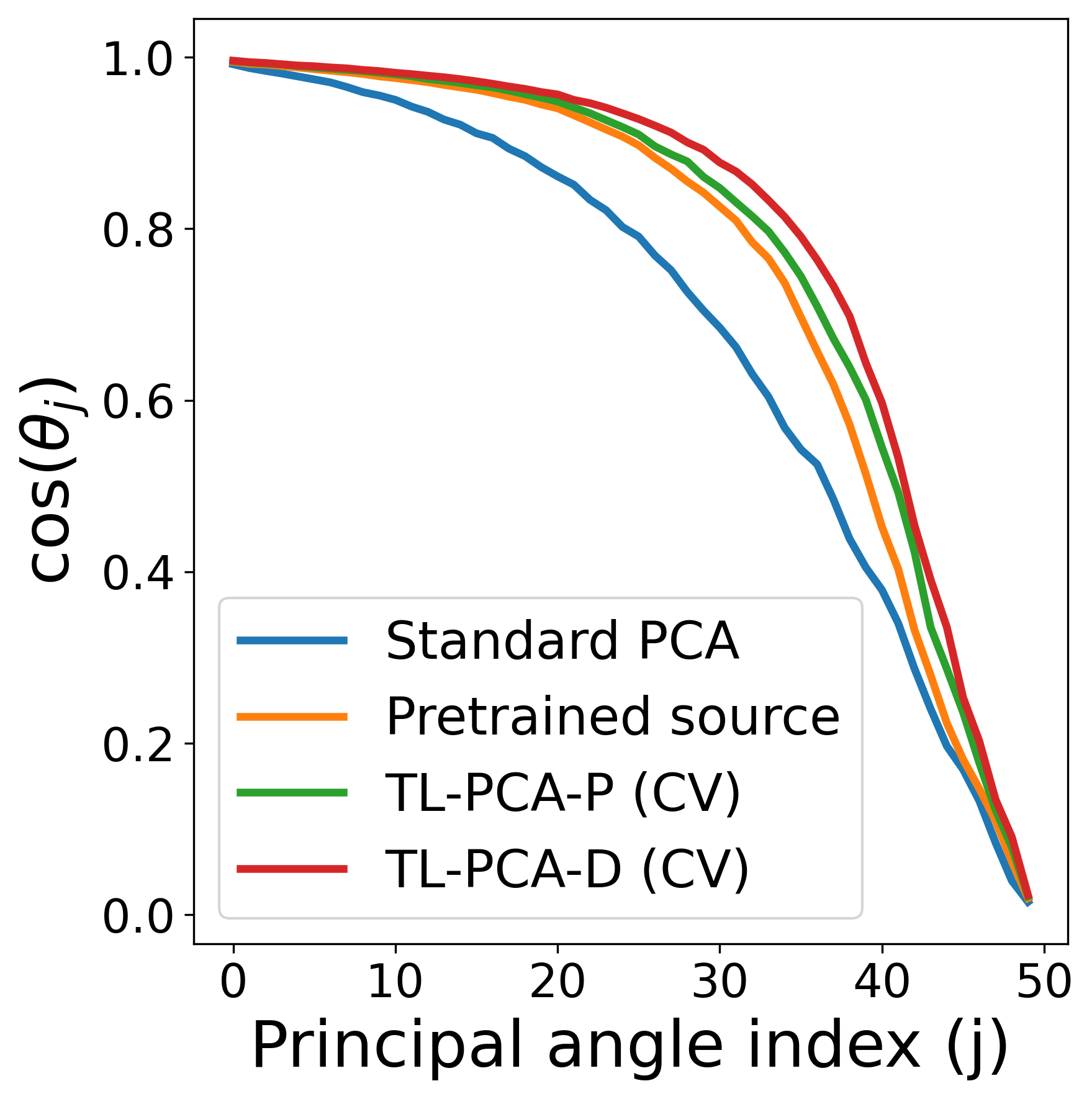}}
    \caption{Principal angles w.r.t.~ideal PCA (Omniglot to MNIST). }
    \label{fig:principal_angles_graphs_main_paper}
\end{figure}
\begin{figure}[t]
    \centering
    \subfloat[$k=50$]{\includegraphics[width=0.48\linewidth]{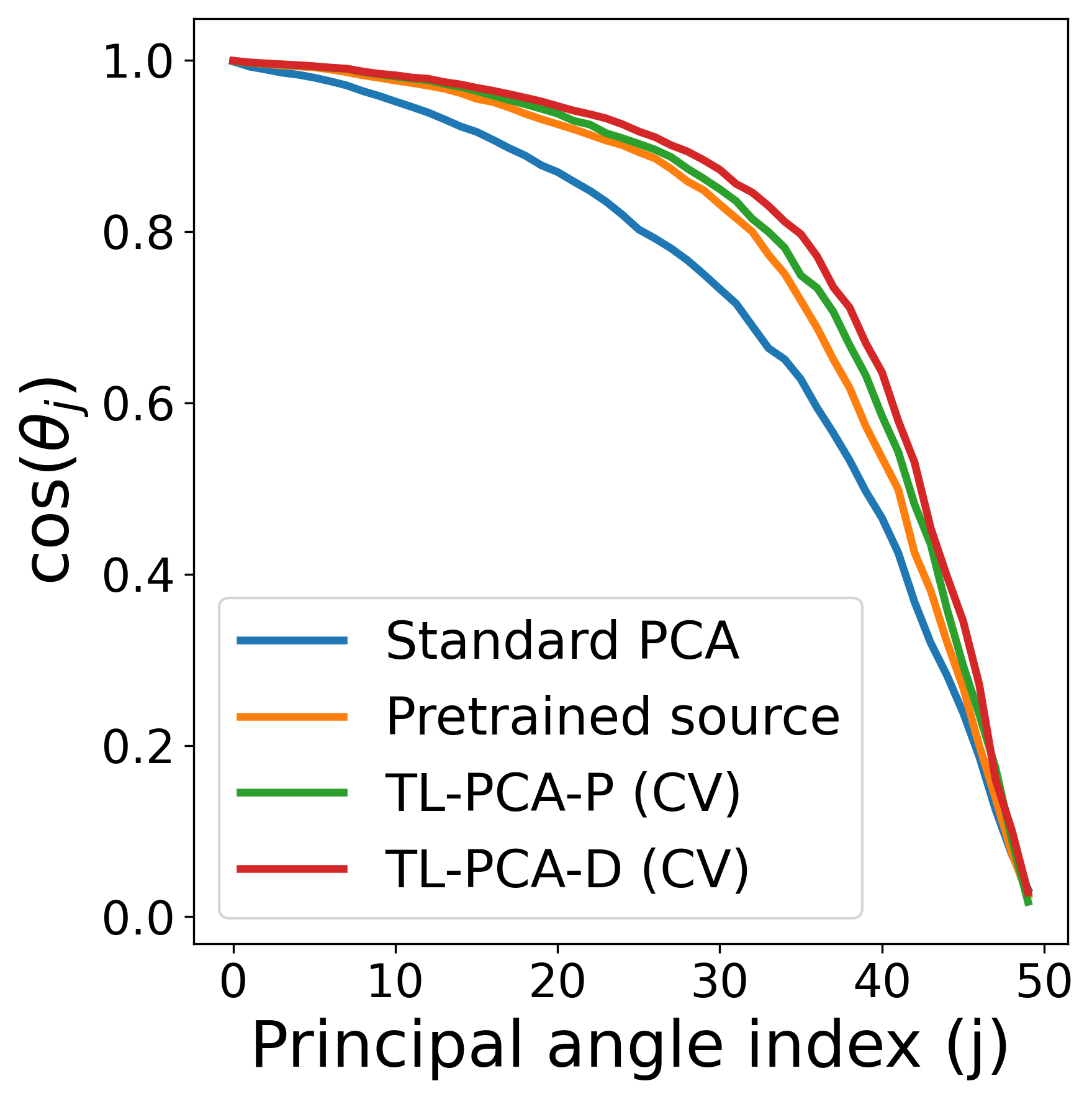}}
    \subfloat[$k=150$]{\includegraphics[width=0.48\linewidth]{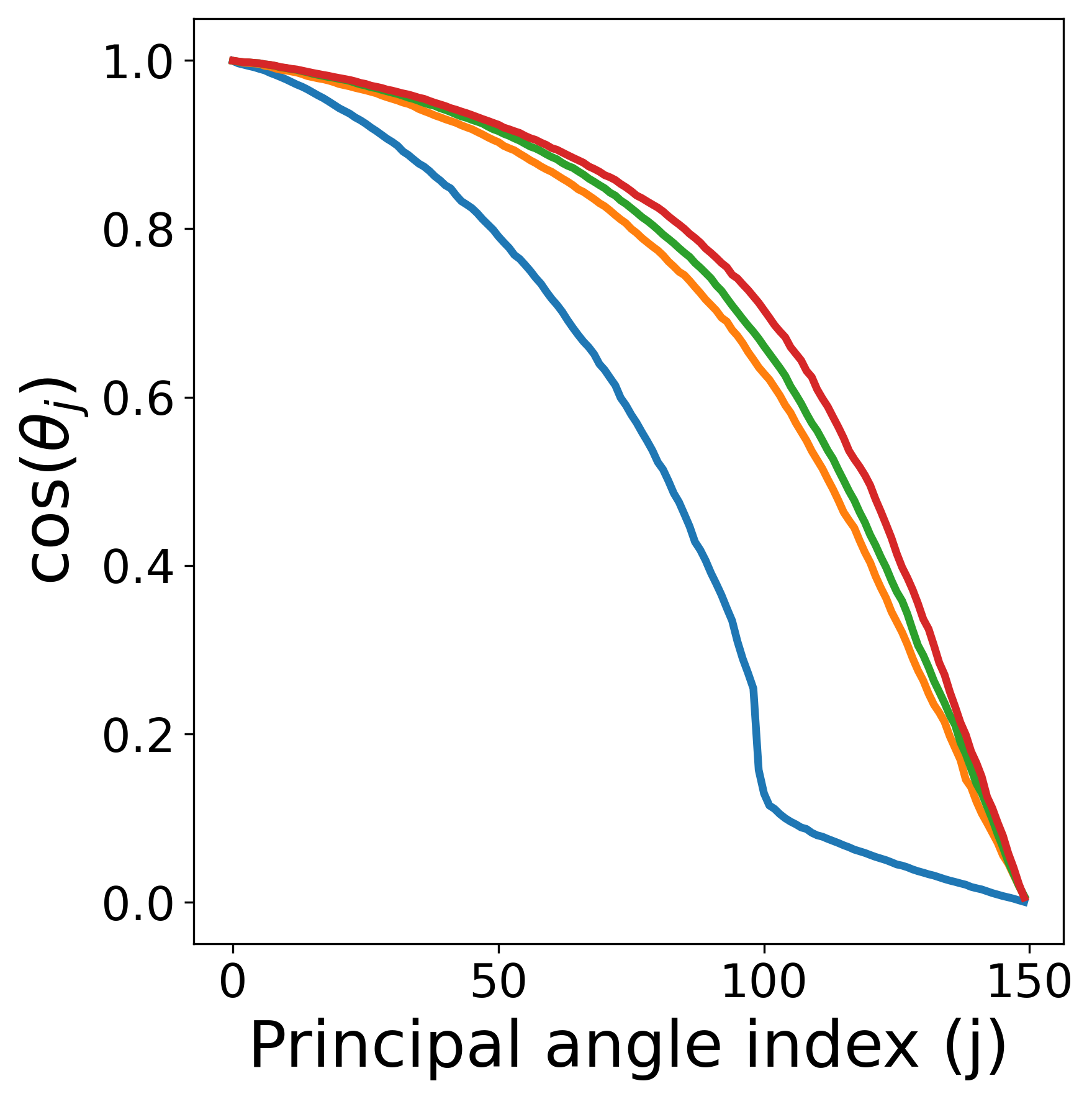}}    
    \caption{Principal angles w.r.t.~ideal PCA (Tiny ImageNet to CelebA). }
    \label{fig:principal_angles_graphs_main_paper_celeba}
\end{figure}

\subsection{Usefulness of the Source Task in Improving the Learned Subspace}


Using the source task knowledge, either through the pretrained source model (in TL-PCA-P) or directly through the source covariance matrix (in TL-PCA-D), the rank of the eigendecomposed matrix (i.e., $\mtx{M}_{\rm P}$ or $\mtx{M}_{\rm D}$) is usually higher than in standard PCA. This results in principal vectors that stem from a larger amount of information, taken from both the target and source tasks. The main question is whether the  principal directions of TL-PCA are useful for representing test data of the \textit{target} task. In Section \ref{sec:Experimental Results} we showed that TL-PCA indeed provides useful principal directions that can reduce the reconstruction error. 

In this section, we give a principal-angles perspective on the usefulness of the TL-PCA to mitigate lack of data examples in the target task. 
We assume that the $n$ data examples given in $\mathcal{D}$ for the target task are too few, and ideally we wish to have much more target data examples $n_{\rm ideal}\gg n$ in an ideal dataset $\mathcal{D}_{\rm ideal}$. Namely, for $n_{\rm ideal}$, the sample covariance matrix would much better approximate the true covariance matrix of the distribution $P_{\vec{x}}$, and accordingly the standard PCA for $\mathcal{D}_{\rm ideal}$ would provide an excellent $k$-dimensional subspace for the target data representation. 
Accordingly, we use principal angles to answer the following question: \textit{Can TL-PCA learn a subspace more similar to the subspace learned by standard PCA of a much larger target dataset?} 

In Figures \ref{fig:principal_angles_graphs_main_paper}, \ref{fig:principal_angles_graphs_main_paper_celeba}, \ref{appendix:fig:principal_angles_graphs_mnist},  \ref{appendix:fig:principal_angles_graphs_svhn}, \ref{appendix:fig:principal_angles_graphs_celeba} we show the principal angles computed between an ``ideal'' subspace learned by standard PCA of $n_{\rm ideal}=2000$ target examples to each of the subspaces learned by (i) standard PCA of $n=50$ target examples, (ii) TL-PCA-P and TL-PCA-D of $n=50$ target examples, (iii) pretrained source model by standard PCA of $\widetilde{n}=500$ source examples.  The graphs show the average cosine value of each of the first $80$ principal angles, hence, a larger cosine value implies a smaller angle and smaller distance of the examined subspace from the ideal subspace. 

Figures \ref{fig:principal_angles_graphs_main_paper}, \ref{fig:principal_angles_graphs_main_paper_celeba}, \ref{appendix:fig:principal_angles_graphs_mnist},  \ref{appendix:fig:principal_angles_graphs_svhn}, \ref{appendix:fig:principal_angles_graphs_celeba} confirm that our TL-PCA approach learns a subspace which is much closer to the subspace learned by standard PCA of much more target examples. For a middle range of angle index $j$, the TL-PCA-D can get closer than TL-PCA-P to the ideal subspace. 
The results show that the source pretrained subspace is relatively farther from the ideal target subspace for a lower dimension $k$, but it might be relatively closer to the angles of TL-PCA-P (but not TL-PCA-D) for larger $k>n$.
Yet, both TL-PCA-D and TL-PCA-P show excellent ability to mitigate the lack of target data examples and perform better than the alternatives for a wide range of settings. 
The principal angles further confirm the error evaluations (Section \ref{sec:Experimental Results}), showcasing the benefits of TL-PCA.



\section{Conclusion}
In this paper, we propose a new transfer learning approach to PCA. Our TL-PCA has two versions, one that uses a pretrained source model and another that uses the source data. The computation of our TL-PCA is an extended form of the standard PCA, specifically, by eigendecomposition of a matrix the merges the target and source information. 
The TL-PCA performs better than its alternatives of standard PCA. 

We believe that our TL-PCA idea opens up a new research direction of both theoretical and practical aspects. On the theory side, an analytical theory that characterizes the TL-PCA performance is of interest. On the practical side, our TL-PCA procedures, as well as future versions of the general TL-PCA idea, can be proposed for diverse  applications.

\bibliography{transfer_learning_pca_refs}

\begin{thebibliography}{}

\bibitem[Bj{\"o}rck and Golub, 1973]{bjorck1973numerical}
Bj{\"o}rck, A. and Golub, G.~H. (1973).
\newblock Numerical methods for computing angles between linear subspaces.
\newblock {\em Mathematics of Computation}, 27(123):579--594.

\bibitem[Craig et~al., 2024]{craig2024pretraining}
Craig, E., Pilanci, M., Menestrel, T.~L., Narasimhan, B., Rivas, M., Dehghannasiri, R., Salzman, J., Taylor, J., and Tibshirani, R. (2024).
\newblock Pretraining and the lasso.
\newblock {\em arXiv preprint arXiv:2401.12911}.

\bibitem[Dar and Baraniuk, 2022]{dar2020double}
Dar, Y. and Baraniuk, R.~G. (2022).
\newblock Double double descent: On generalization errors in transfer learning between linear regression tasks.
\newblock {\em SIAM Journal on Mathematics of Data Science}, 4(4):1447--1472.

\bibitem[Dar et~al., 2024]{dar2024common}
Dar, Y., LeJeune, D., and Baraniuk, R.~G. (2024).
\newblock The common intuition to transfer learning can win or lose: Case studies for linear regression.
\newblock {\em SIAM Journal on Mathematics of Data Science}.

\bibitem[Dar et~al., 2020]{dar2020subspace}
Dar, Y., Mayer, P., Luzi, L., and Baraniuk, R.~G. (2020).
\newblock Subspace fitting meets regression: The effects of supervision and orthonormality constraints on double descent of generalization errors.
\newblock In {\em International Conference on Machine Learning (ICML)}, pages 2366--2375.

\bibitem[Dhifallah and Lu, 2021]{dhifallah2021phase}
Dhifallah, O. and Lu, Y.~M. (2021).
\newblock Phase transitions in transfer learning for high-dimensional perceptrons.
\newblock {\em Entropy}, 23(4):400.

\bibitem[Edelman et~al., 1998]{edelman1998geometry}
Edelman, A., Arias, T.~A., and Smith, S.~T. (1998).
\newblock The geometry of algorithms with orthogonality constraints.
\newblock {\em SIAM Journal on Matrix Analysis and Applications}, 20(2):303--353.

\bibitem[Gerace et~al., 2022]{gerace2021probing}
Gerace, F., Saglietti, L., Mannelli, S.~S., Saxe, A., and Zdeborov{\'a}, L. (2022).
\newblock Probing transfer learning with a model of synthetic correlated datasets.
\newblock {\em Mach. Learn.: Sci. Technol.}, 3(1):015030.

\bibitem[Hamm and Lee, 2008]{grassmann2008hamm}
Hamm, J. and Lee, D.~D. (2008).
\newblock Grassmann discriminant analysis: a unifying view on subspace-based learning.
\newblock In {\em International Conference on Machine Learning}, page 376–383.

\bibitem[Krizhevsky and Hinton, 2009]{krizhevsky2009learning}
Krizhevsky, A. and Hinton, G. (2009).
\newblock Learning multiple layers of features from tiny images.

\bibitem[Lake et~al., 2015]{lake2015human}
Lake, B.~M., Salakhutdinov, R., and Tenenbaum, J.~B. (2015).
\newblock Human-level concept learning through probabilistic program induction.
\newblock {\em Science}, 350(6266):1332--1338.

\bibitem[Le and Yang, 2015]{le2015tiny}
Le, Y. and Yang, X. (2015).
\newblock Tiny imagenet visual recognition challenge.
\newblock {\em CS 231N}, 7(7):3.

\bibitem[Lecun et~al., 1998]{726791}
Lecun, Y., Bottou, L., Bengio, Y., and Haffner, P. (1998).
\newblock Gradient-based learning applied to document recognition.
\newblock {\em Proceedings of the IEEE}, 86(11):2278--2324.

\bibitem[Liu et~al., 2015]{liu2015faceattributes}
Liu, Z., Luo, P., Wang, X., and Tang, X. (2015).
\newblock Deep learning face attributes in the wild.
\newblock In {\em Proceedings of International Conference on Computer Vision (ICCV)}.

\bibitem[Netzer et~al., 2011]{37648}
Netzer, Y., Wang, T., Coates, A., Bissacco, A., Wu, B., and Ng, A.~Y. (2011).
\newblock Reading digits in natural images with unsupervised feature learning.
\newblock In {\em {NIPS} Workshop on Deep Learning and Unsupervised Feature Learning 2011}.

\bibitem[Obst et~al., 2021]{obst2021transfer}
Obst, D., Ghattas, B., Cugliari, J., Oppenheim, G., Claudel, S., and Goude, Y. (2021).
\newblock Transfer learning for linear regression: a statistical test of gain.
\newblock {\em arXiv preprint arXiv:2102.09504}.

\bibitem[Pan and Yang, 2009]{pan2009survey}
Pan, S.~J. and Yang, Q. (2009).
\newblock A survey on transfer learning.
\newblock {\em IEEE transactions on knowledge and data engineering}, 22(10):1345--1359.

\bibitem[Panagopoulos et~al., 2016]{panagopoulos2016constrained}
Panagopoulos, O.~P., Pappu, V., Xanthopoulos, P., and Pardalos, P.~M. (2016).
\newblock Constrained subspace classifier for high dimensional datasets.
\newblock {\em Omega}, 59:40--46.
\newblock Business Analytics.

\bibitem[Song et~al., 2024]{song2024generalization}
Song, Y., Bhattacharya, S., and Sur, P. (2024).
\newblock Generalization error of min-norm interpolators in transfer learning.
\newblock {\em arXiv preprint arXiv:2406.13944}.

\end{thebibliography}

\appendix
\renewcommand\thefigure{\thesection.\arabic{figure}}    
\setcounter{figure}{0}  

\renewcommand\thetable{\thesection.\arabic{table}}    
\setcounter{table}{0} 

\counterwithin*{figure}{section}
\counterwithin*{table}{section}

\section*{Appendices}

\section{Proof of TL-PCA-P Maximization Form} \label{proof_of_target_max_problem}
The optimization problem for the target task is formulated as follows:
\begin{align}
\label{appendix:proof-TL-PCA-P Maximization Form - 1}
&\mtx{U}_k = \nonumber\\\nonumber
&= \underset{\substack{\mtx{W} \in \mathbb{R}^{d \times k}\\\mtx{W}^T \mtx{W} = \mtx{I}_k}}{\arg\min} \left\lbrace \frac{1}{n} \Frobnormsquared{ (\mtx{I}_d - \mtx{W}\mtx{W}^T) \mtx{X} } + \frac{\alpha}{2} h(\mtx{W};\widetilde{\mtx{U}}_m) \right\rbrace \\\nonumber
&= \underset{\substack{\mtx{W} \in \mathbb{R}^{d \times k}\\\mtx{W}^T \mtx{W} = \mtx{I}_k}}{\arg\max} \Bigg\lbrace \frac{1}{n} \mtxtrace{\mtx{W}^T \mtx{X} \mtx{X}^T \mtx{W} } \\
&\qquad\qquad\qquad - \frac{\alpha}{2} \| \mtx{W} \mtx{W}^T - \widetilde{\mtx{U}}_m\widetilde{\mtx{U}}_m^T \|_F^2 \Bigg\rbrace 
\end{align}
Here, we used the relation $\Frobnormsquared{ (\mtx{I}_d - \mtx{W}\mtx{W}^T) \mtx{X} } = \Frobnormsquared{\mtx{X}} - \mtxtrace{ \mtx{W}^T \mtx{X} \mtx{X}^T \mtx{W} }$ and the defintion of the penalty $ h(\mtx{W};\widetilde{\mtx{U}}_m)$ from (\ref{eq:TL-PCA-P - penalty function - projection F norm}).
Next, we develop the second term of (\ref{appendix:proof-TL-PCA-P Maximization Form - 1}) as follows:
\begin{equation}
\begin{aligned}
&\| \mtx{W} \mtx{W}^T - \widetilde{\mtx{U}}_m\widetilde{\mtx{U}}_m^T \|_F^2 = \\
&=\text{Tr}\{ (\mtx{W} \mtx{W}^T - 
\widetilde{\mtx{U}}_m \widetilde{\mtx{U}}_m^T)^T (\mtx{W} \mtx{W}^T - \widetilde{\mtx{U}}_m \widetilde{\mtx{U}}_m^T) \} \\
&= \| \mtx{W} \|_F^2 + \| \widetilde{\mtx{U}}_m \|_F^2 - 2 \| \mtx{W}^T \widetilde{\mtx{U}}_m \|_F^2 \\
&= \mtxtrace{ \mtx{W} \mtx{W}^T } + \mtxtrace{ \widetilde{\mtx{U}}_m \widetilde{\mtx{U}}_m^T } - 2 \mtxtrace{ \mtx{W}^T \widetilde{\mtx{U}}_m \widetilde{\mtx{U}}_m^T \mtx{W} } \\
&= k + m - 2\mtxtrace{ \mtx{W}^T \widetilde{\mtx{U}}_m \widetilde{\mtx{U}}_m^T \mtx{W} }
\label{appendix:proof-TL-PCA-P Maximization Form - 2}
\end{aligned}
\end{equation}
The last equality is due to the $k$ orthonormal columns of $\mtx{W}$ and $m$ orthonormal columns of $\widetilde{\mtx{U}}_m$.

Substituting the expression developed in (\ref{appendix:proof-TL-PCA-P Maximization Form - 2}) into the maximization problem (\ref{appendix:proof-TL-PCA-P Maximization Form - 1}), and joining the traces, gives the maximization form of (\ref{eq:TL-PCA-P - optimization - maximization problem}) and completes the proof.

\section{Proof of TL-PCA-D Maximization Form} 
\label{appendix:sec:Proof of TL-PCA-D Maximization Form}
The optimization problem for TL-PCA-D is formulated as follows:
\begin{align}
&\mtx{U}_k = \\\nonumber
&= \underset{\substack{\mtx{W} \in \mathbb{R}^{d \times k}\\\mtx{W}^T \mtx{W} = \mtx{I}_k}}{\arg\min} \left\lbrace  \frac{1}{n} \| (\mtx{I}_d - \mtx{W} \mtx{W}^T) \mtx{X} \|_F^2 + \alpha g(\mtx{W}; \widetilde{\mtx{X}}) \right\rbrace \\\nonumber
&=  \underset{\substack{\mtx{W} \in \mathbb{R}^{d \times k}\\\mtx{W}^T \mtx{W} = \mtx{I}_k}}{\arg\min} \Bigg\lbrace  \frac{1}{n} \| (\mtx{I}_d - \mtx{W} \mtx{W}^T) \mtx{X} \|_F^2  \\\nonumber
&\qquad\qquad\qquad+ \frac{\alpha}{\widetilde{n}} \| (\mtx{I}_d - \mtx{W} \mtx{W}^T) \widetilde{\mtx{X}} \|_F^2 \Bigg\rbrace \\\nonumber
&=  \underset{\substack{\mtx{W} \in \mathbb{R}^{d \times k}\\\mtx{W}^T \mtx{W} = \mtx{I}_k}}{\arg\min} \Bigg\lbrace   \frac{1}{n} \left( \| \mtx{X} \|_F^2 - \mtxtrace{\mtx{X}^T \mtx{W} \mtx{W}^T \mtx{X}} \right)  \\\nonumber
&\qquad\qquad\qquad+ \frac{\alpha}{\widetilde{n}}\left( \| \widetilde{\mtx{X}} \|_F^2 - \mtxtrace{\widetilde{\mtx{X}}^T \mtx{W} \mtx{W}^T \widetilde{\mtx{X}}} \right) \Bigg\rbrace
\end{align}




At this point, since \( \| \mtx{X} \|_F^2 \) and \( \| \widetilde{\mtx{X}} \|_F^2 \) are constants with respect to \( \mtx{W} \), they can be dropped from the minimization. Thus, using the cyclic property of the trace, the optimization problem can be written as 
\begin{align}
&\mtx{U}_k =  \underset{\substack{\mtx{W} \in \mathbb{R}^{d \times k}\\\mtx{W}^T \mtx{W} = \mtx{I}_k}}{\arg\max}  \Bigg\lbrace \frac{1}{n} \mtxtrace{\mtx{W}^T \mtx{X} \mtx{X}^T \mtx{W}} \\\nonumber
&\qquad\qquad\qquad +  \frac{\alpha}{\widetilde{n}} \mtxtrace{\mtx{W}^T \widetilde{\mtx{X}} \widetilde{\mtx{X}}^T \mtx{W}} \Bigg\rbrace
\end{align}
By joining the traces in the last expression we get the maximization formulation of (\ref{eq:TL-PCA-D - optimization - maximization problem}) and complete the proof. 

\section{Additional Experimental Details}
\label{appendix:sec:Additional Experimental Details}

\subsection{Datasets}
\label{appendix:subsec:Additional Experimental Details - Datasets}
The datasets used in this work are publicly available and were originally published by the following creators: 
\begin{enumerate}
    \item \textbf{CIFAR-10:} \cite{krizhevsky2009learning}
    \item \textbf{SVHN:}
    \cite{37648}
    \item \textbf{Omniglot:} 
    \cite{lake2015human}
    \item \textbf{MNIST:}
    \cite{726791}
    \item \textbf{Tiny-ImageNet:}
    \cite{le2015tiny}
    \item \textbf{CelebA:}
    \cite{liu2015faceattributes}
\end{enumerate}

\subsection{Computing Resources}
\label{appendix:subsec:Additional Experimental Details - Computing Resources}
We run our experiments on CPUs available to us in an internal cluster.

\section{Additional Experimental Results}
\label{appendix:sec:Additional Experimental Results}
\begin{figure}
    \centering
    \subfloat[Train error]{\includegraphics[width=0.48\linewidth]{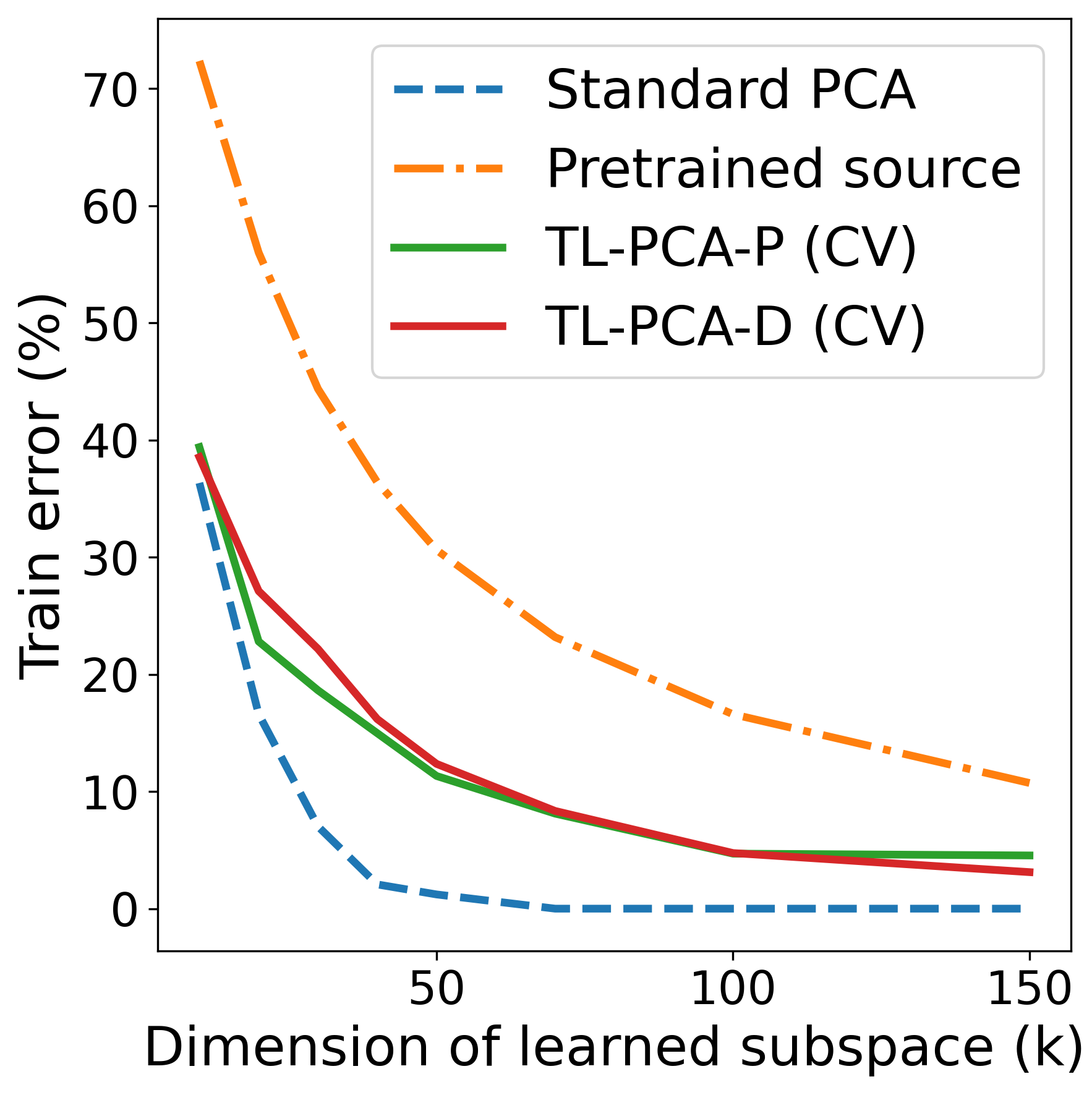}}    
    \subfloat[Test error]{\includegraphics[width=0.48\linewidth]{MNIST_results/comparison_test.png}}    
    \caption{Train and test error comparison of PCA and TL-PCA for Omniglot to MNIST. }
    \label{appendix:fig:error_graphs_mnist_train_test_errors}
\end{figure}

\begin{figure}
    \centering
    \subfloat[Train error]{\includegraphics[width=0.48\linewidth]{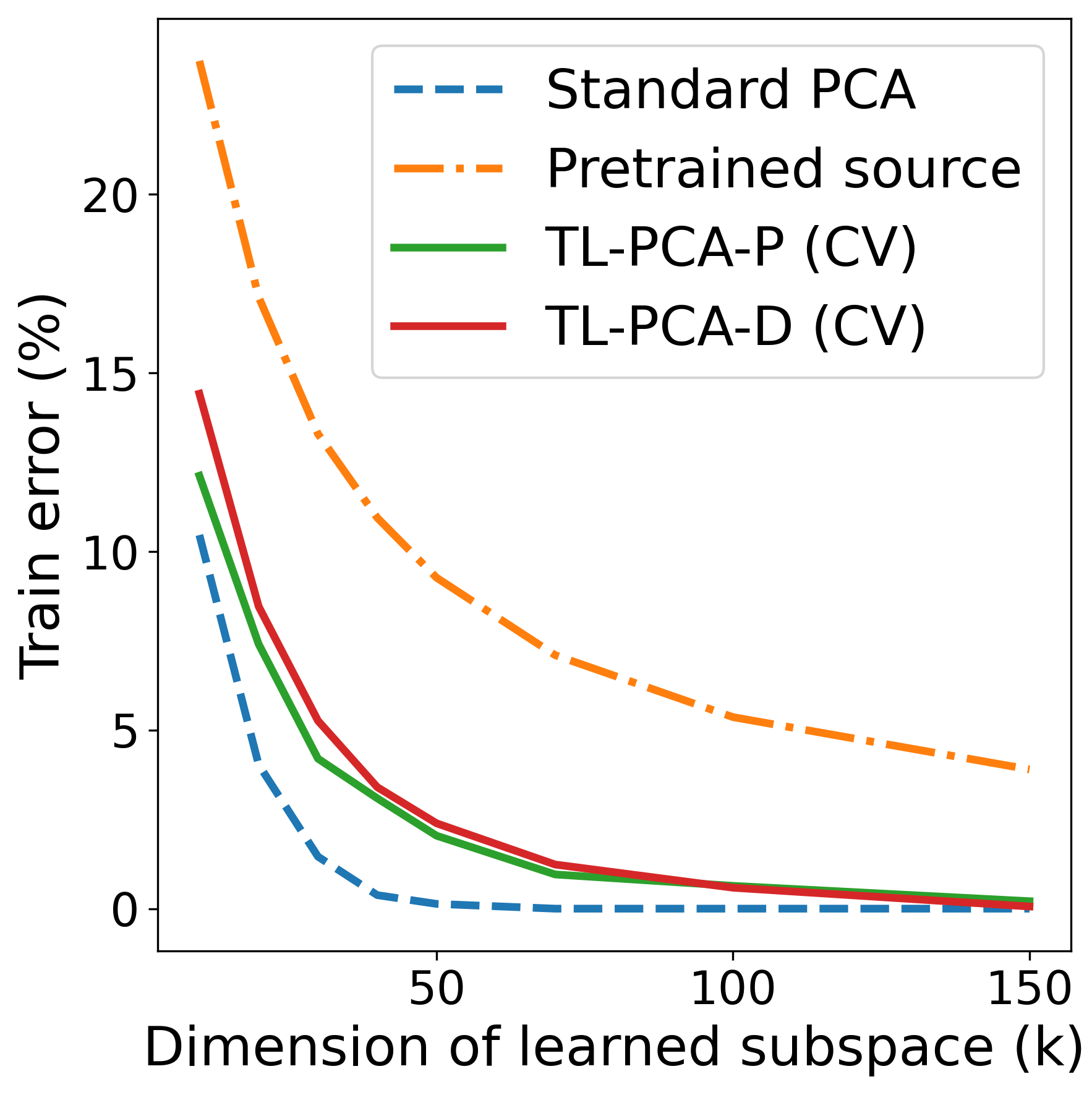}}    
    \subfloat[Test error]{\includegraphics[width=0.48\linewidth]{SVHN_results/comparison_test.png}}    
    \caption{Train and test error comparison of PCA and TL-PCA for CIFAR-10 to SVHN. }
    \label{appendix:fig:error_graphs_svhn_train_test_errors}
\end{figure}

\begin{figure}
    \centering
    \subfloat[Train error]{\includegraphics[width=0.48\linewidth]{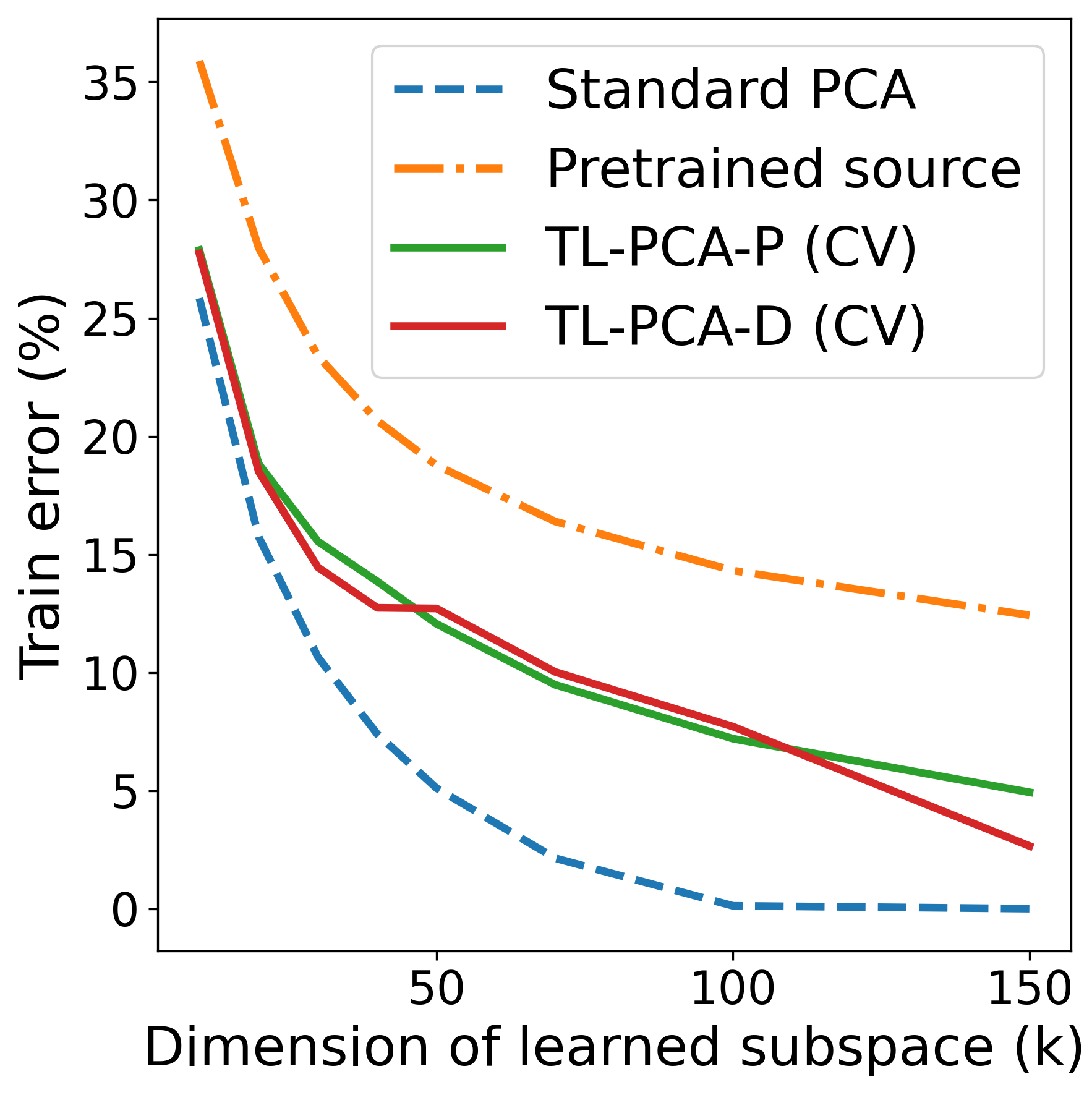}}    
    \subfloat[Test error]{\includegraphics[width=0.48\linewidth]{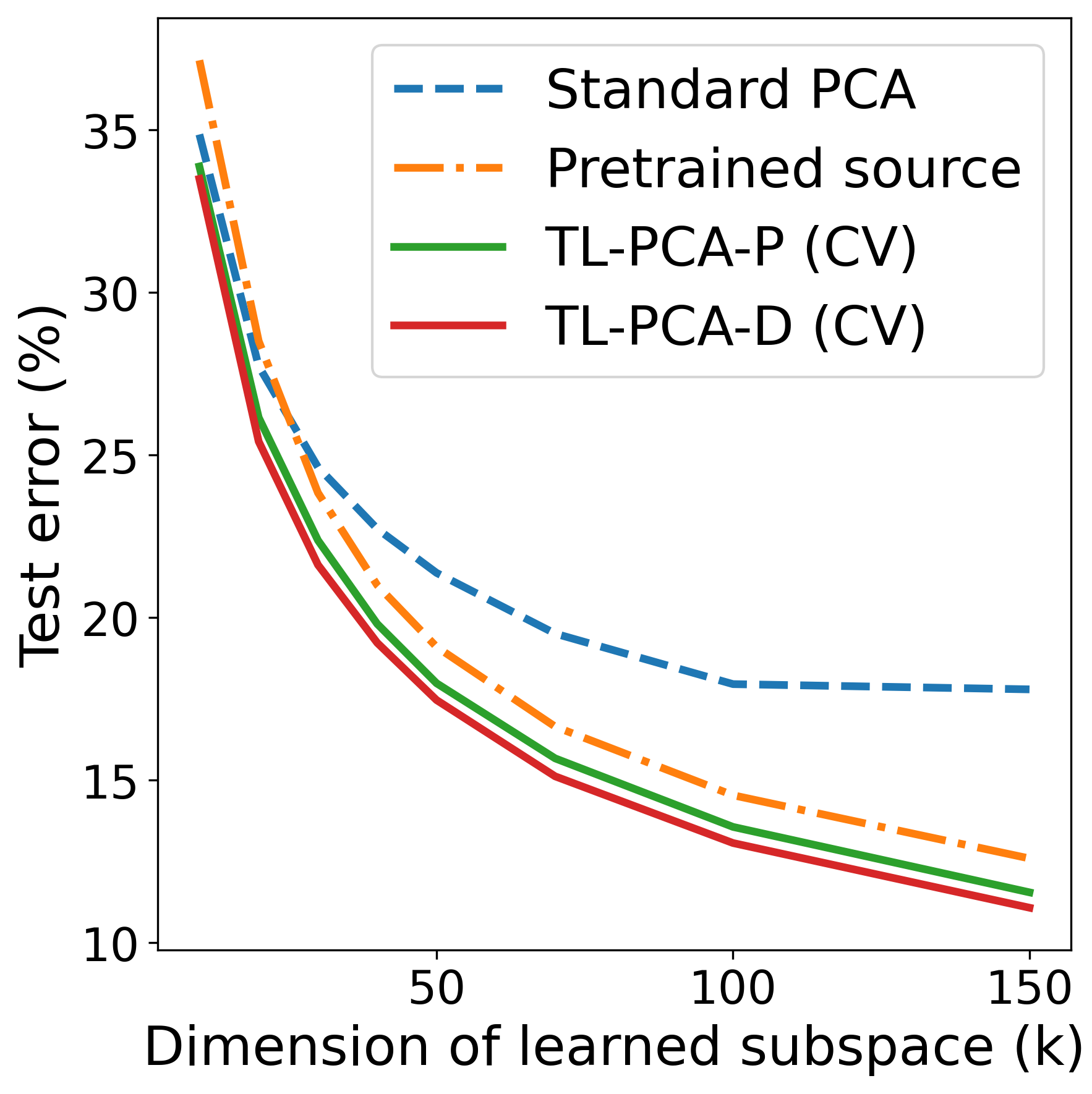}}    
    \caption{Train and test error comparison of PCA and TL-PCA for Tiny ImageNet to CelebA. }
    \label{appendix:fig:error_graphs_celeba_train_test_errors}
\end{figure}

We provide here experimental results in addition to those shown in the main paper. The full errors of CIFAR10-to-SVHN are reported in Table \ref{app:table_svhn} that extends Table \ref{table_svhn} from the main text, due to lack of space. 
Additional error graphs are shown in Figures \ref{appendix:fig:error_graphs_mnist_train_test_errors}, \ref{appendix:fig:error_graphs_svhn_train_test_errors}, \ref{appendix:fig:error_graphs_celeba_train_test_errors} for TL-PCA with cross validation (CV), as in the main paper Figure \ref{fig:error_graphs_main_paper}. 

\begin{figure*}
    \centering
    \subfloat[]{\includegraphics[width=0.33\linewidth]{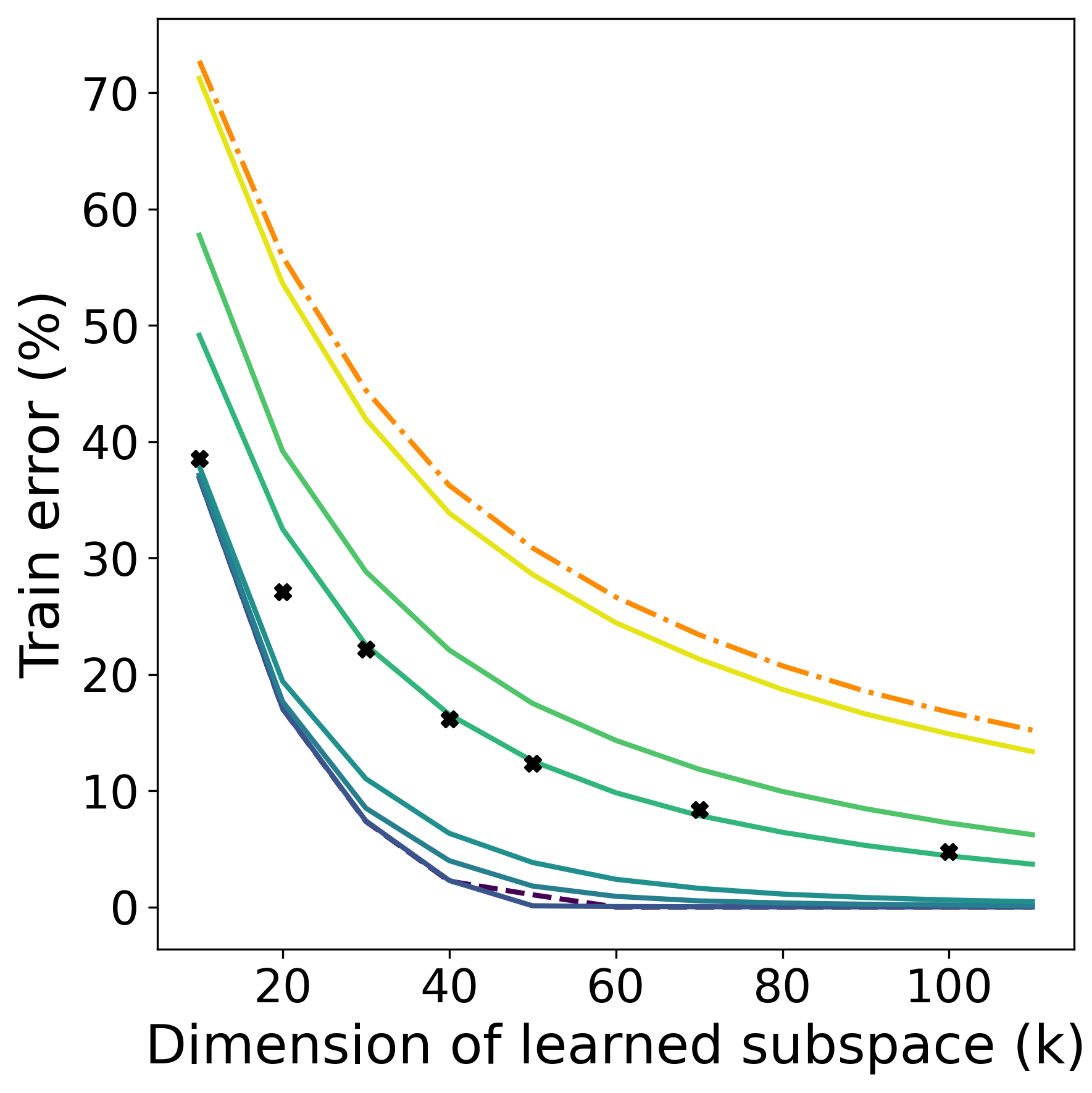}}
    \subfloat[]{\includegraphics[width=0.33\linewidth]{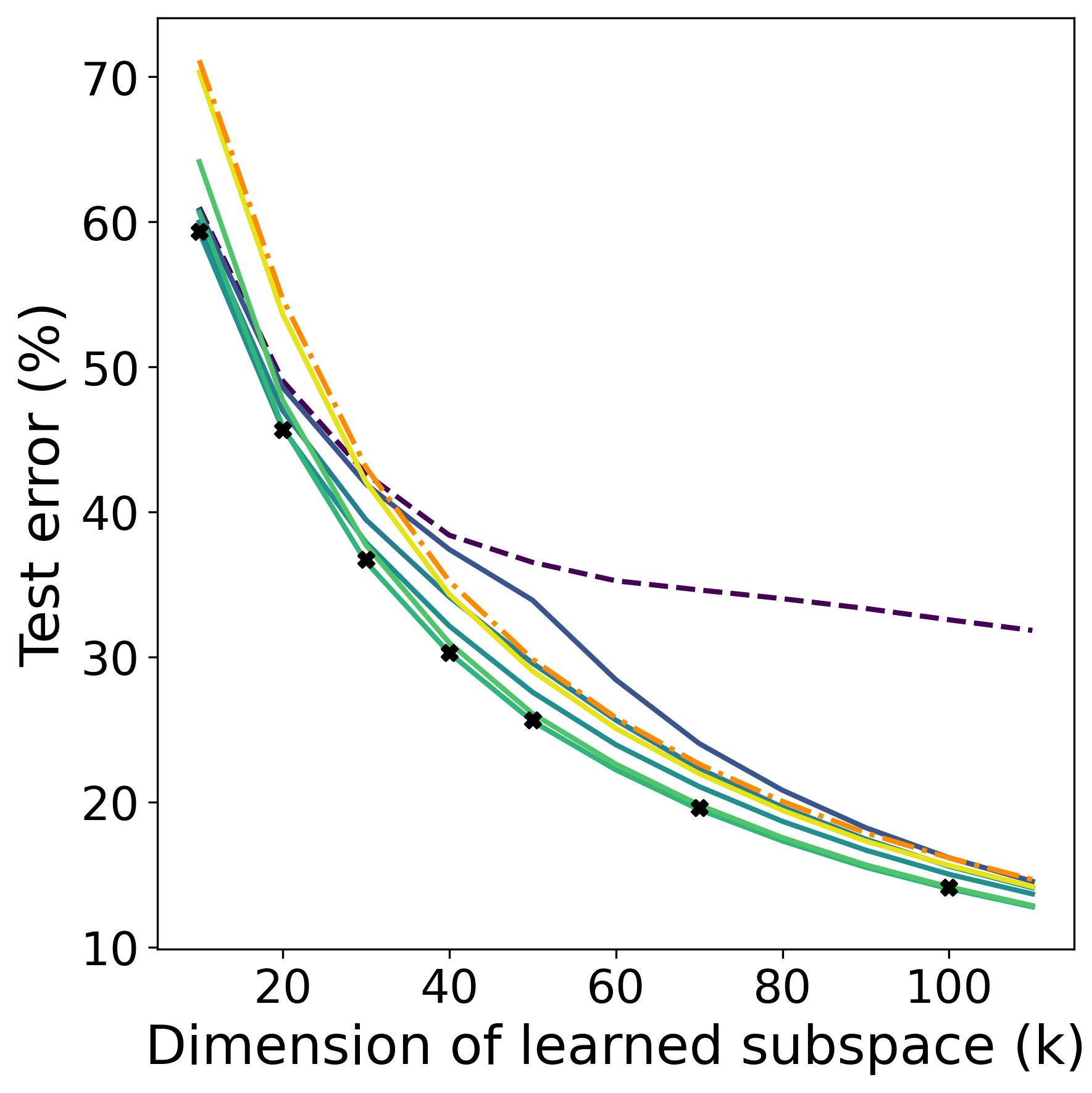}}
    \subfloat{\includegraphics[width=0.18\linewidth]{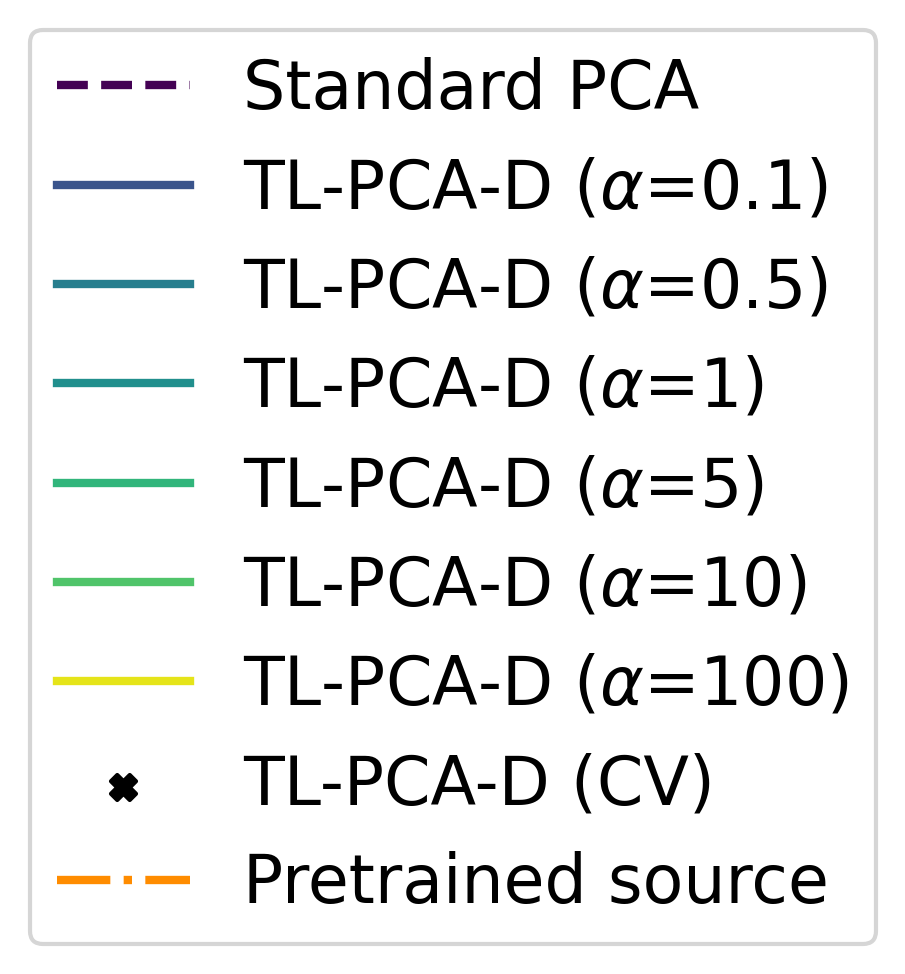}}
    \caption{TL-PCA-D train and test errors as a function of k for different values of $\alpha$ (Omniglot to MNIST). }
    \label{appendix:fig:tlpcad_train_test_errors_mnist}
\end{figure*}

\begin{figure*}
    \centering
    \subfloat[]{\includegraphics[width=0.33\linewidth]{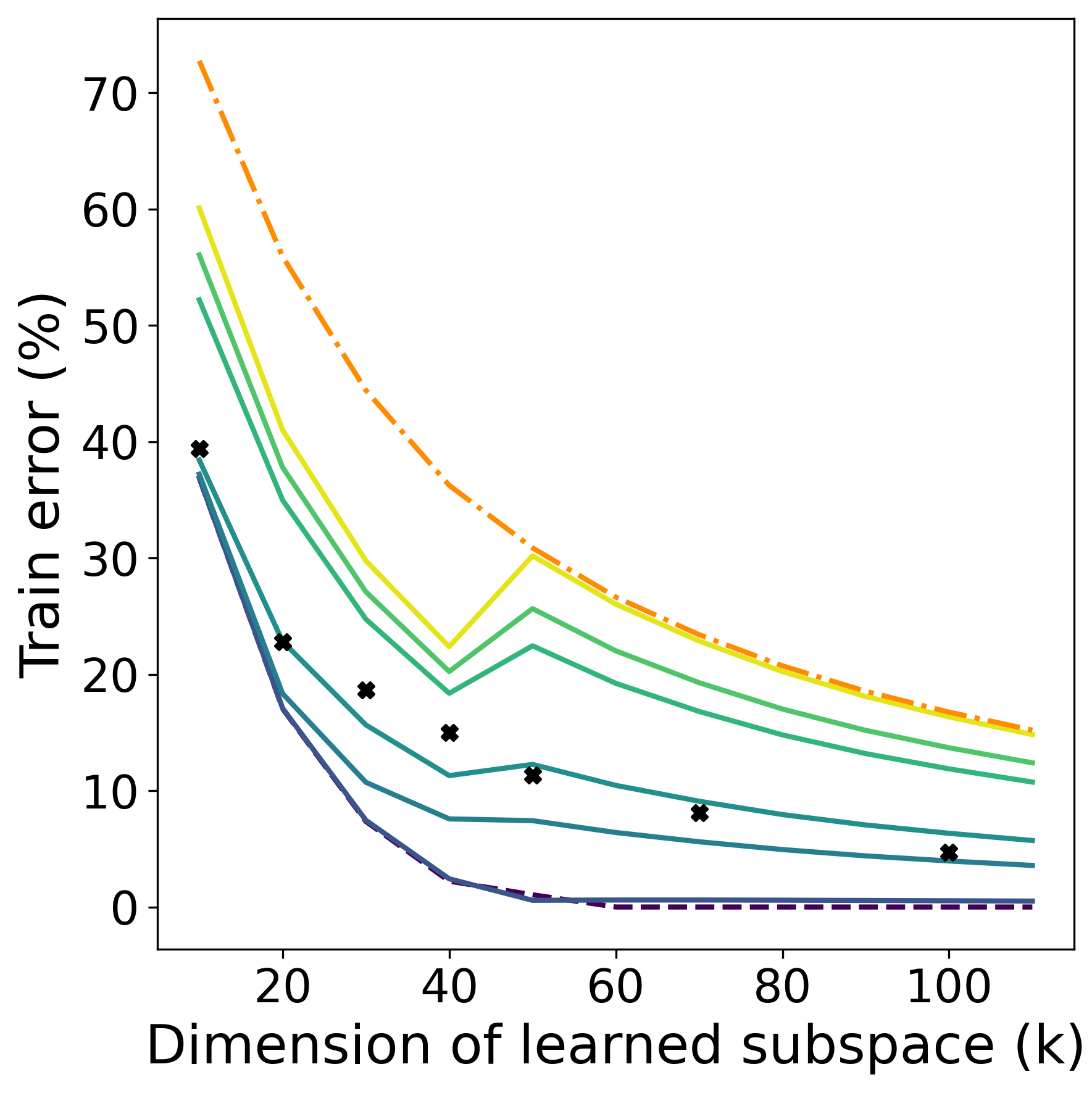}}
    \subfloat[]{\includegraphics[width=0.33\linewidth]{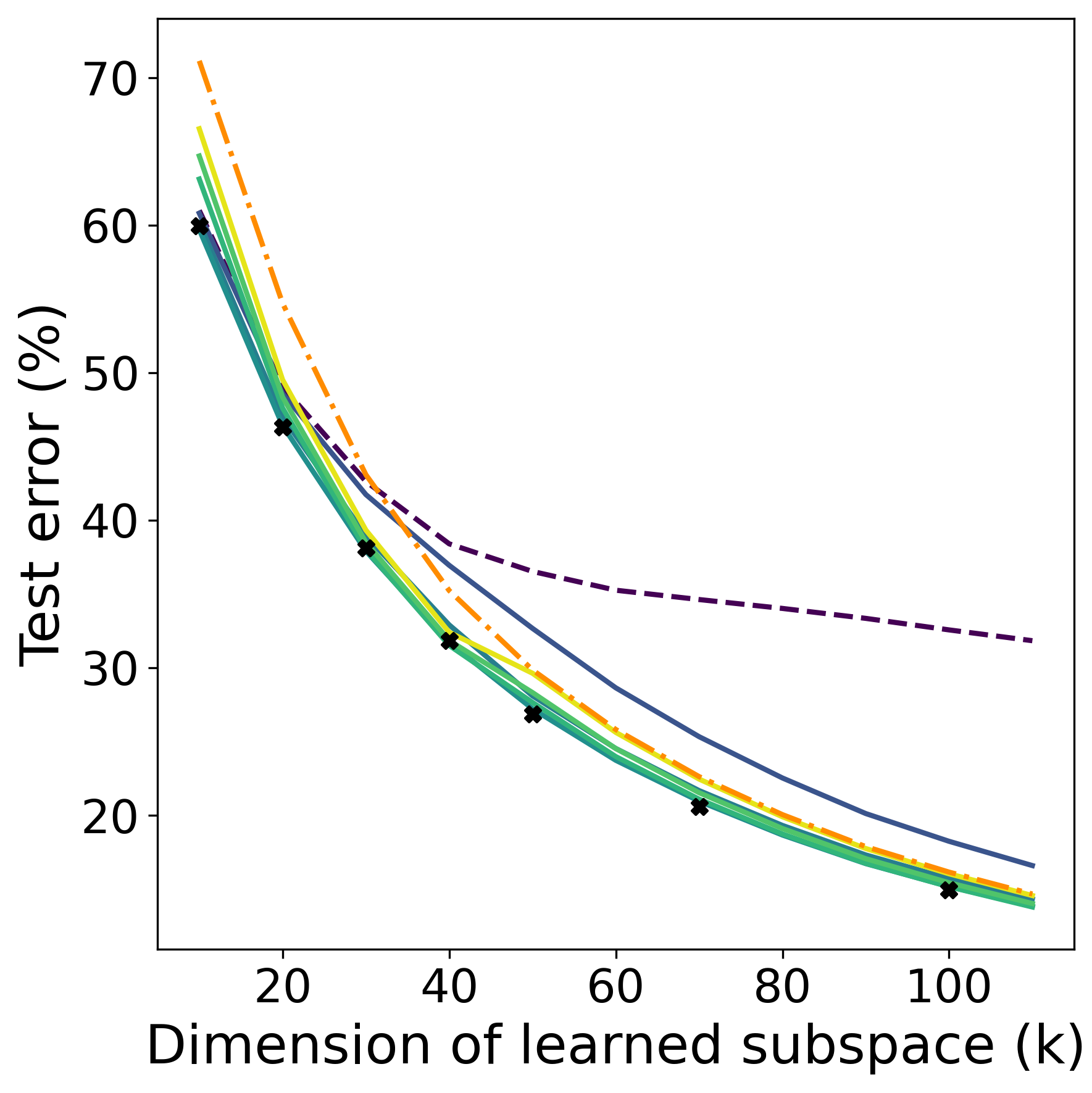}}
    \subfloat{\includegraphics[width=0.18\linewidth]{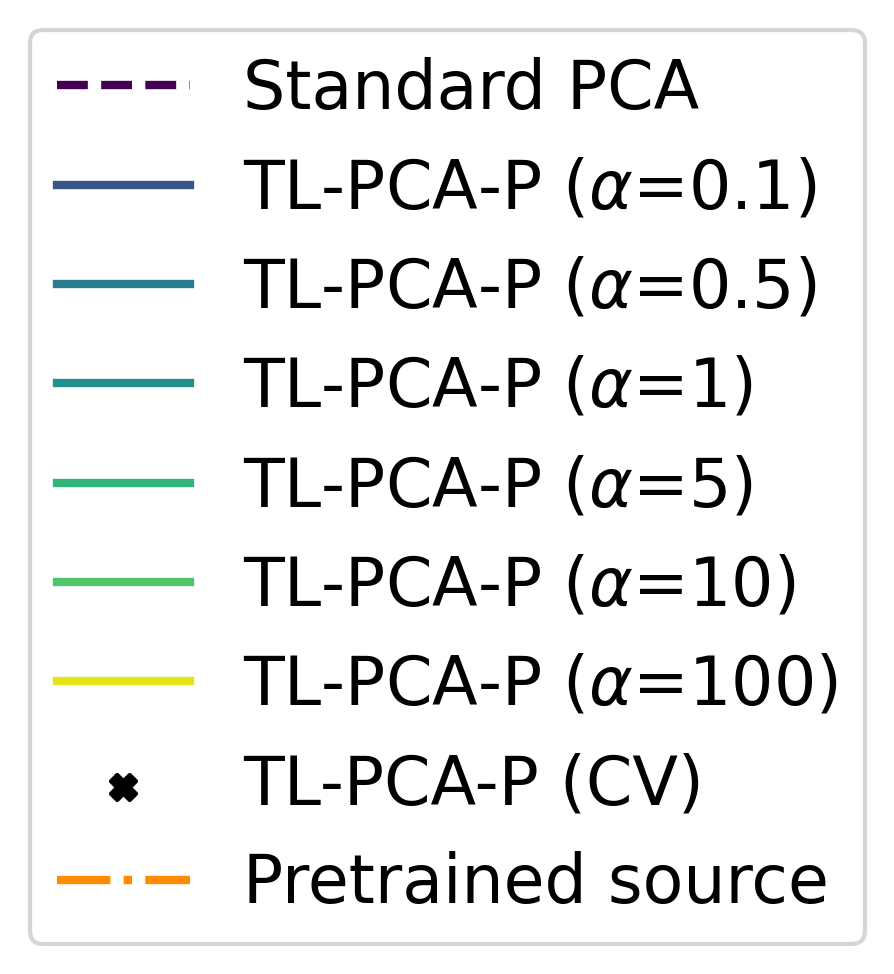}}
    \caption{TL-PCA-P train and test errors as a function of k for various values of $\alpha$, with $m=\left\lceil 0.8\cdot k \right\rceil$ principal directions transferred from the pretrained source (Omniglot to MNIST). }
    \label{appendix:fig:tlpcap_train_test_errors_mnist_08}
\end{figure*}

\begin{figure*}
    \centering
    \subfloat[]{\includegraphics[width=0.33\linewidth]{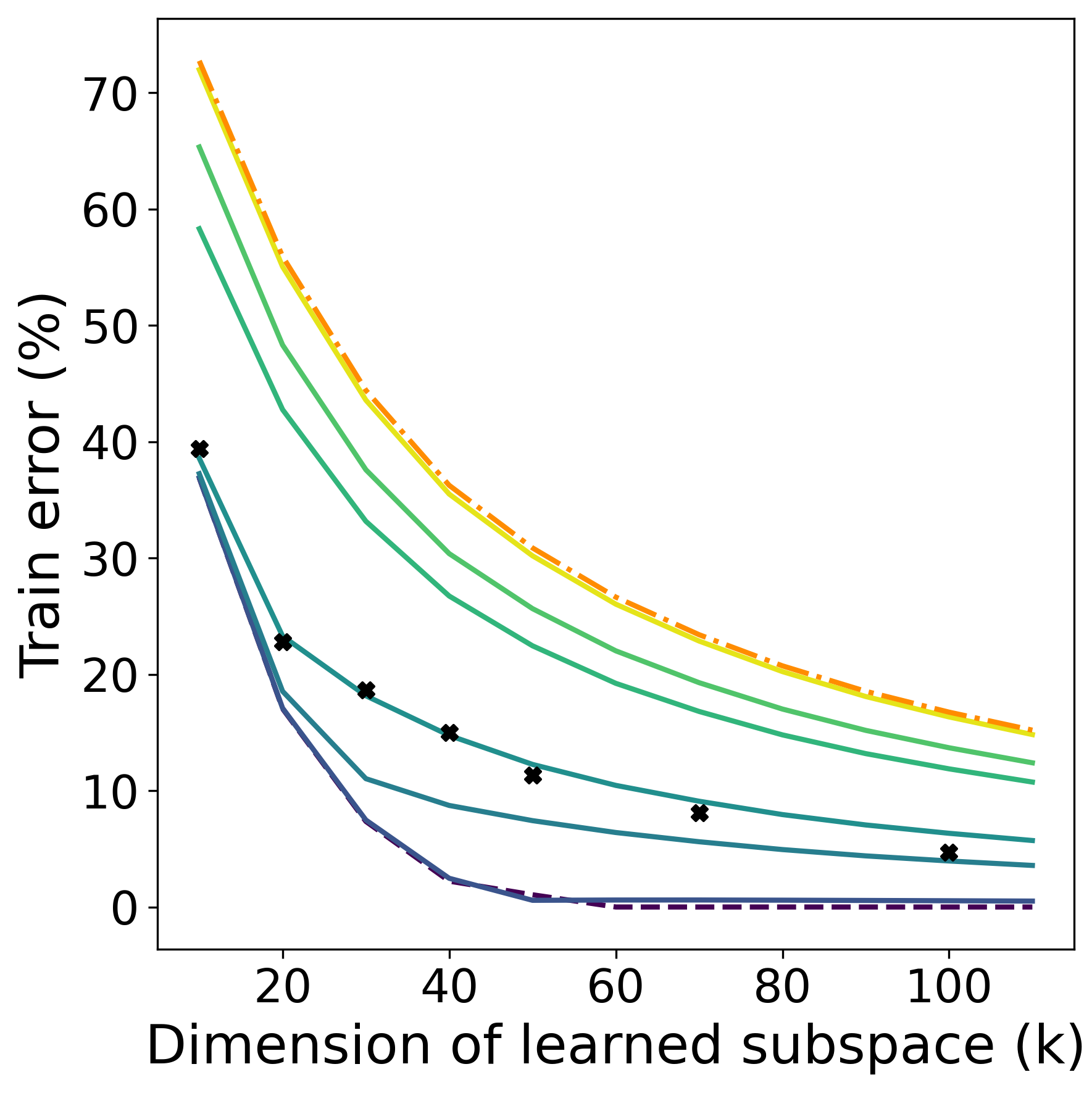}}
    \subfloat[]{\includegraphics[width=0.33\linewidth]{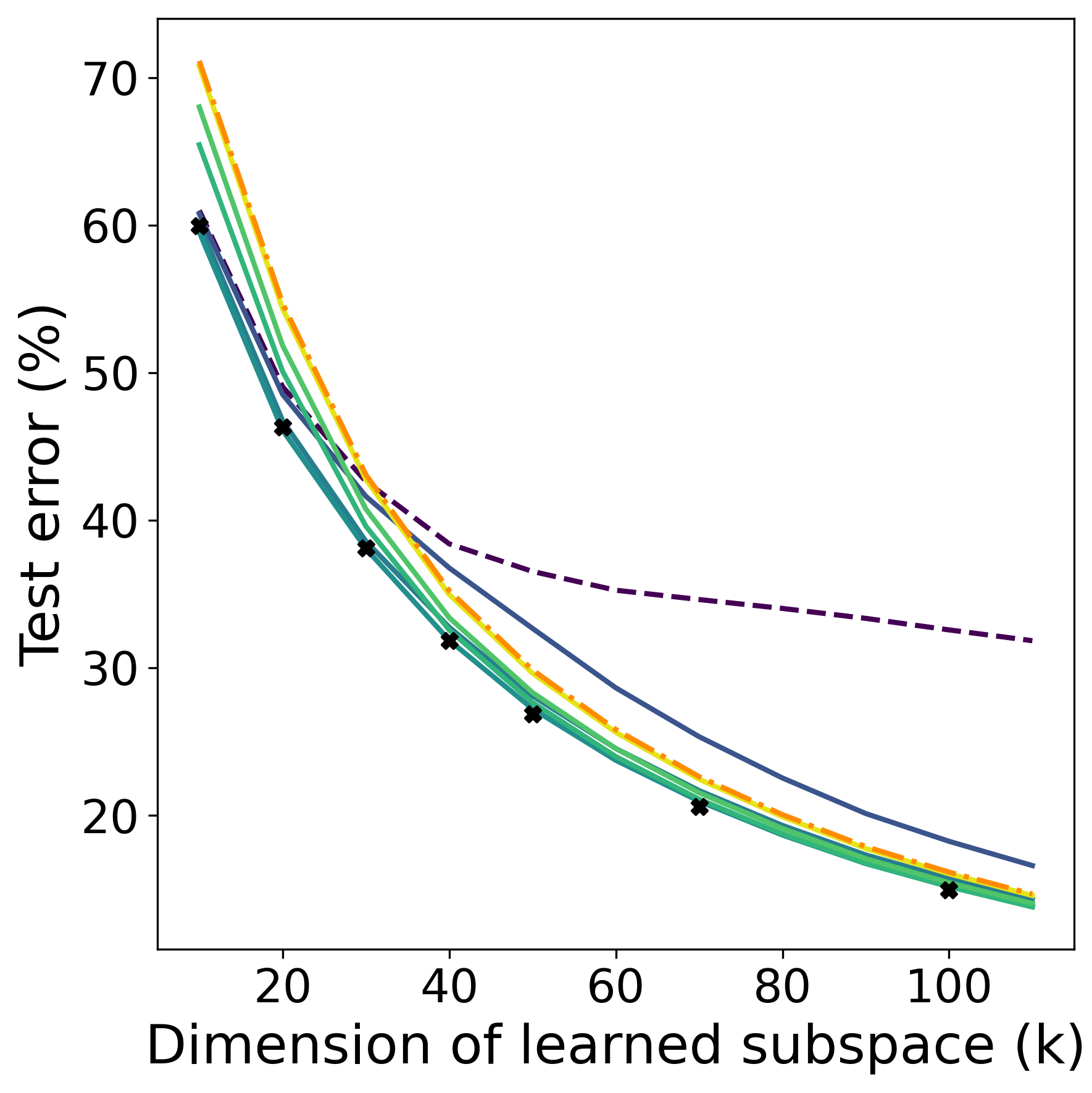}}
    \subfloat{\includegraphics[width=0.18\linewidth]{MNIST_results/alpha_values_legend.png}}
    \caption{TL-PCA-P train and test errors as a function of k for various values of $\alpha$, with $m=k$ principal directions transferred from the pretrained source (Omniglot to MNIST). }
    \label{appendix:fig:tlpcap_train_test_errors_mnist_1}
\end{figure*}

\begin{figure*}
    \centering
    \subfloat[]{\includegraphics[width=0.33\linewidth]{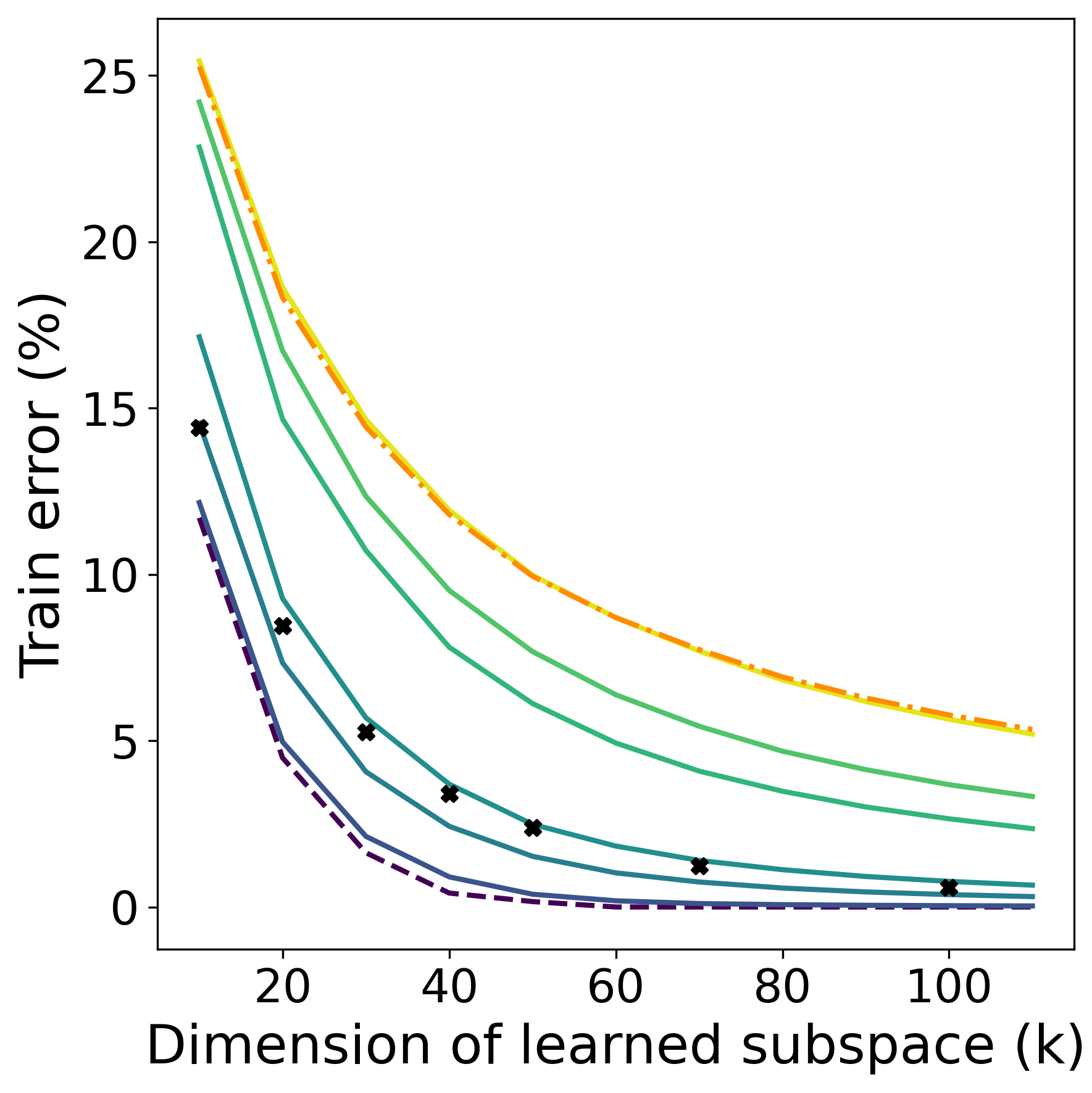}}
    \subfloat[]{\includegraphics[width=0.33\linewidth]{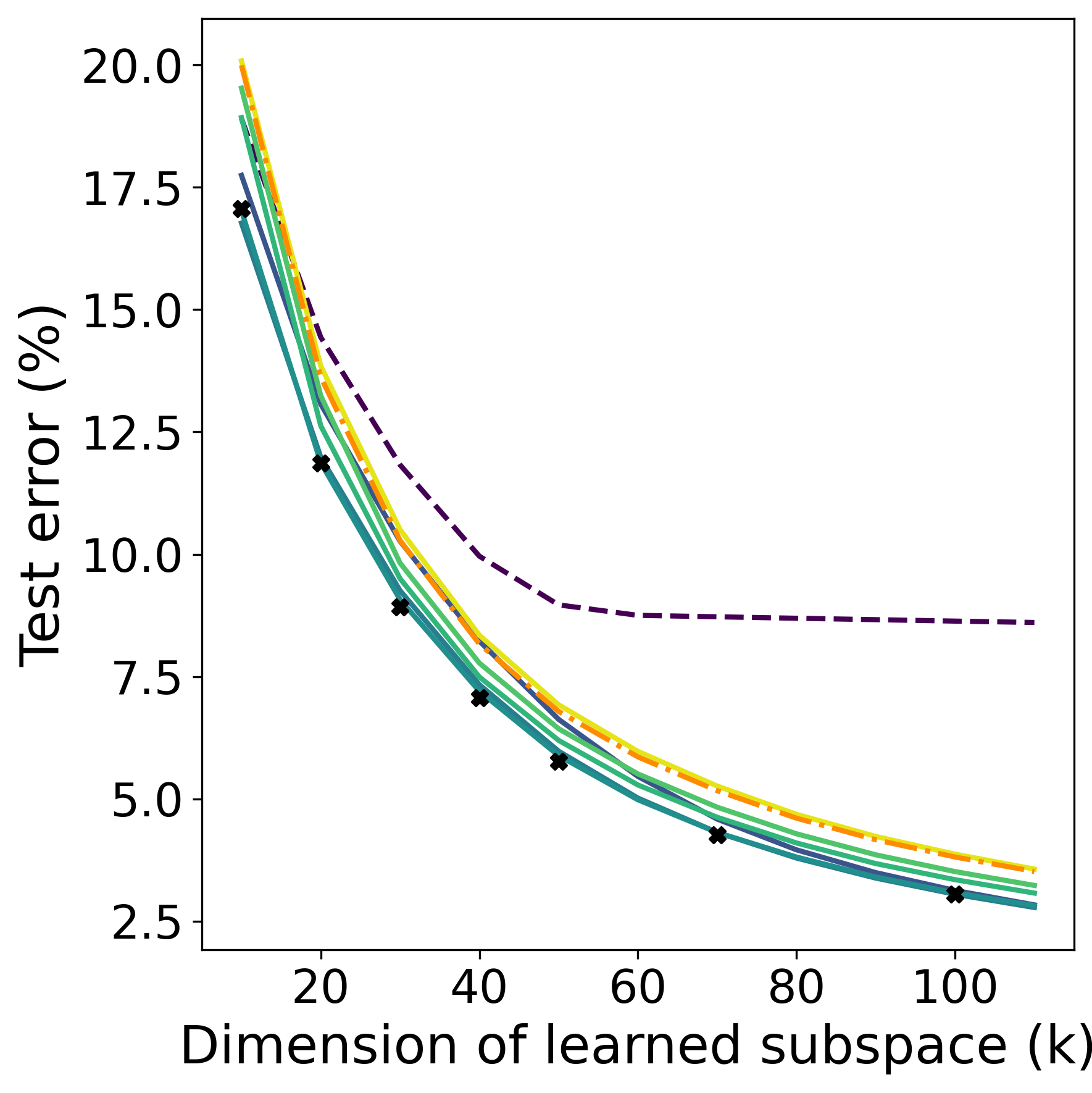}}
    \subfloat{\includegraphics[width=0.18\linewidth]{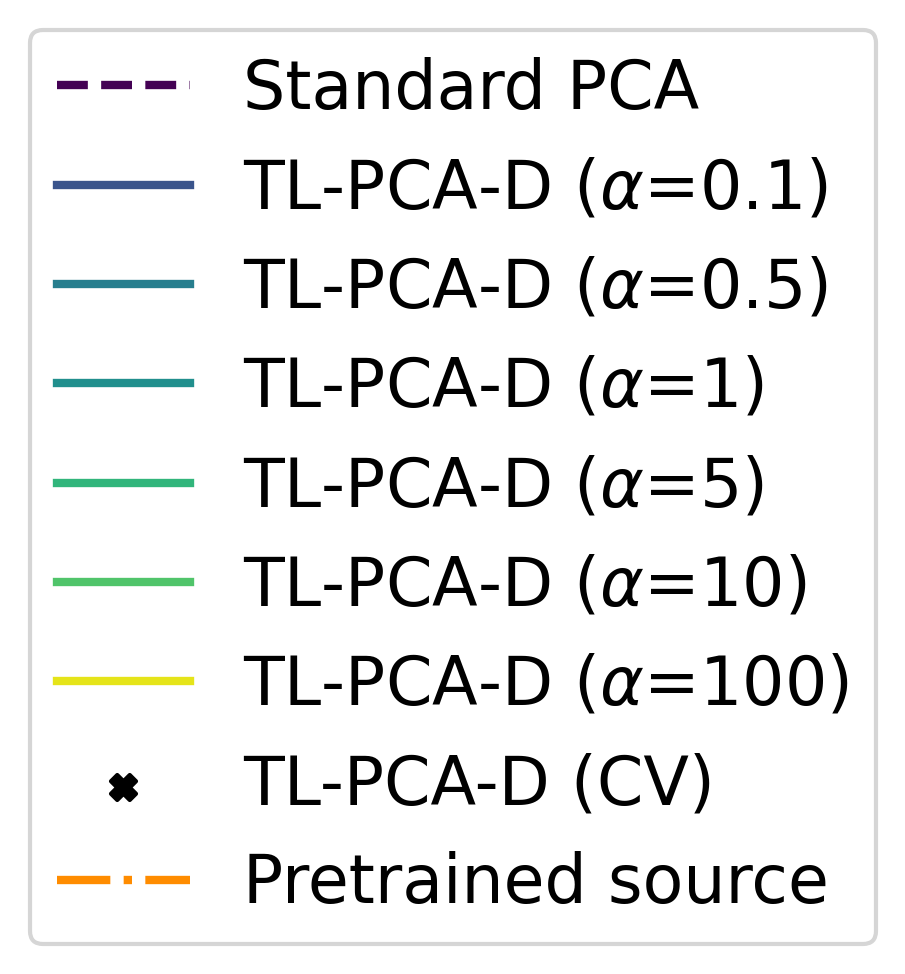}}
    \caption{TL-PCA-D train and test errors as a function of k for different values of $\alpha$ (CIFAR-10 to SVHN). }
    \label{appendix:fig:tlpcad_train_test_errors_svhn}
\end{figure*}

\begin{figure*}
    \centering
    \subfloat[]{\includegraphics[width=0.33\linewidth]{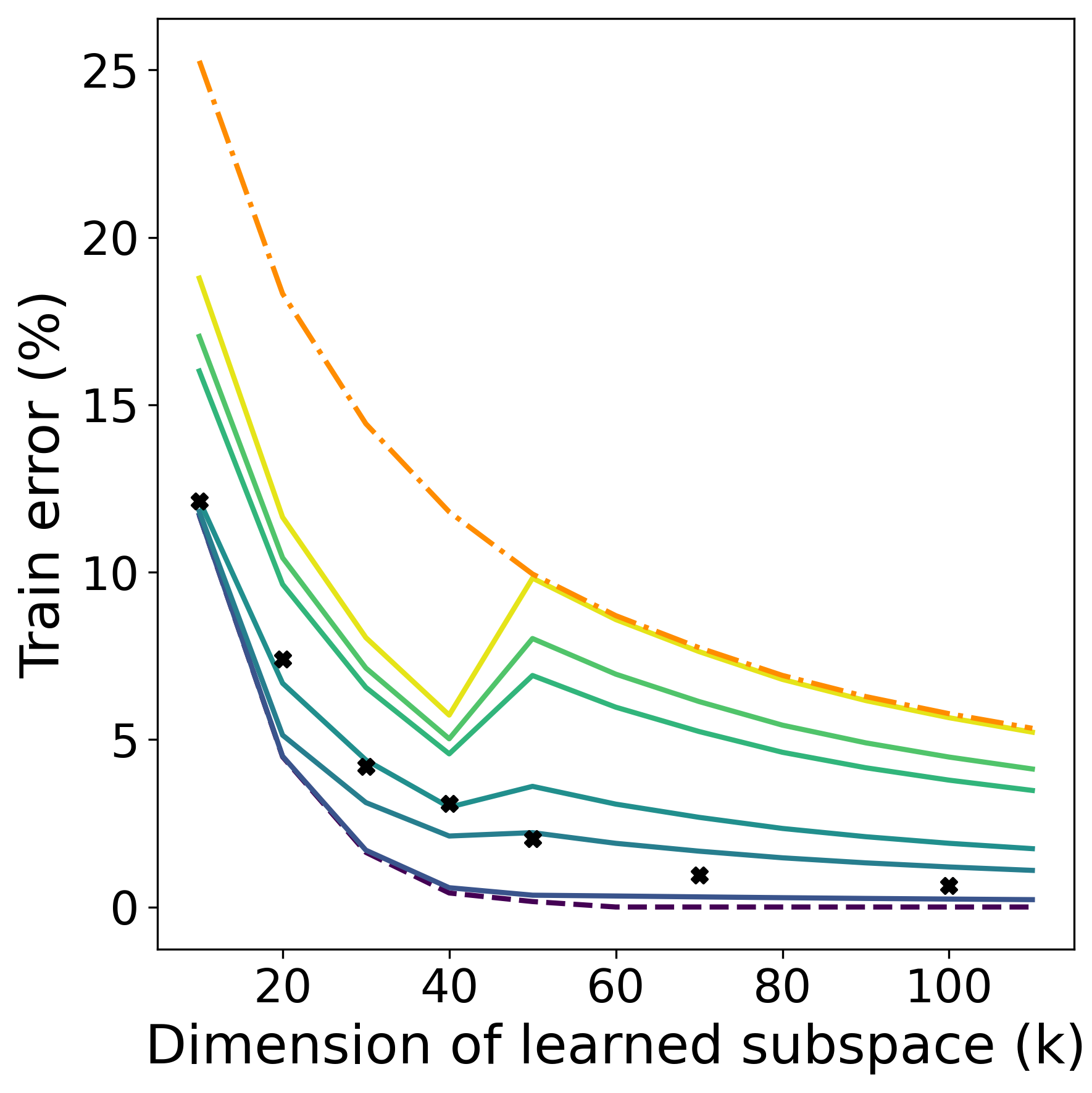}}
    \subfloat[]{\includegraphics[width=0.33\linewidth]{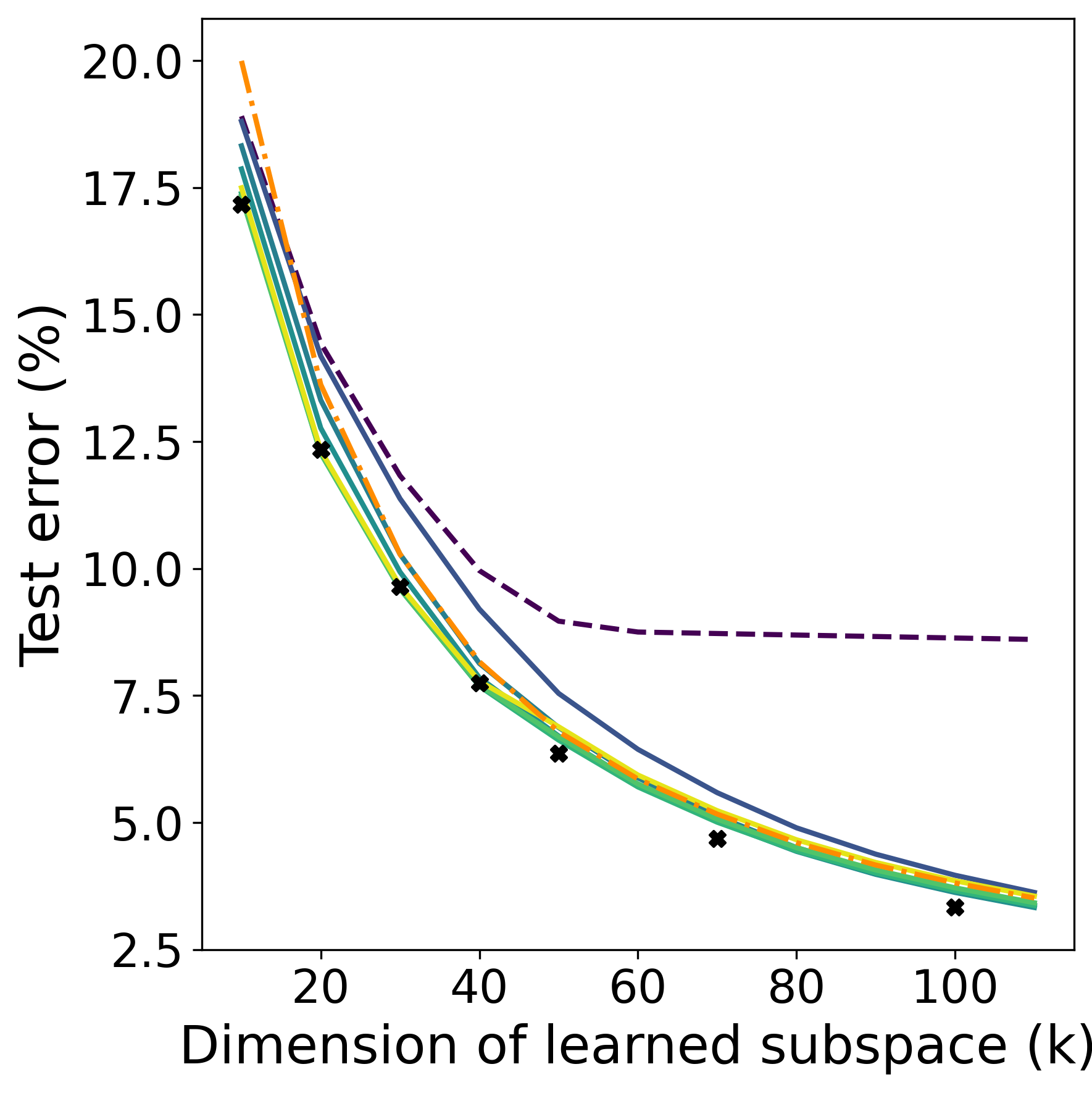}}
    \subfloat{\includegraphics[width=0.18\linewidth]{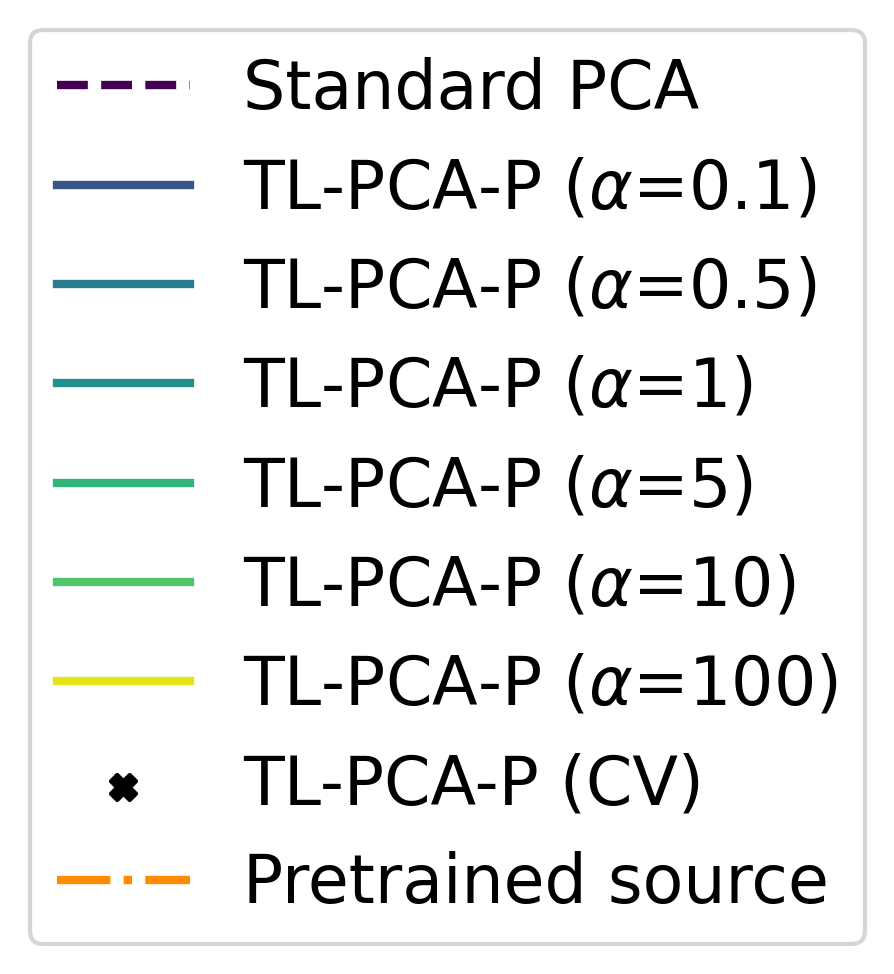}}
    \caption{TL-PCA-P train and test errors as a function of k for various values of $\alpha$, with $m=\left\lceil 0.8\cdot k \right\rceil$ principal directions transferred from the pretrained source (CIFAR-10 to SVHN). }
    \label{appendix:fig:tlpcap_train_test_errors_svhn_08}
\end{figure*}

\begin{figure*}
    \centering
    \subfloat[]{\includegraphics[width=0.33\linewidth]{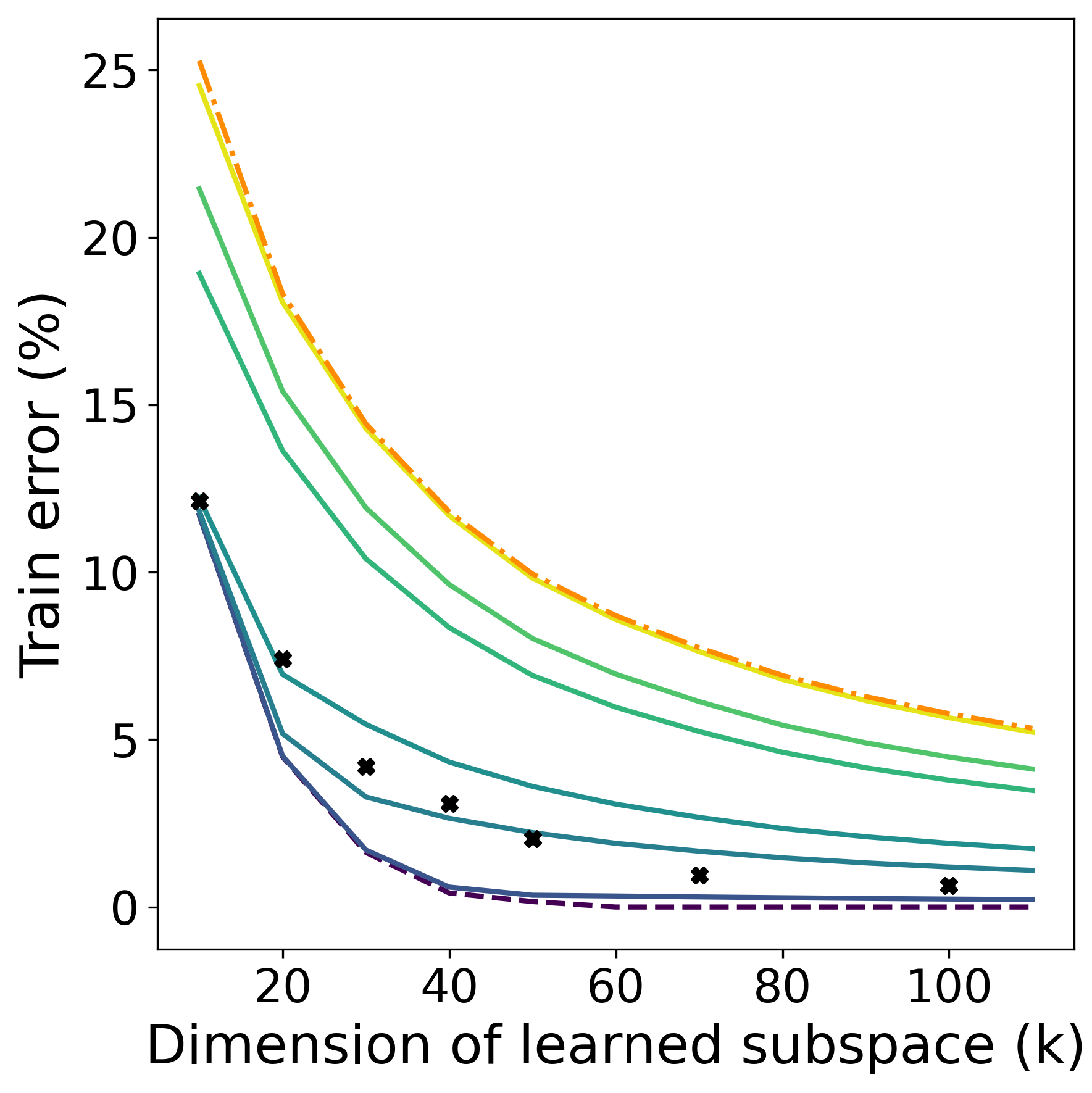}}
    \subfloat[]{\includegraphics[width=0.33\linewidth]{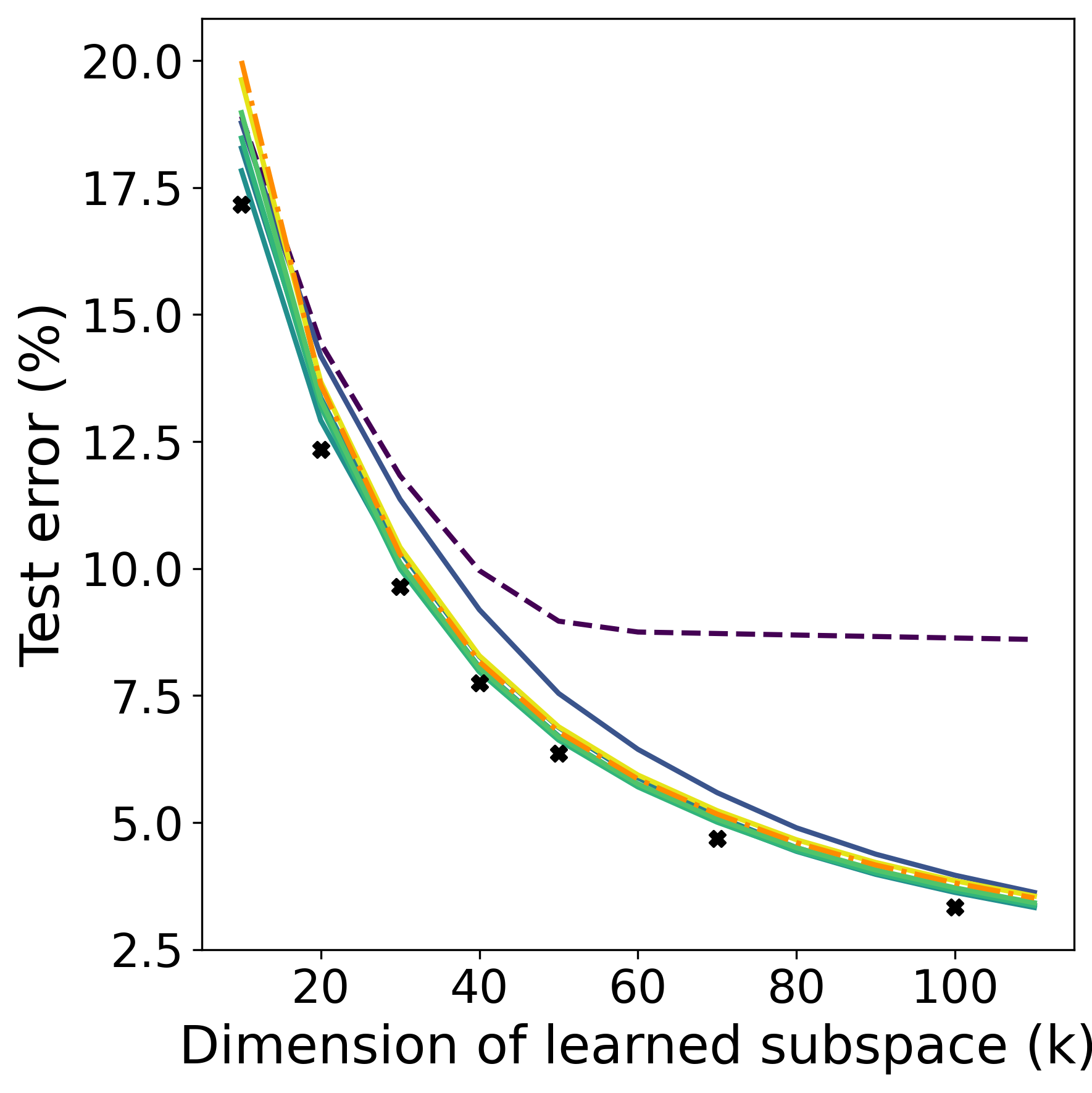}}
    \subfloat{\includegraphics[width=0.18\linewidth]{SVHN_results/alpha_values_legend.png}}
    \caption{TL-PCA-P train and test errors as a function of k for various values of $\alpha$, with $m=k$ principal directions transferred from the pretrained source (CIFAR-10 to SVHN). }
    \label{appendix:fig:tlpcap_train_test_errors_svhn_1}
\end{figure*}

\begin{figure*}
    \centering
    \subfloat[]{\includegraphics[width=0.33\linewidth]{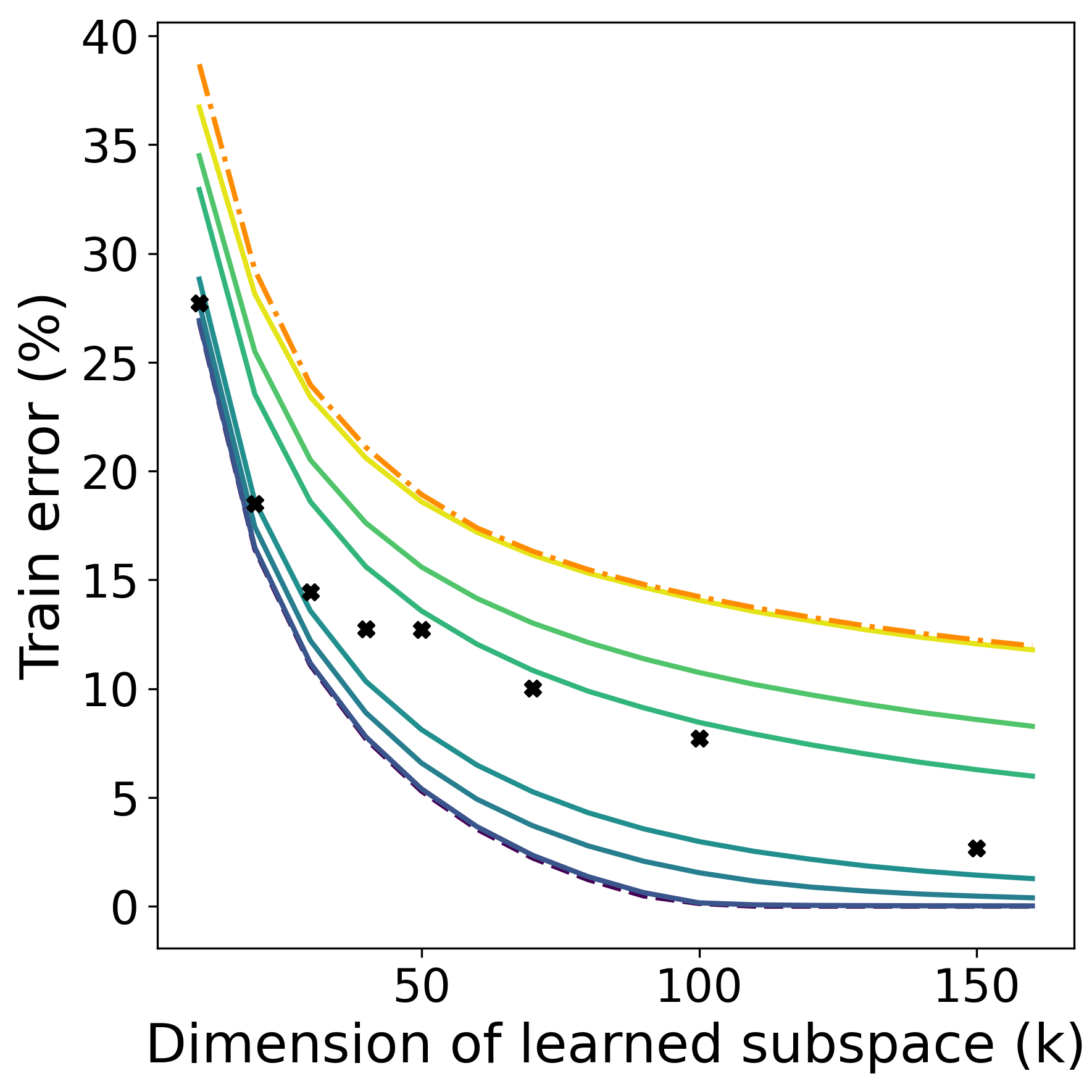}}
    \subfloat[]{\includegraphics[width=0.33\linewidth]{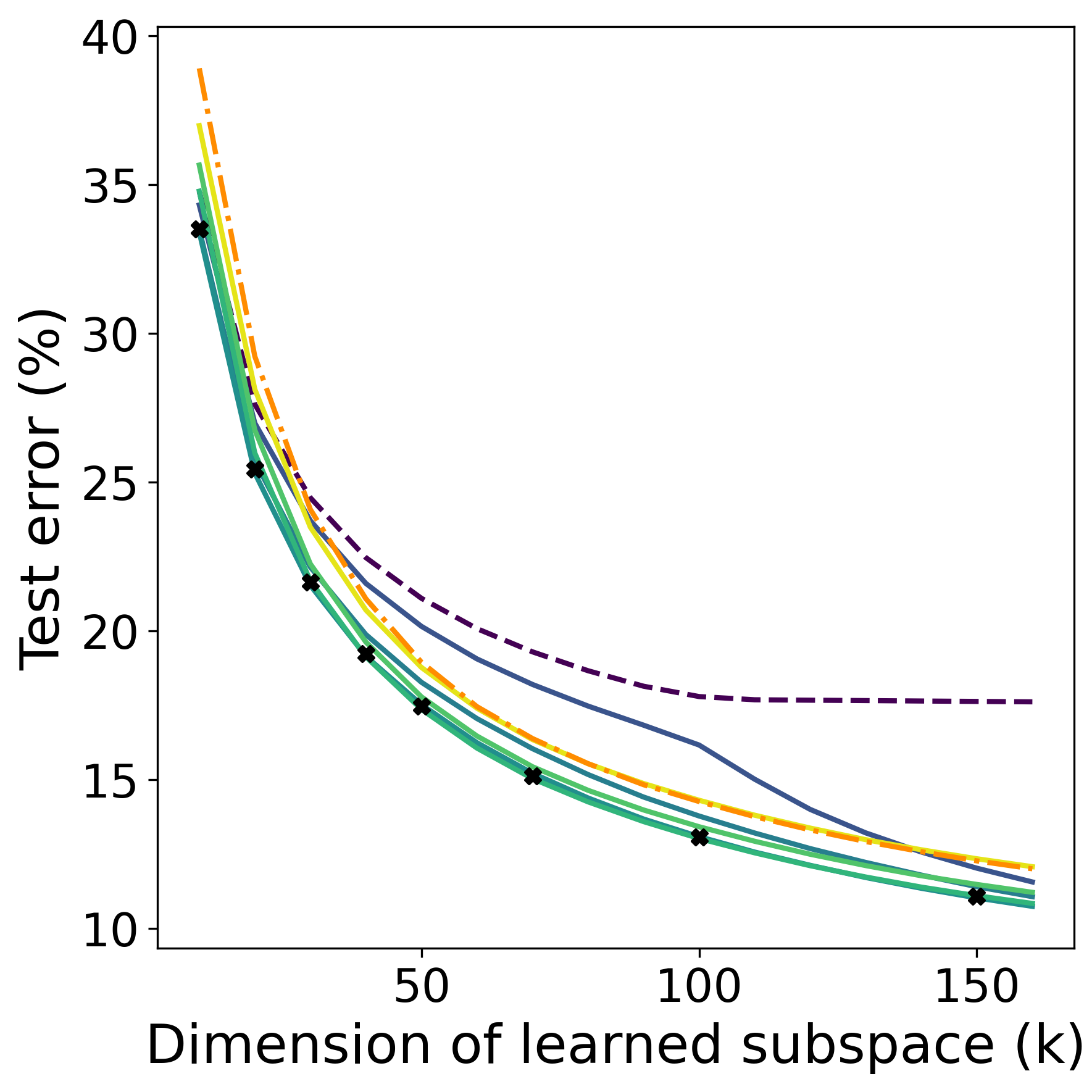}}
    \subfloat{\includegraphics[width=0.18\linewidth]{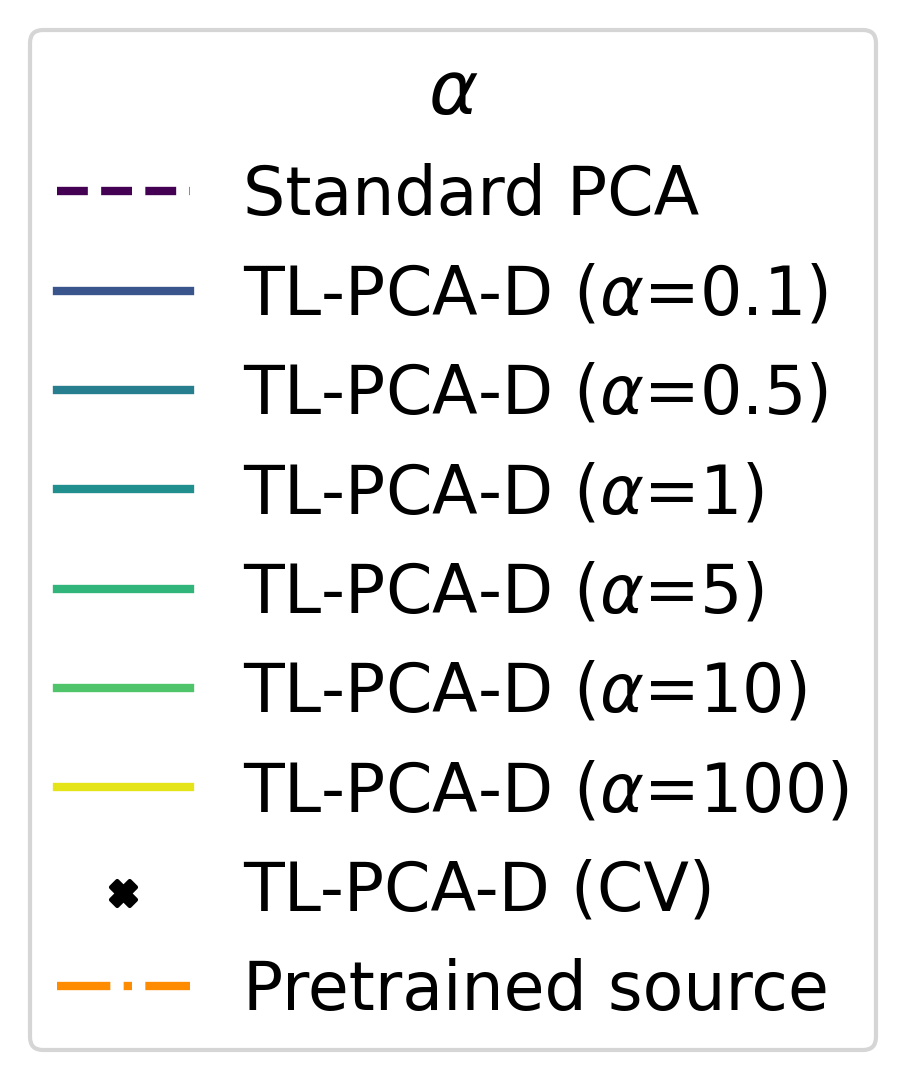}}
    \caption{TL-PCA-D train and test errors as a function of k for different values of $\alpha$ (Tiny-ImageNet to CelebA). }
    \label{appendix:fig:tlpcad_train_test_errors_celeba}
\end{figure*}

\begin{figure*}
    \centering
    \subfloat[]{\includegraphics[width=0.33\linewidth]{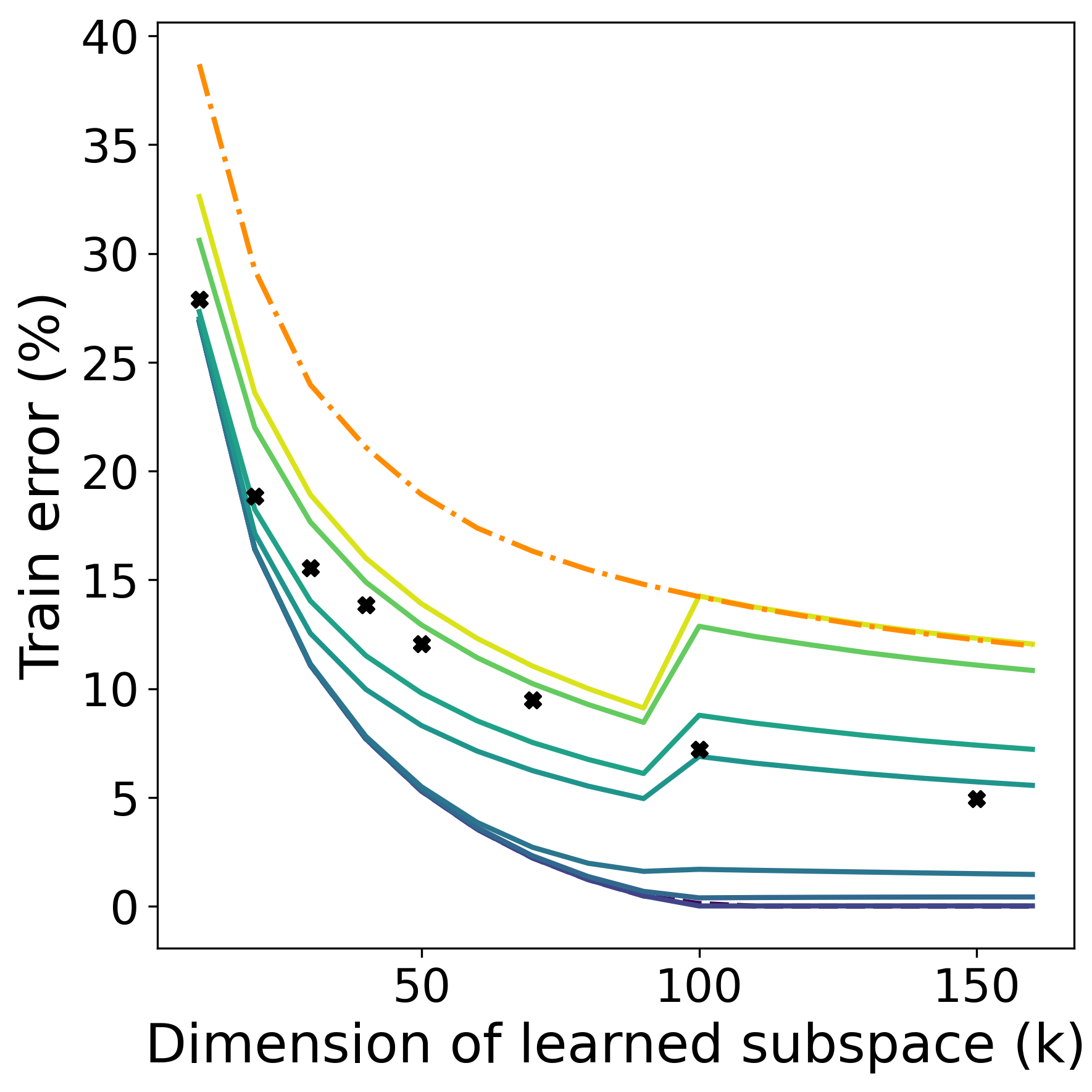}}
    \subfloat[]{\includegraphics[width=0.33\linewidth]{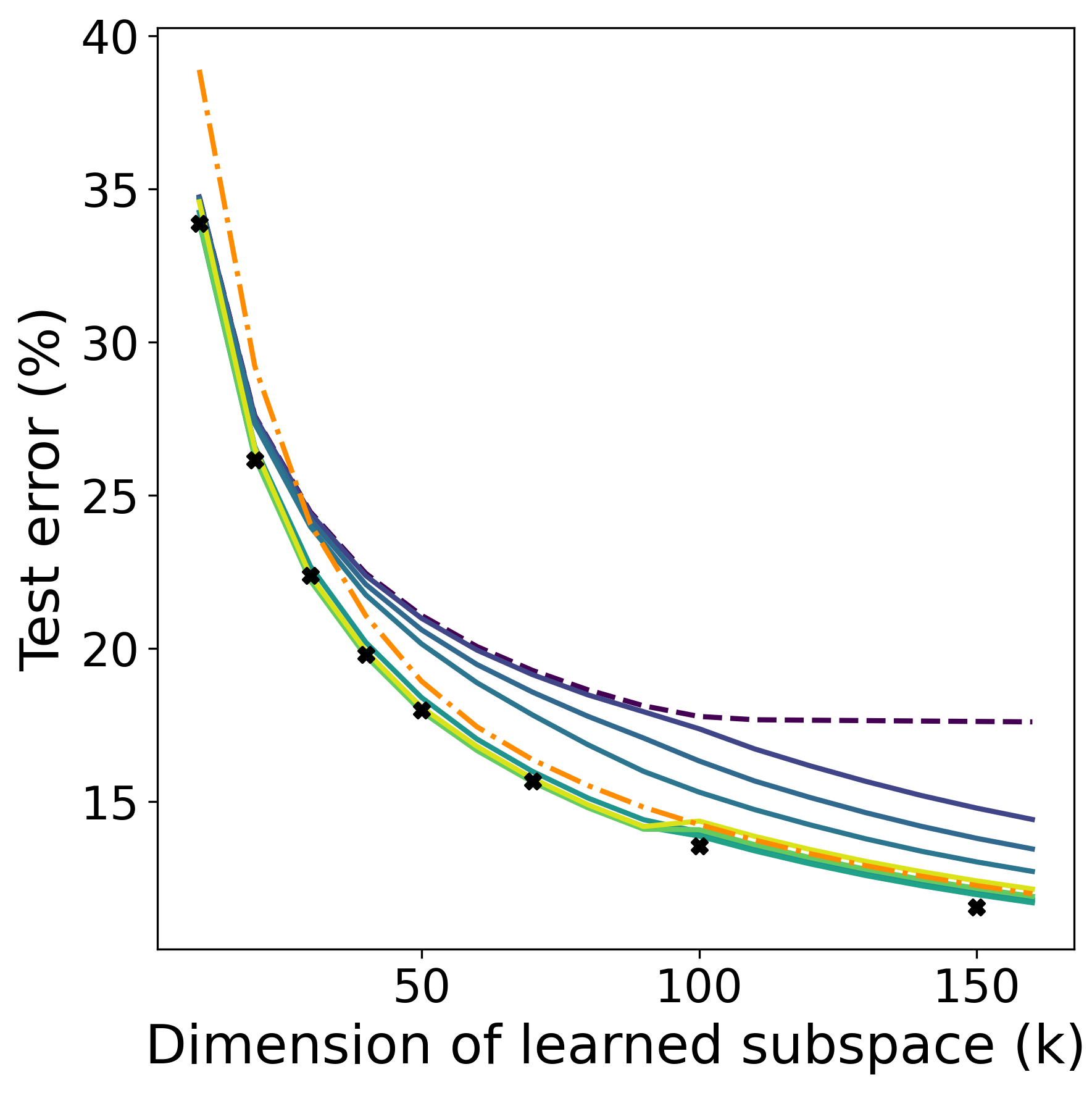}}
    \subfloat{\includegraphics[width=0.18\linewidth]{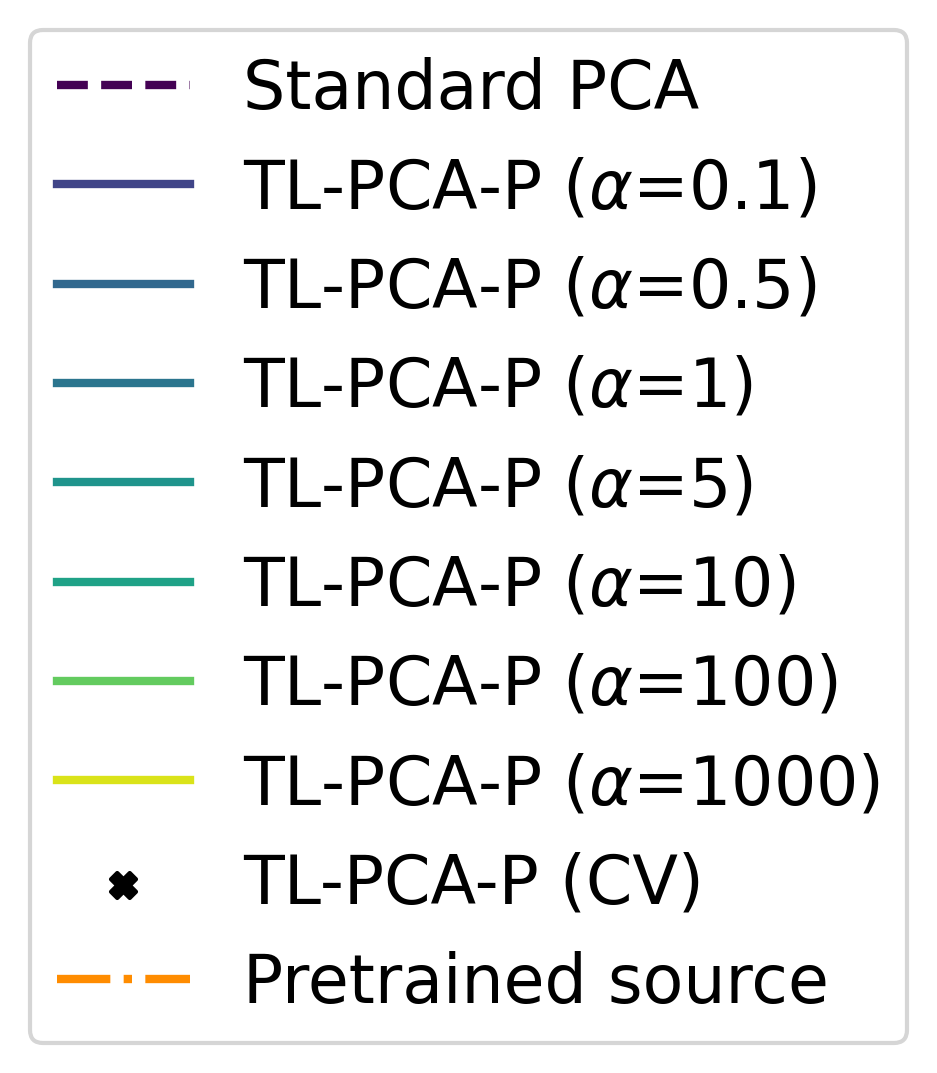}}
    \caption{TL-PCA-P train and test errors as a function of k for various values of $\alpha$, with $m=\left\lceil 0.8\cdot k \right\rceil$ principal directions transferred from the pretrained source (Tiny-ImageNet to CelebA). }
    \label{appendix:fig:tlpcap_train_test_errors_celeba_08}
\end{figure*}

\begin{figure*}
    \centering
    \subfloat[]{\includegraphics[width=0.33\linewidth]{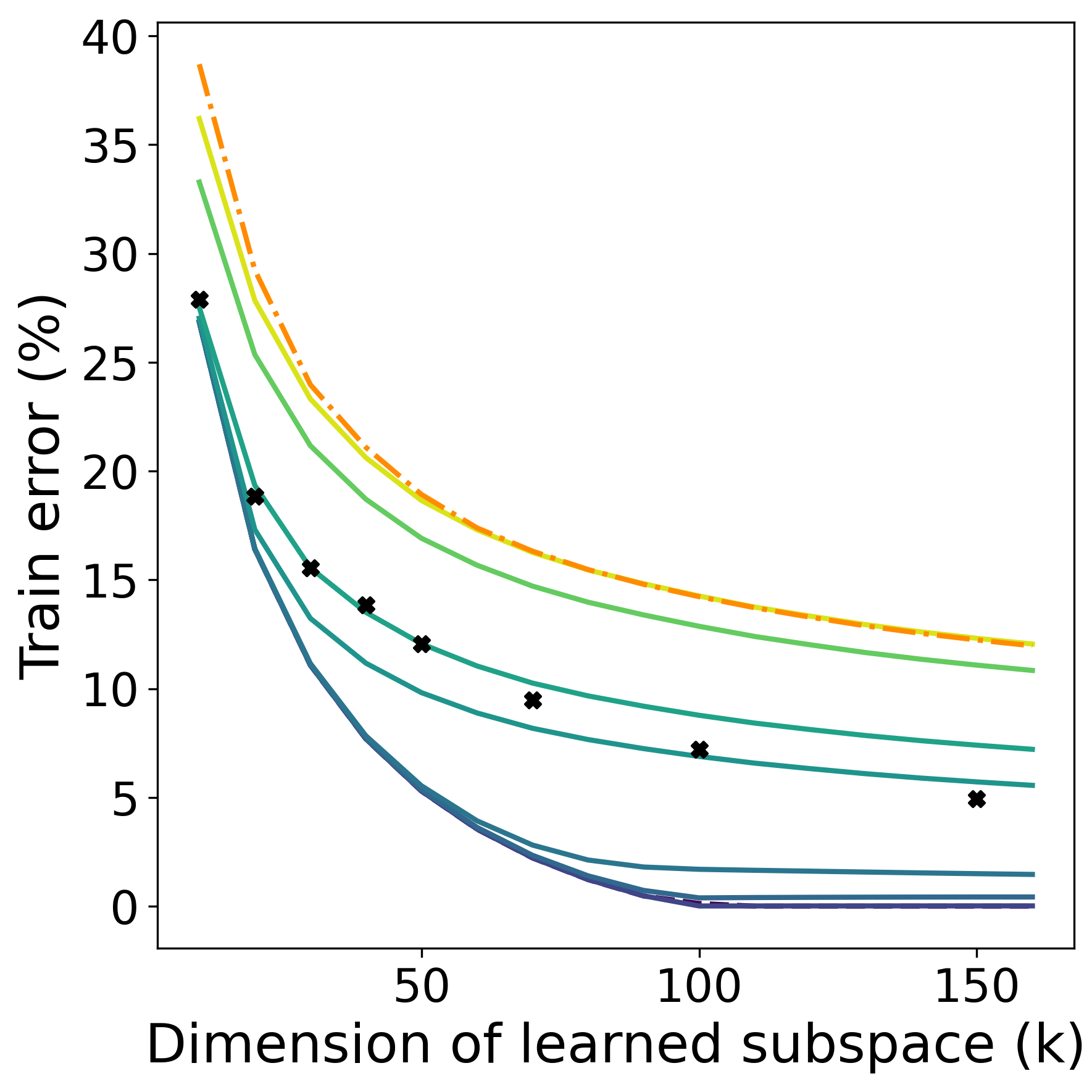}}
    \subfloat[]{\includegraphics[width=0.33\linewidth]{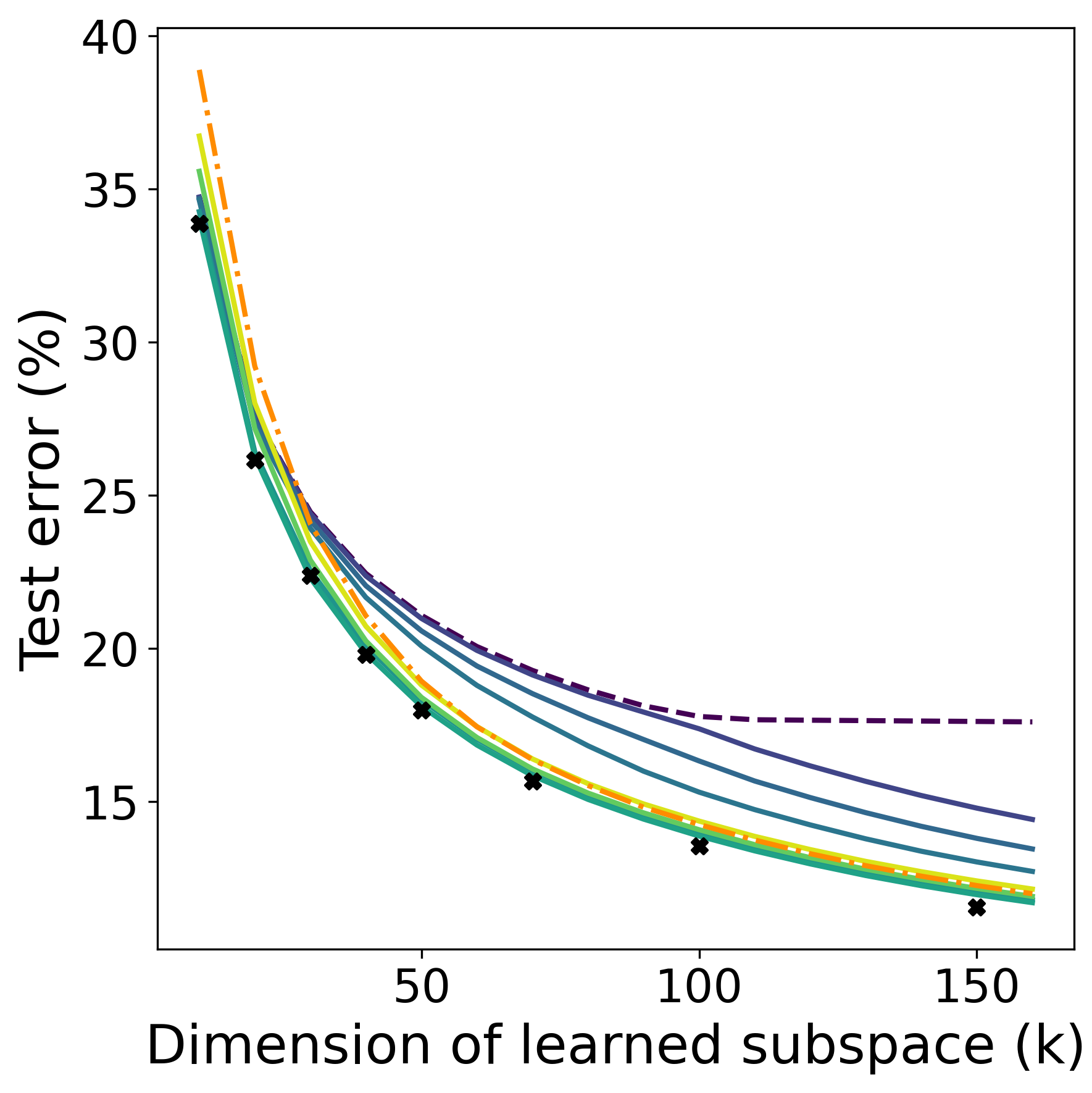}}
    \subfloat{\includegraphics[width=0.18\linewidth]{CELEBA_results/alpha_values_legend.png}}
    \caption{TL-PCA-P train and test errors as a function of k for various values of $\alpha$, with $m=k$ principal directions transferred from the pretrained source (Tiny-ImageNet to CelebA). }
    \label{appendix:fig:tlpcap_train_test_errors_celeba_1}
\end{figure*}

In Figures \ref{appendix:fig:tlpcad_train_test_errors_mnist}-\ref{appendix:fig:tlpcap_train_test_errors_celeba_1} we show error graphs including for TL-PCA with specific $\alpha$ values (i.e., without cross validation) and with markers that denote the cross validation results. These graphs are useful to observe the effect of the $\alpha$ value on the transfer learning strength.

\begin{figure}
    \centering
    \includegraphics[width=0.98\linewidth]{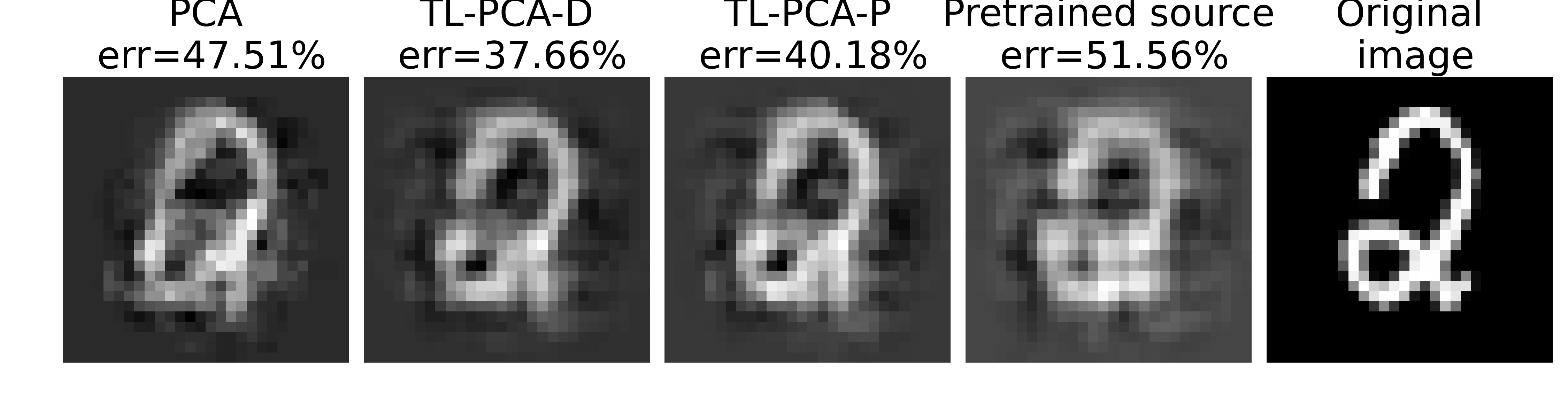}\\
    \includegraphics[width=0.98\linewidth]{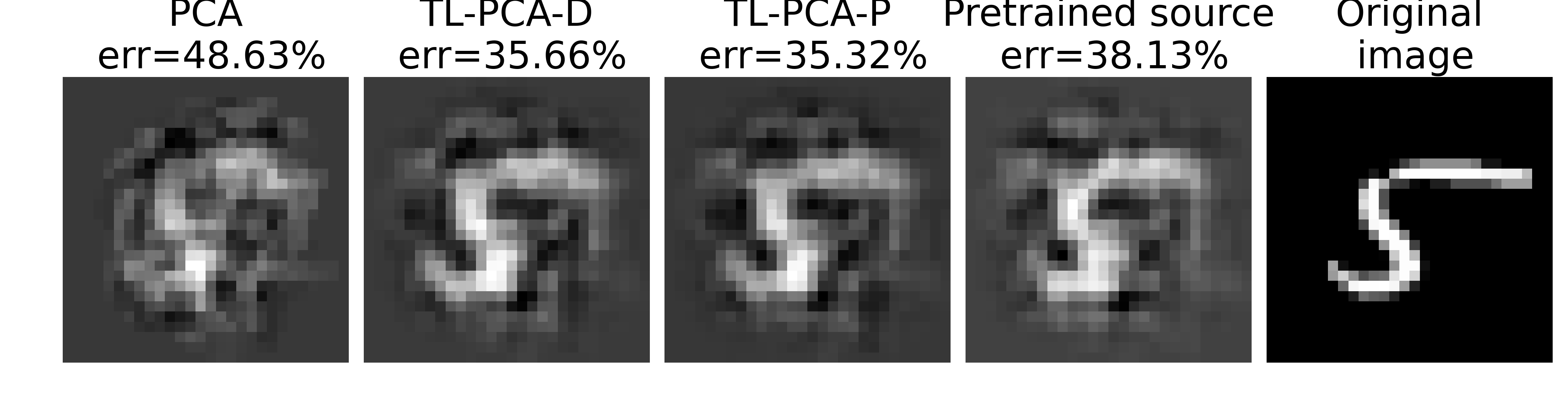}
    \includegraphics[width=0.98\linewidth]{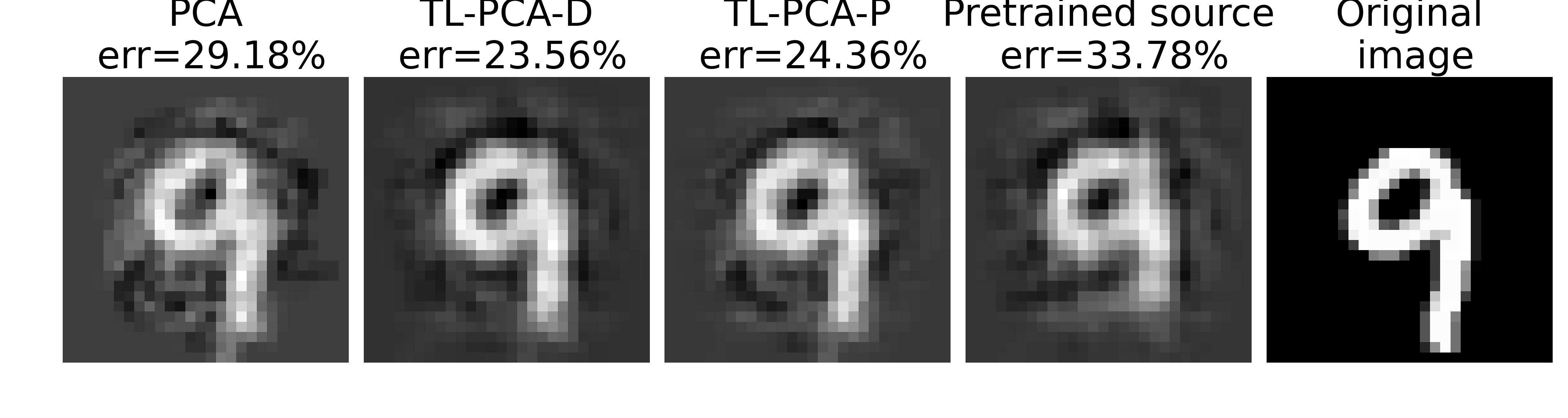}
    \caption{Comparison of the reconstructions of MNIST test examples ($k=40$).}
    \label{fig:mnist_rec_k40}
\end{figure}

\begin{figure}
    \centering
    \includegraphics[width=0.98\linewidth]{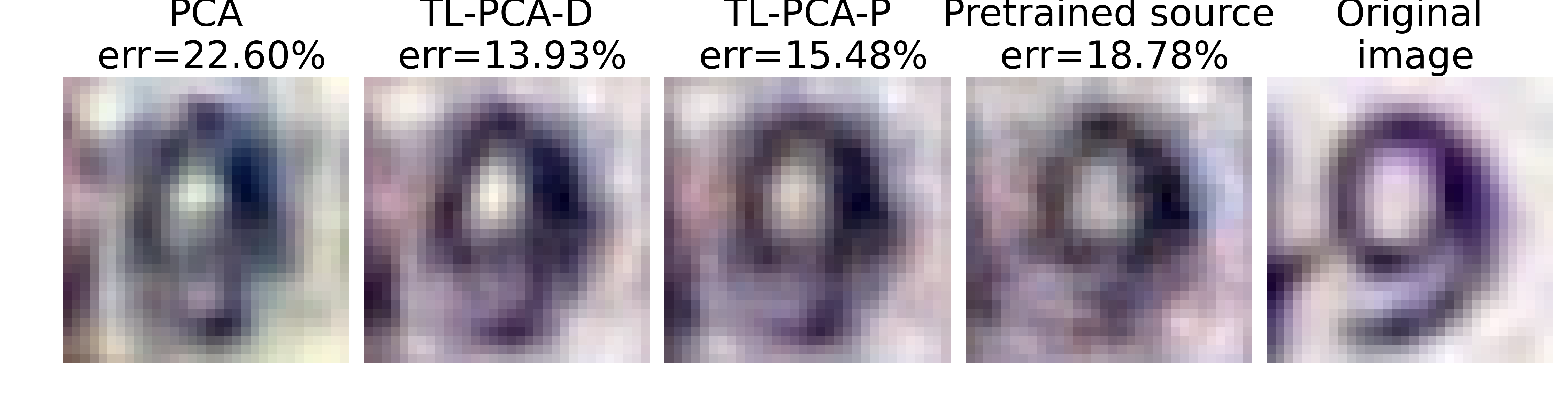}
    \includegraphics[width=0.98\linewidth]{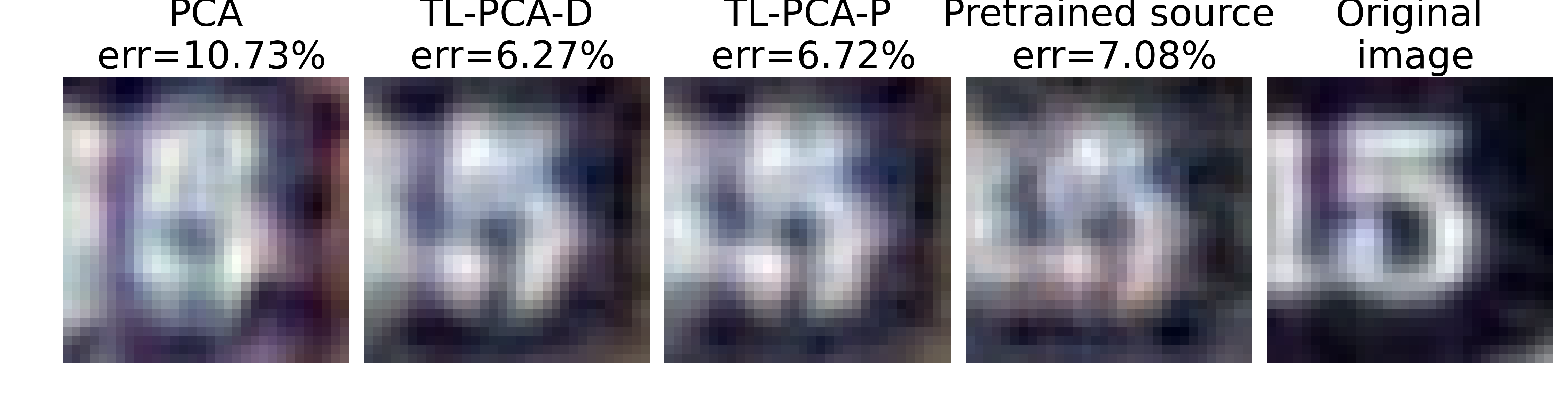}
    \includegraphics[width=0.98\linewidth]{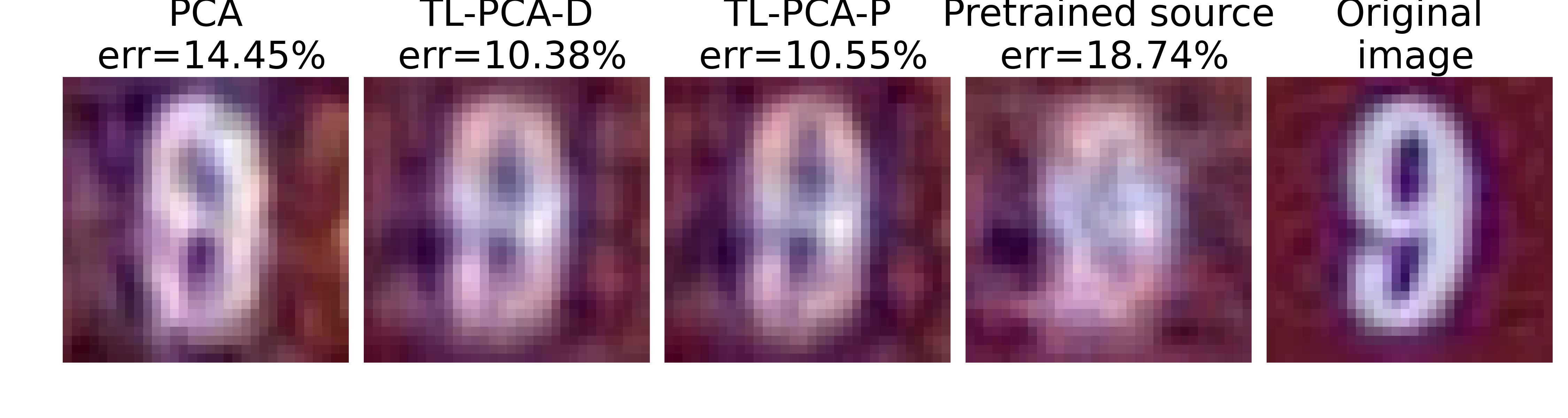}
    \includegraphics[width=0.98\linewidth]{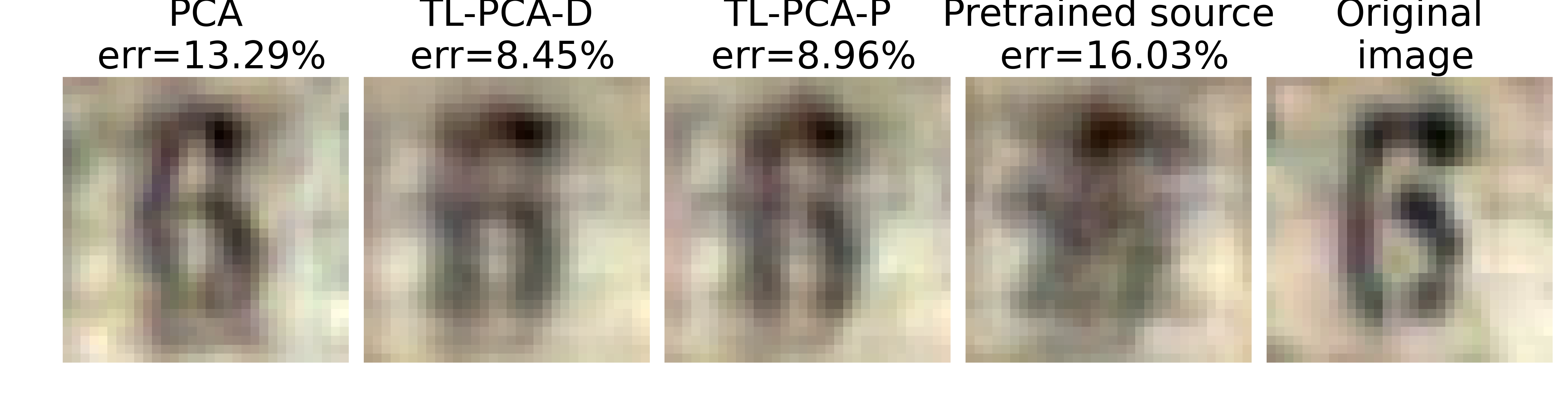}
    \caption{Comparison of the reconstructions of SVHN test examples ($k=40$).}
    \label{fig:svhn_rec_k40}
\end{figure}

\begin{figure}
    \centering
    \includegraphics[width=0.98\linewidth]{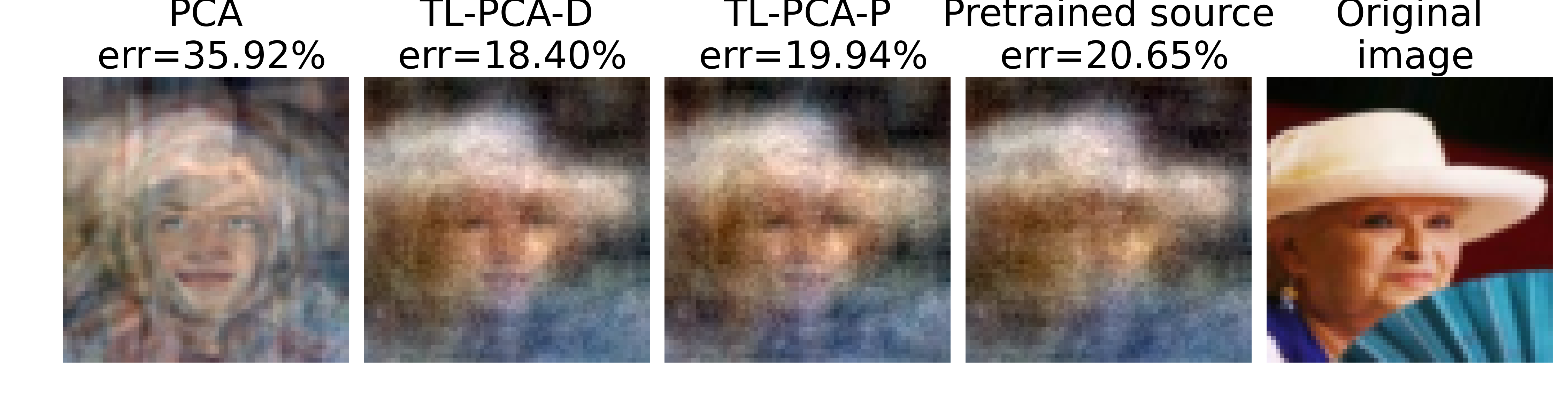}\\
\includegraphics[width=0.98\linewidth]{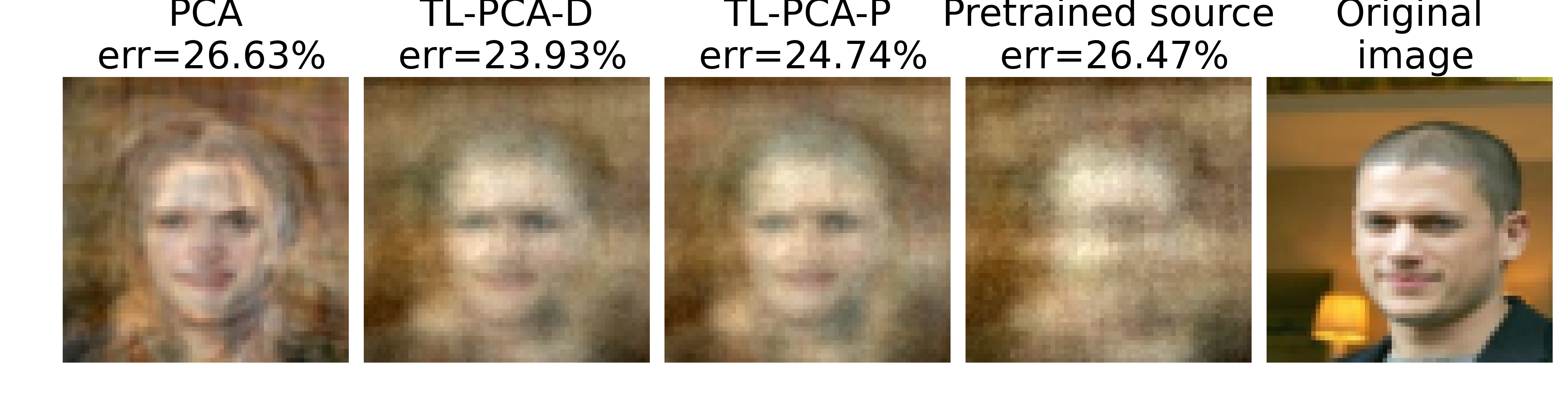}
\\
\includegraphics[width=0.98\linewidth]{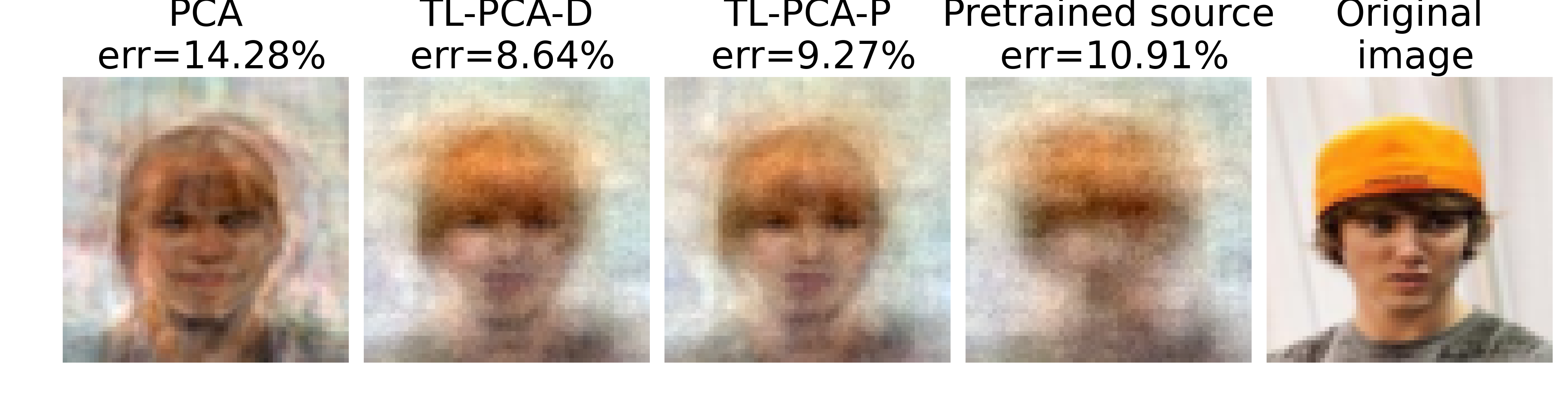}
\\    \includegraphics[width=0.98\linewidth]{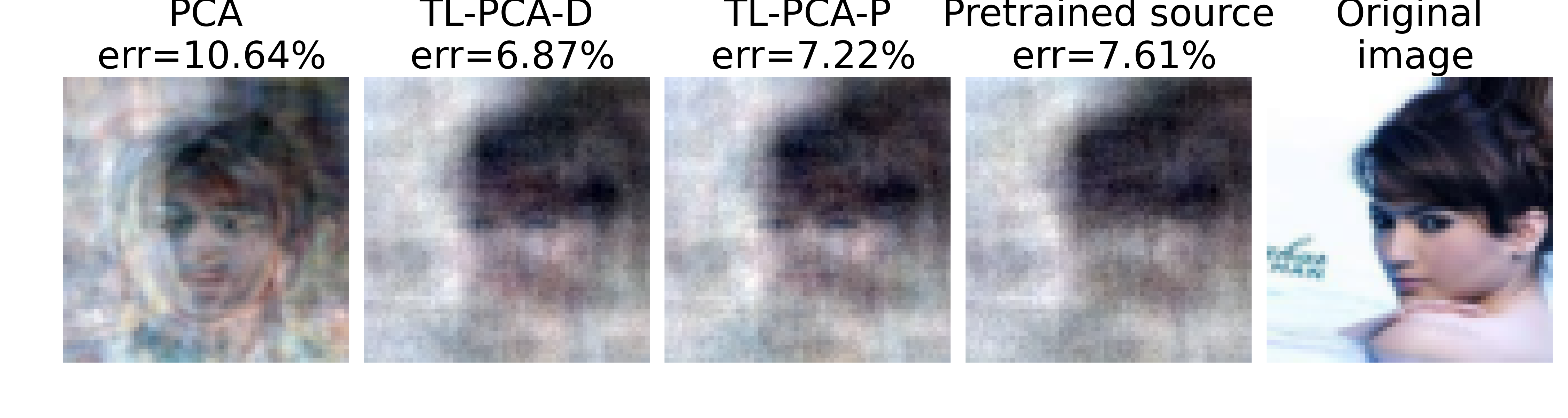}   
\\    \includegraphics[width=0.98\linewidth]{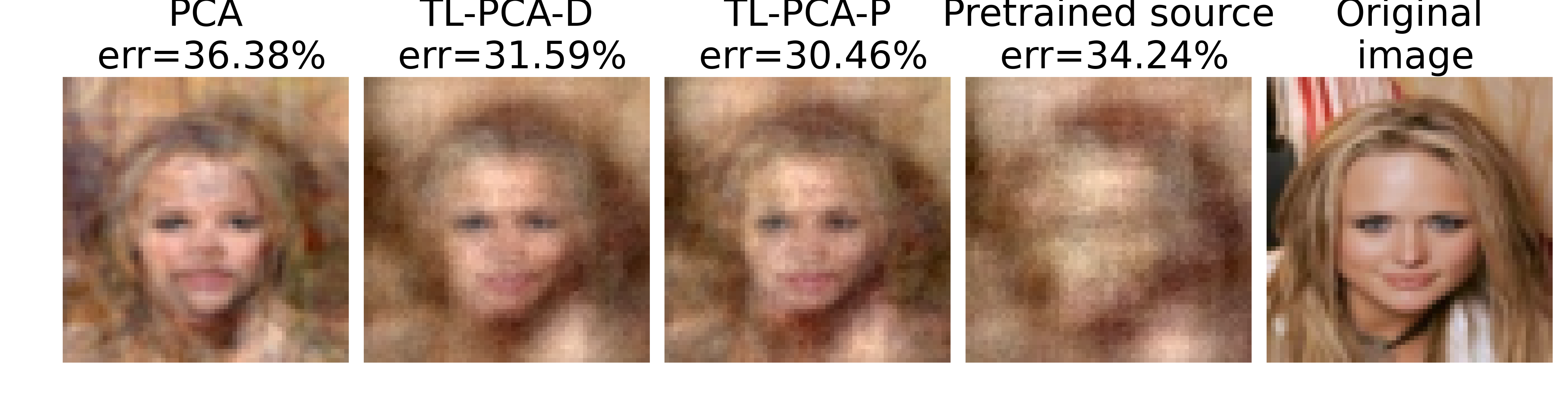}
    \caption{Comparison of the reconstructions of CelebA test examples ($k=80$).}
    \label{appendix:fig:celeba_rec_k80_5}
\end{figure}

In Figures \ref{fig:mnist_rec_k40}, \ref{fig:svhn_rec_k40}, \ref{appendix:fig:celeba_rec_k80_5} we show additional visual comparisons of images reconstructed by the TL-PCA approach and its competing methods. These further demonstrate the improved performance of TL-PCA.

\begin{table*}[h]
\caption{Test error rates for CIFAR10-to-SVHN with different methods at various values of $k$}
\label{app:table_svhn}
\begin{center}
\footnotesize
\begin{tabular}
{p{2.5cm}p{1.3cm}p{1.3cm}p{1.3cm}p{1.3cm}p{1.3cm}p{1.3cm}p{1.3cm}p{1.3cm}}
\toprule
& \multicolumn{1}{c}{$k=10$} & \multicolumn{1}{c}{$k=20$} & \multicolumn{1}{c}{$k=30$} & \multicolumn{1}{c}{$k=40$} & \multicolumn{1}{c}{$k=50$} & \multicolumn{1}{c}{$k=70$} & \multicolumn{1}{c}{$k=100$} & \multicolumn{1}{c}{$k=150$} \\
\midrule
TL-PCA-D (CV) & \textbf{17.07}\tiny{$\pm0.34$} & \textbf{11.86}\tiny{$\pm0.34$} & \textbf{8.92}\tiny{$\pm0.23$} & \textbf{7.07}\tiny{$\pm0.15$} & \textbf{5.77}\tiny{$\pm0.12$} & \textbf{4.26}\tiny{$\pm0.12$} & \textbf{3.05}\tiny{$\pm0.07$} & \textbf{2.06}\tiny{$\pm0.06$} \\ 
TL-PCA-P (CV) & \textbf{17.17}\tiny{$\pm0.50$} & \textbf{12.34}\tiny{$\pm0.40$} & 9.65\tiny{$\pm0.29$} & 7.75\tiny{$\pm0.27$} & 6.36\tiny{$\pm0.18$} & 4.69\tiny{$\pm0.15$} & 3.33\tiny{$\pm0.07$} & 2.24\tiny{$\pm0.06$} \\
\midrule
Pretrained source \tiny{(w.~target mean)} & 19.95\tiny{$\pm0.44$} & 13.83\tiny{$\pm0.26$} & 10.39\tiny{$\pm0.17$} & 8.36\tiny{$\pm0.18$} & 6.93\tiny{$\pm0.14$} & 5.21\tiny{$\pm0.12$} & 3.87\tiny{$\pm0.09$} & 2.75\tiny{$\pm0.06$} \\
Pretrained source \tiny{(w.~source mean)} & 19.87\tiny{$\pm0.32$} & 13.62\tiny{$\pm0.13$} & 10.12\tiny{$\pm0.12$} & 8.15\tiny{$\pm0.13$} & 6.75\tiny{$\pm0.10$} & 5.07\tiny{$\pm0.08$} & 3.77\tiny{$\pm0.07$} & 2.68\tiny{$\pm0.06$} \\
Standard PCA & 18.60\tiny{$\pm0.62$} & 13.94\tiny{$\pm0.45$} & 11.48\tiny{$\pm0.38$} & 9.78\tiny{$\pm0.33$} & 8.79\tiny{$\pm0.23$} & 8.55\tiny{$\pm0.23$} & 8.46\tiny{$\pm0.23$} & 8.32\tiny{$\pm0.23$} \\ 
\hline
\end{tabular}
\end{center}
\end{table*}

\begin{figure*}
    \centering
    \subfloat[$k=20$]{\includegraphics[width=0.25\linewidth]{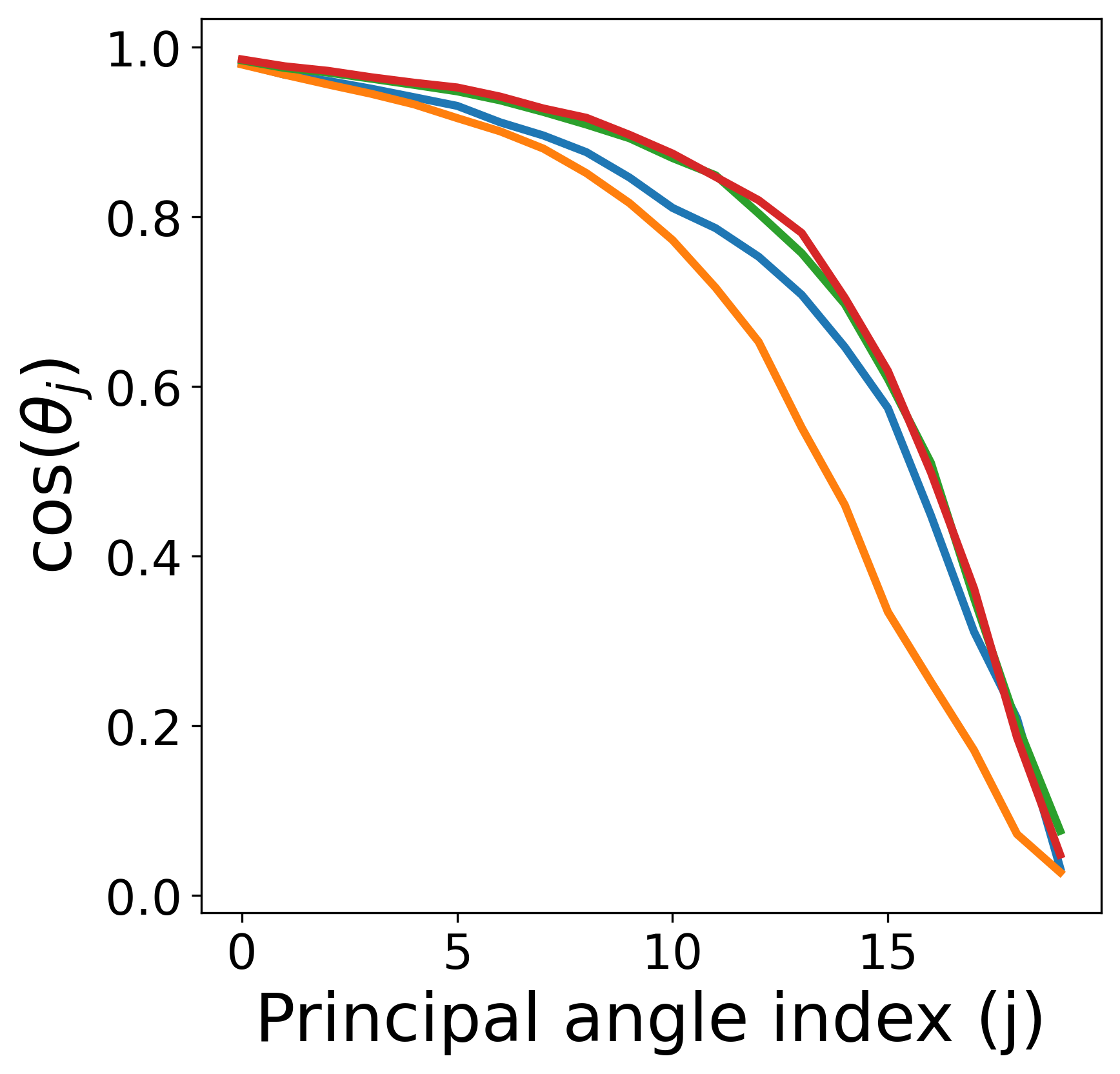}}
    \subfloat[$k=30$]{\includegraphics[width=0.25\linewidth]{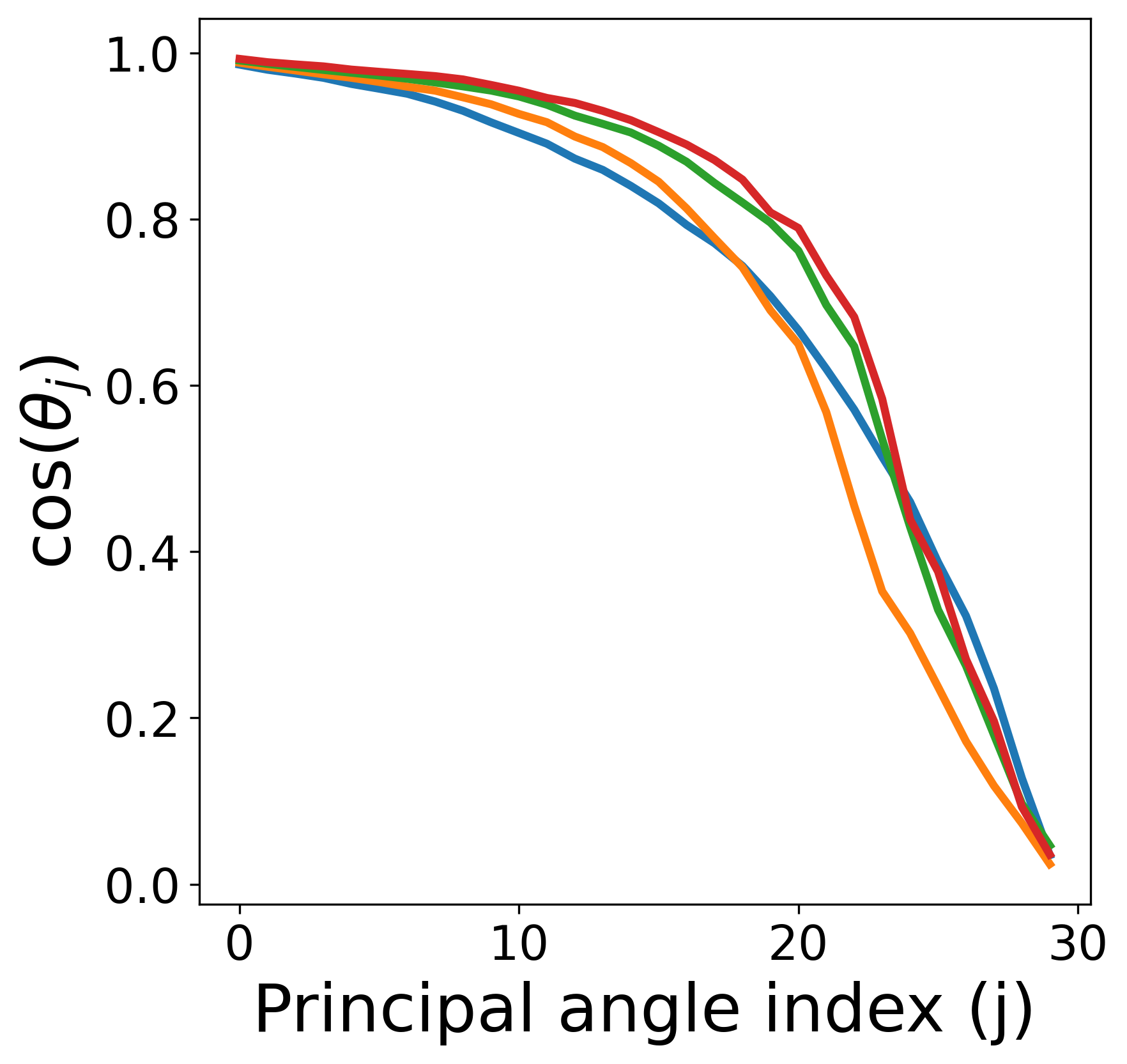}}
    \subfloat[$k=40$]{\includegraphics[width=0.25\linewidth]{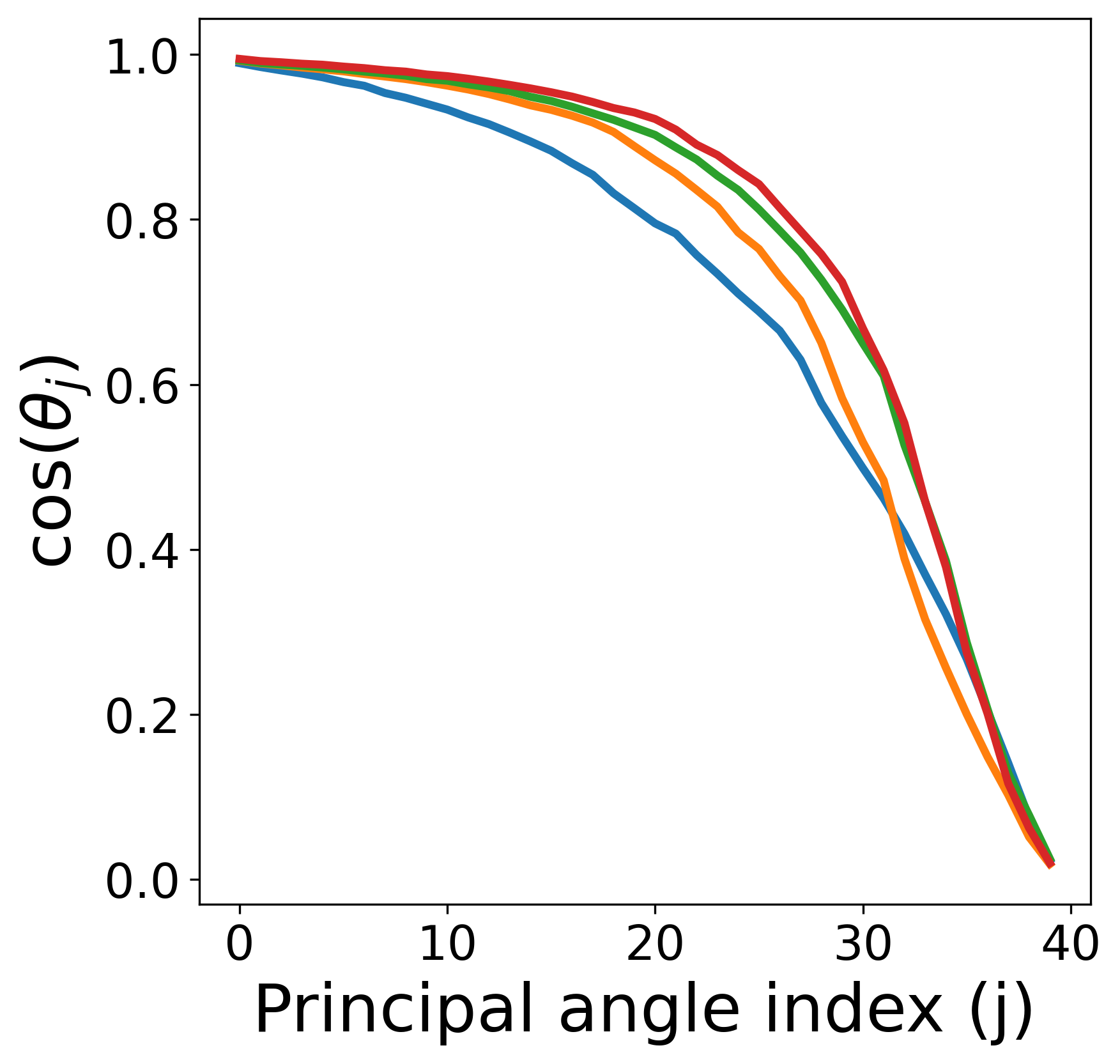}}
    
    \subfloat[$k=50$]{\includegraphics[width=0.25\linewidth]{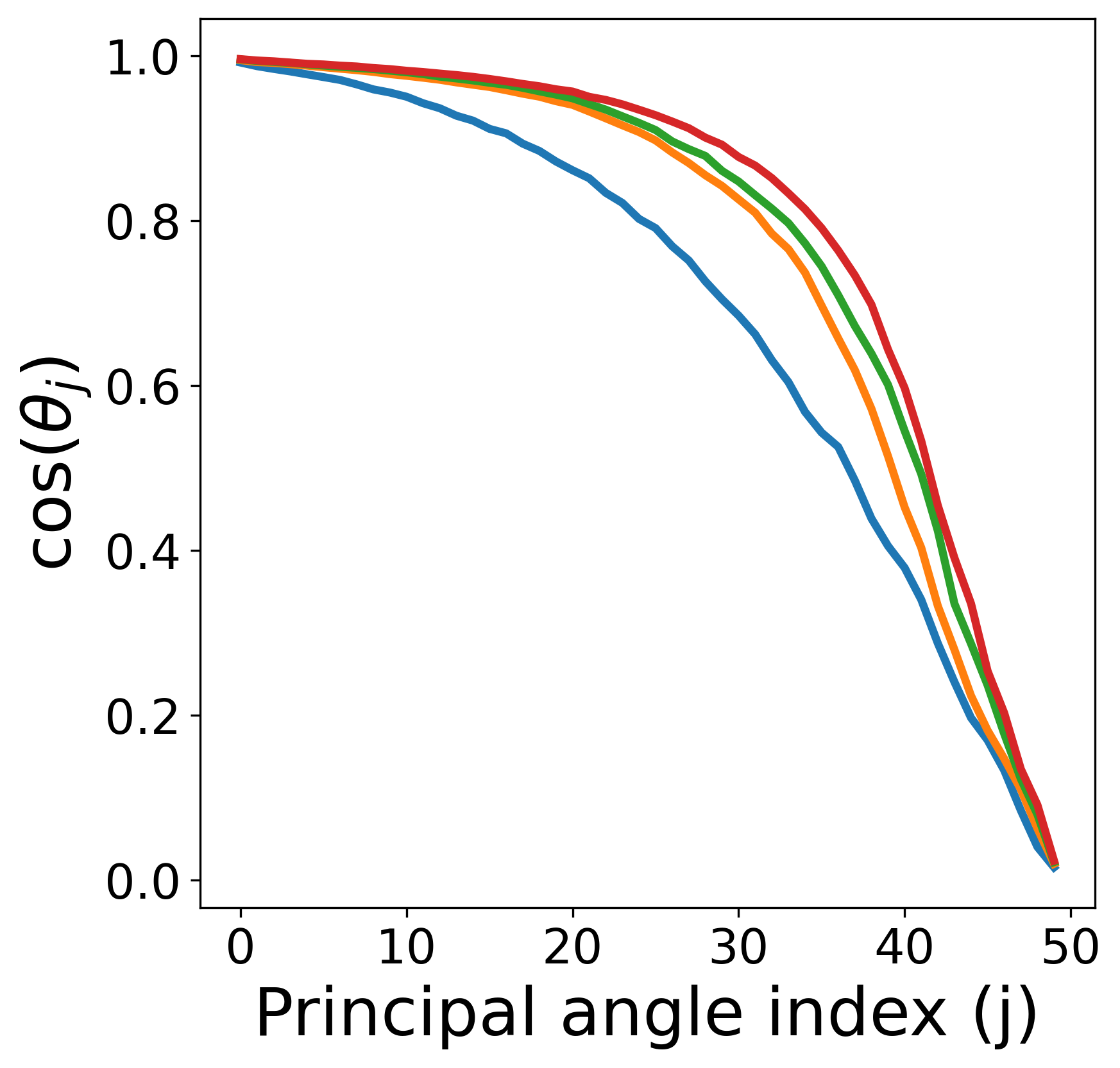}}
    \subfloat[$k=80$]{\includegraphics[width=0.25\linewidth]{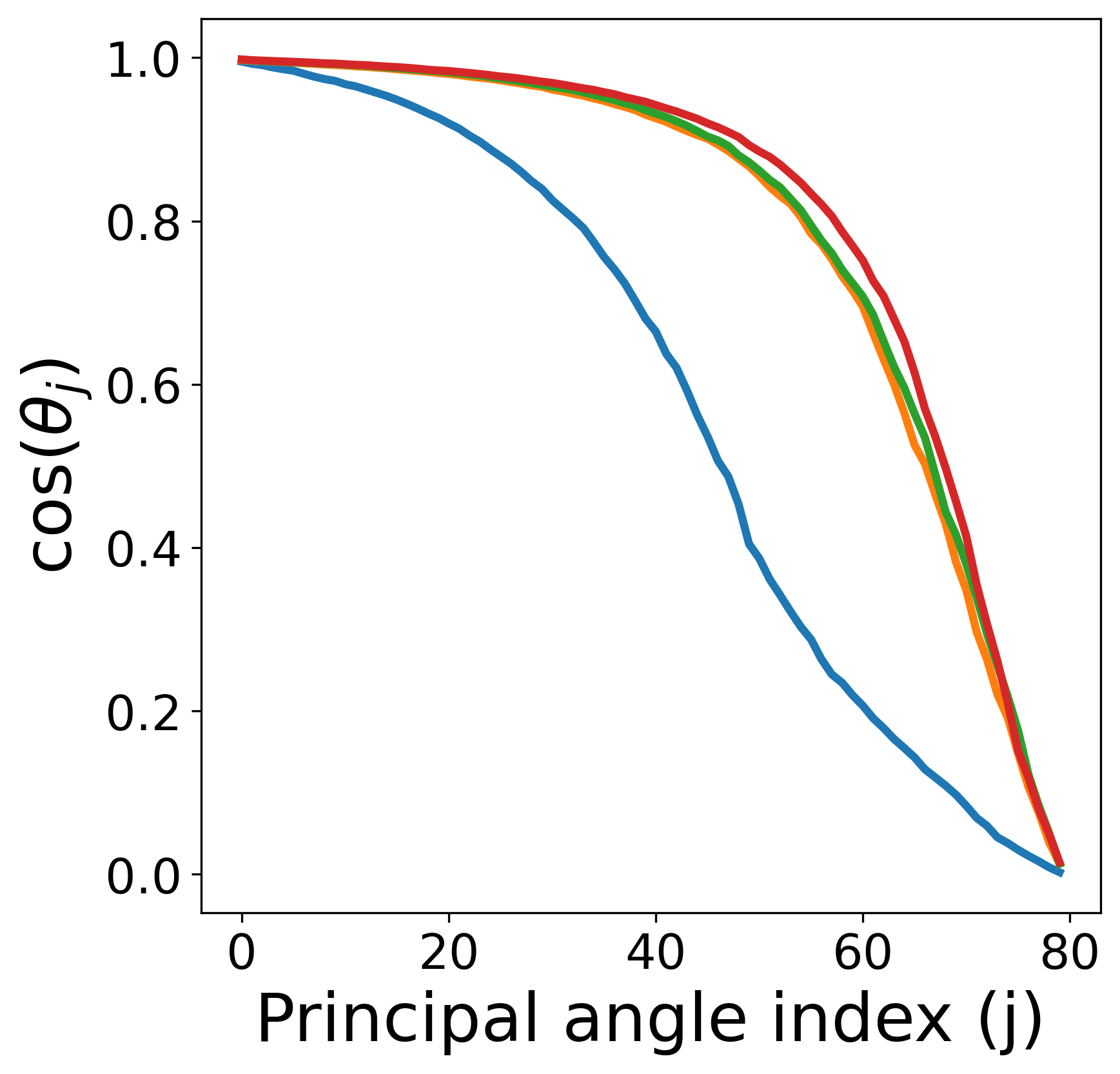}}
    \subfloat{\includegraphics[width=0.2\linewidth]{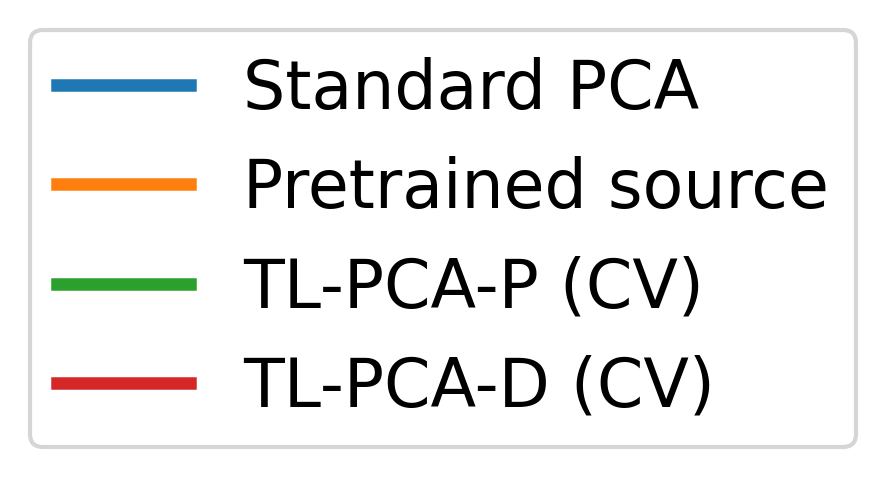}}
    \caption{Principal angles w.r.t.~ideal PCA (Omniglot to MNIST). }
    \label{appendix:fig:principal_angles_graphs_mnist}
\end{figure*}

\begin{figure*}
    \centering
    \subfloat[$k=20$]{\includegraphics[width=0.25\linewidth]{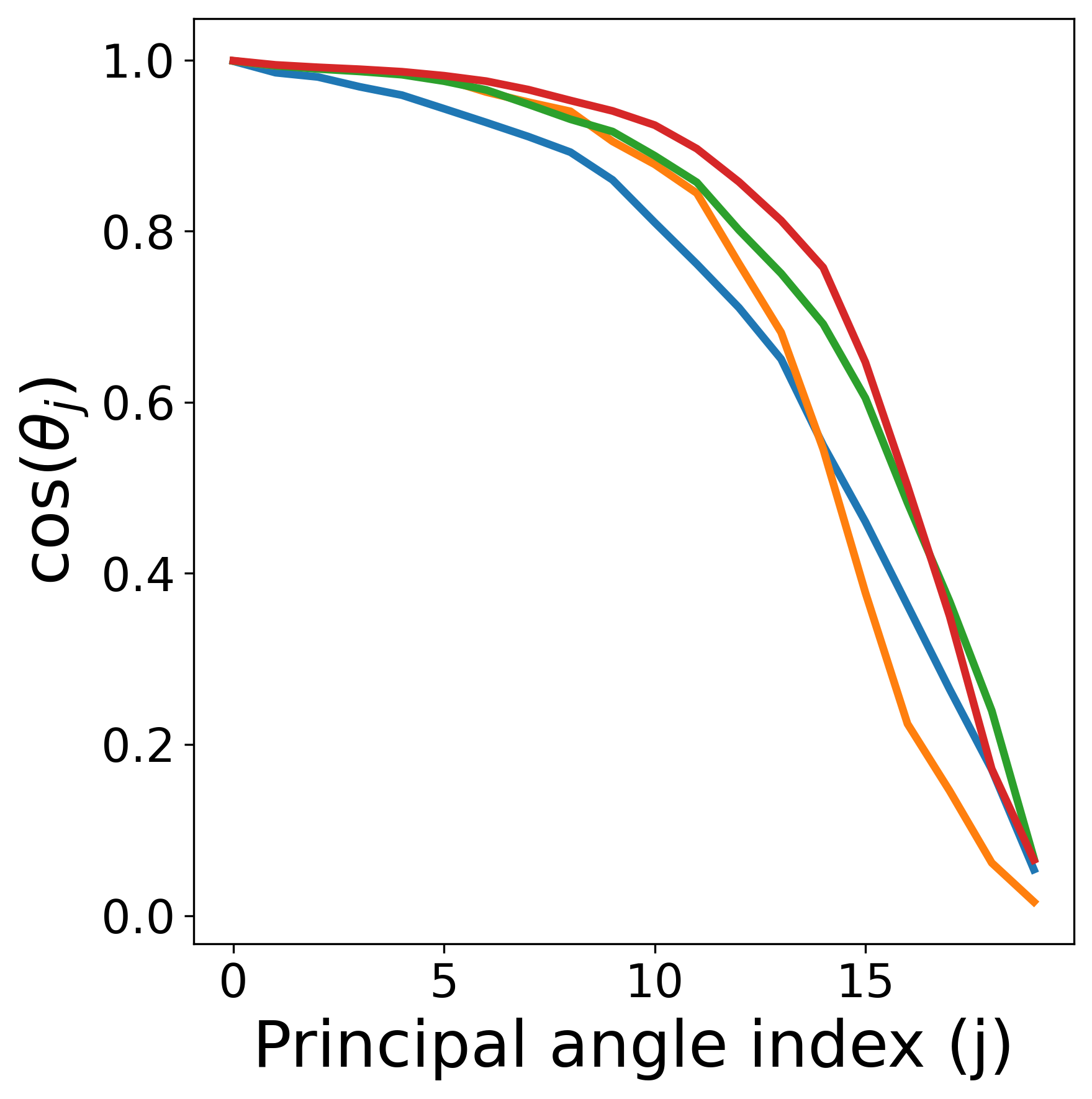}}
    \subfloat[$k=30$]{\includegraphics[width=0.25\linewidth]{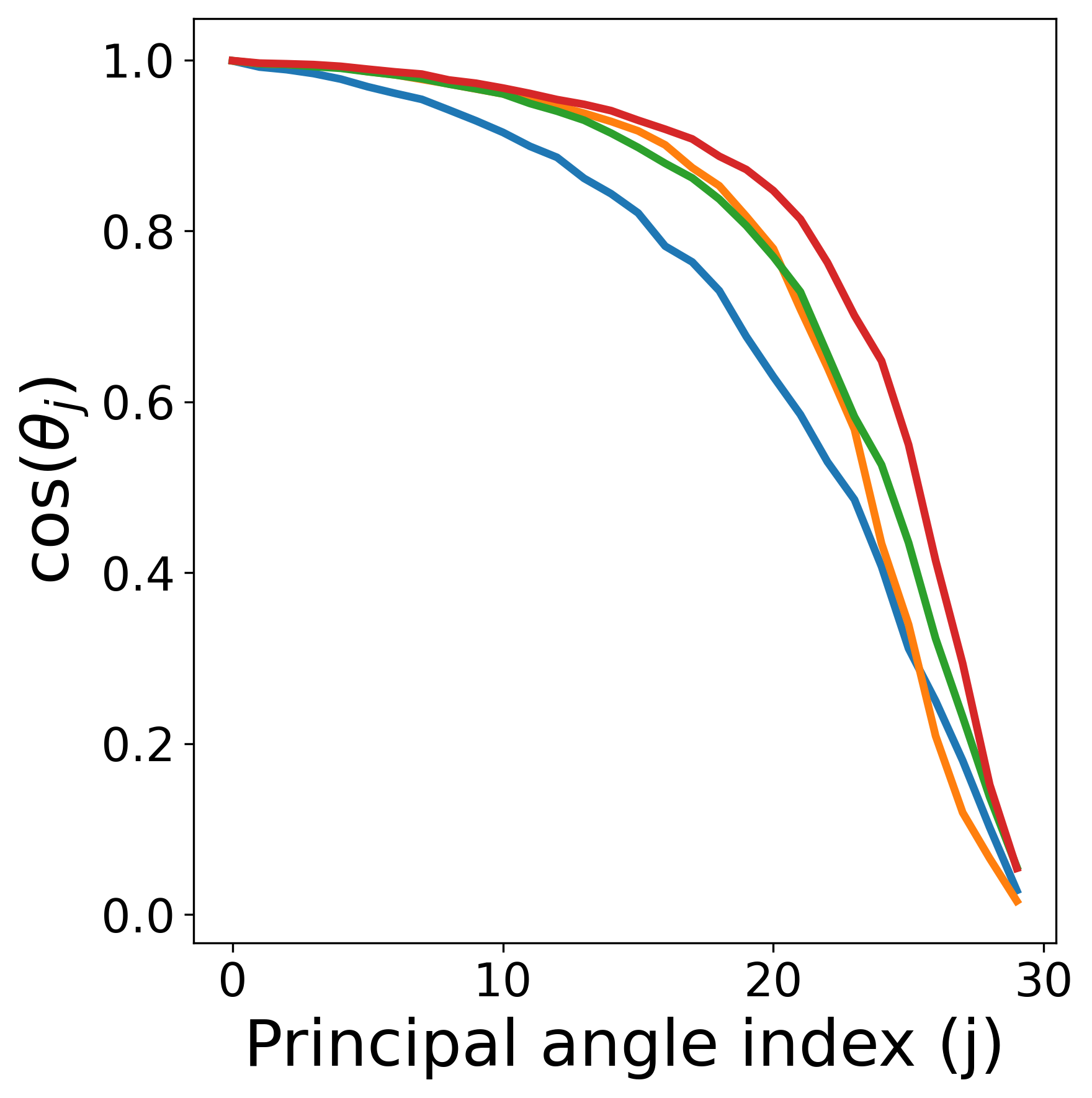}}
    \subfloat[$k=40$]{\includegraphics[width=0.25\linewidth]{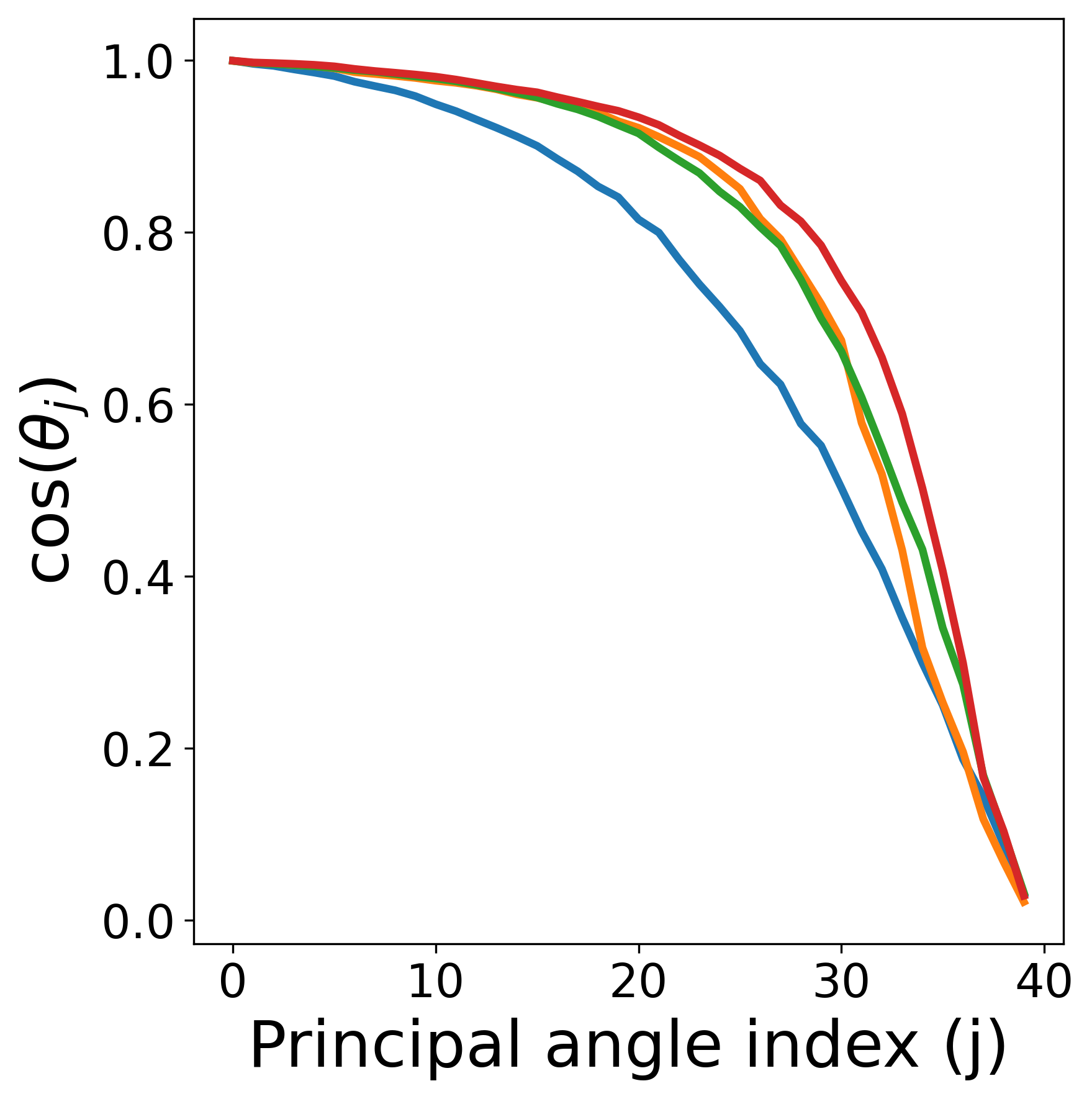}}
    
    \subfloat[$k=50$]{\includegraphics[width=0.25\linewidth]{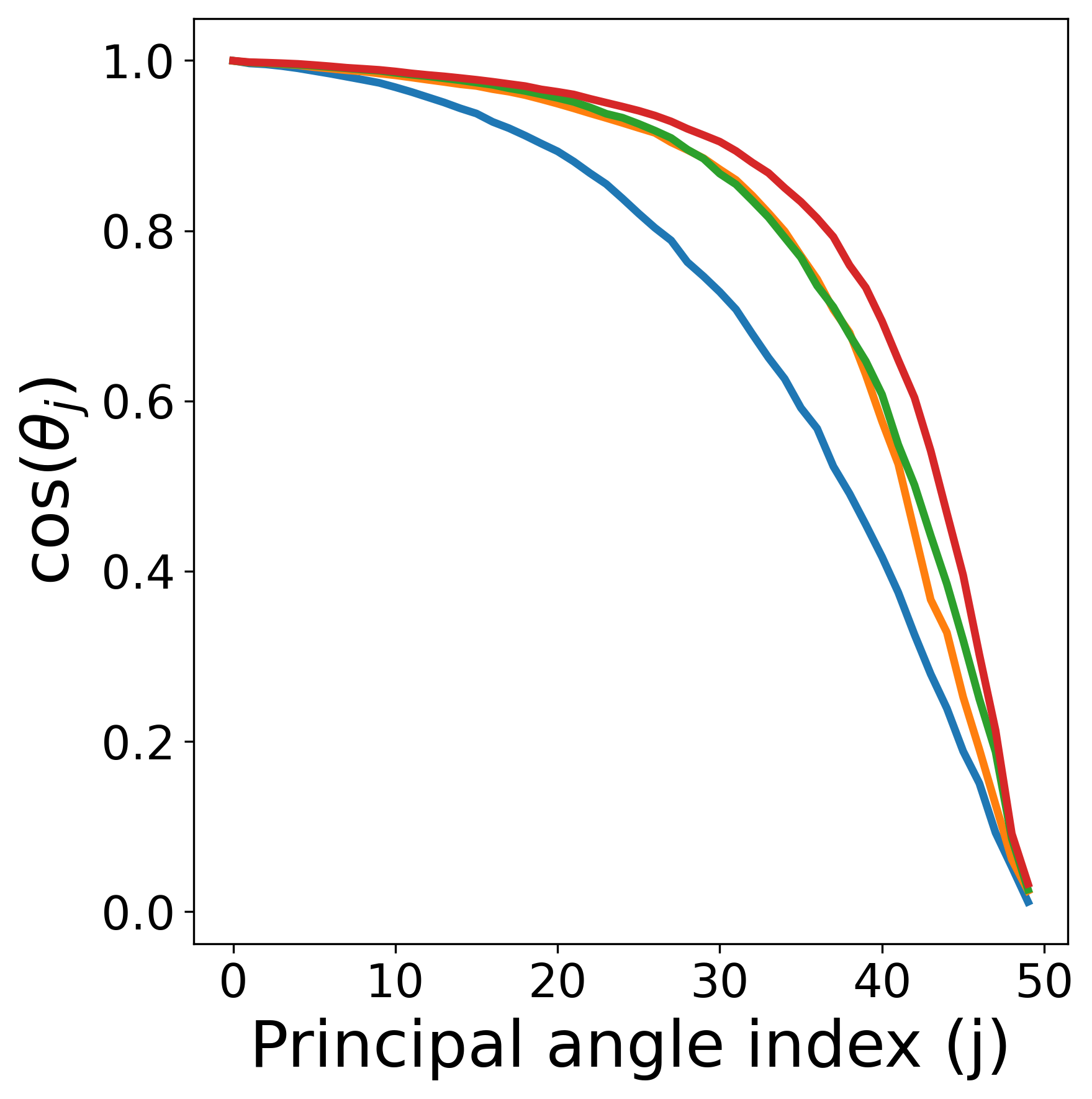}}
    \subfloat[$k=80$]{\includegraphics[width=0.25\linewidth]{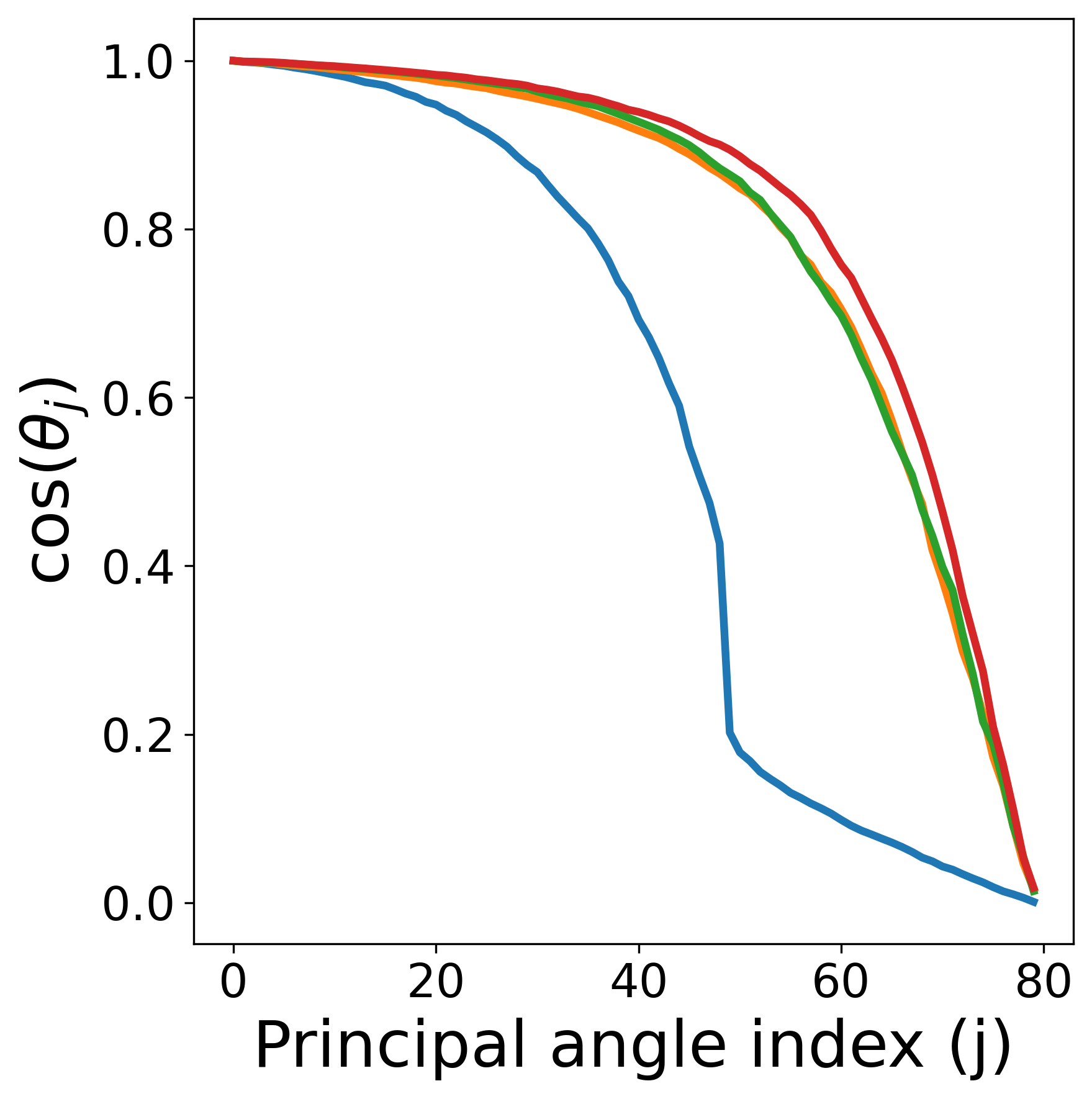}}
    \subfloat{\includegraphics[width=0.2\linewidth]{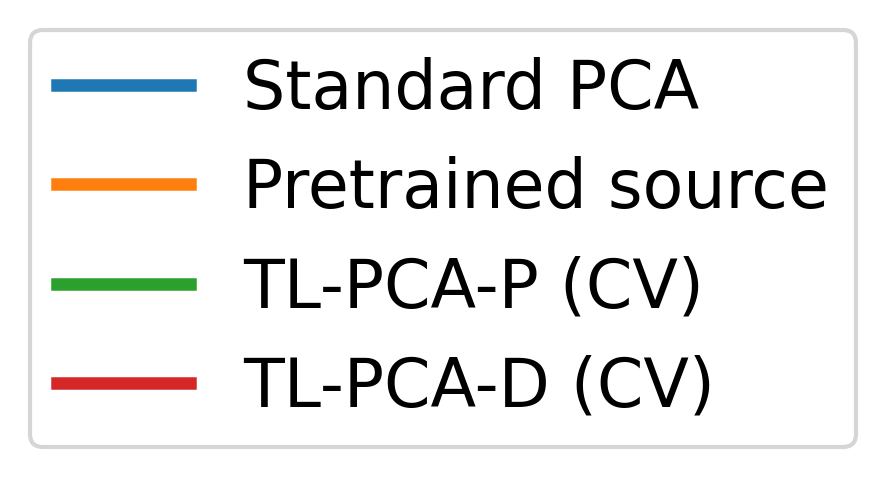}}
    \caption{Principal angles w.r.t.~ideal PCA (CIFAR-10 to SVHN). }
    \label{appendix:fig:principal_angles_graphs_svhn}
\end{figure*}

\begin{figure*}
    \centering
    \subfloat[$k=30$]{\includegraphics[width=0.24\linewidth]{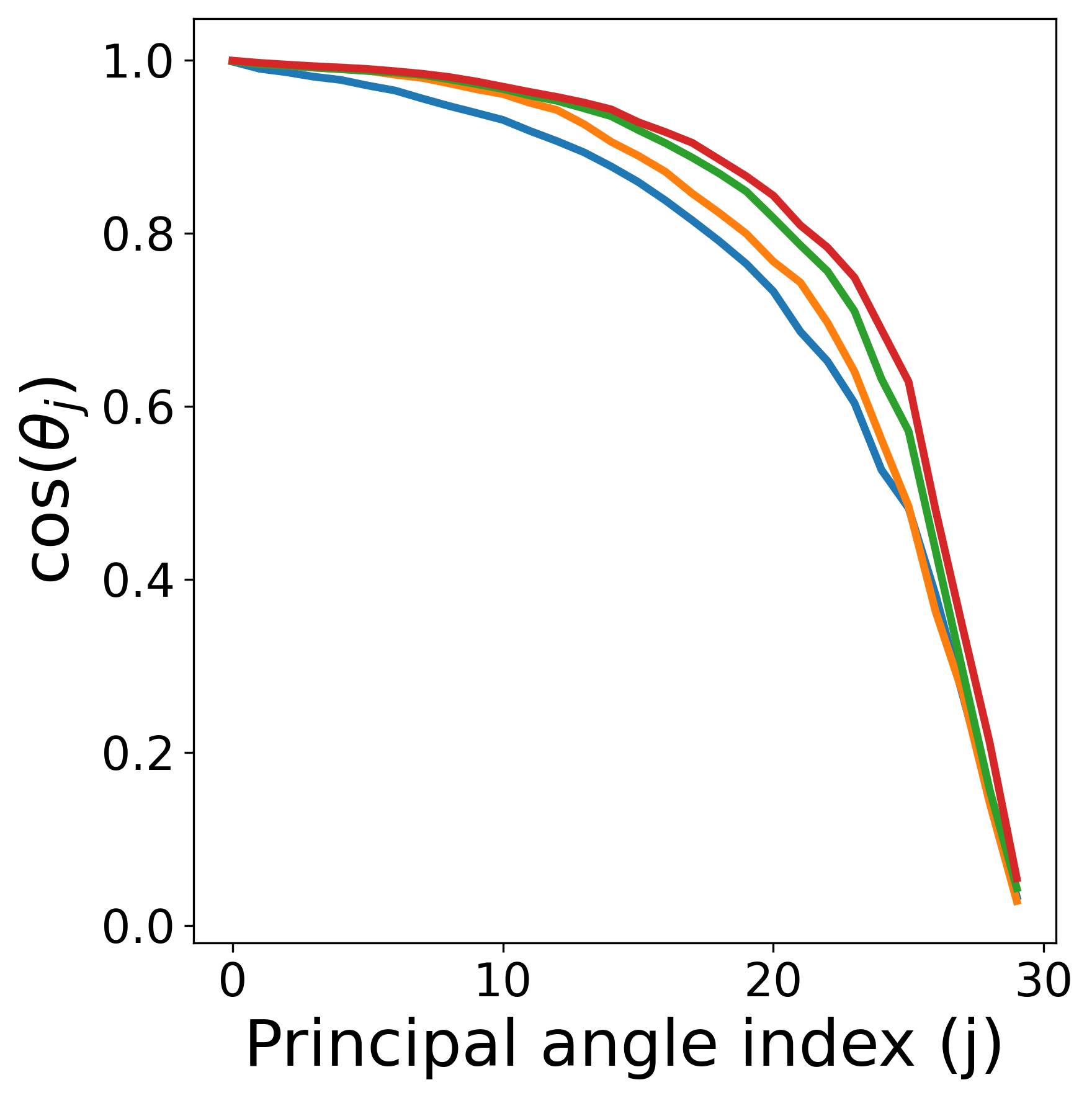}}
    \subfloat[$k=50$]{\includegraphics[width=0.24\linewidth]{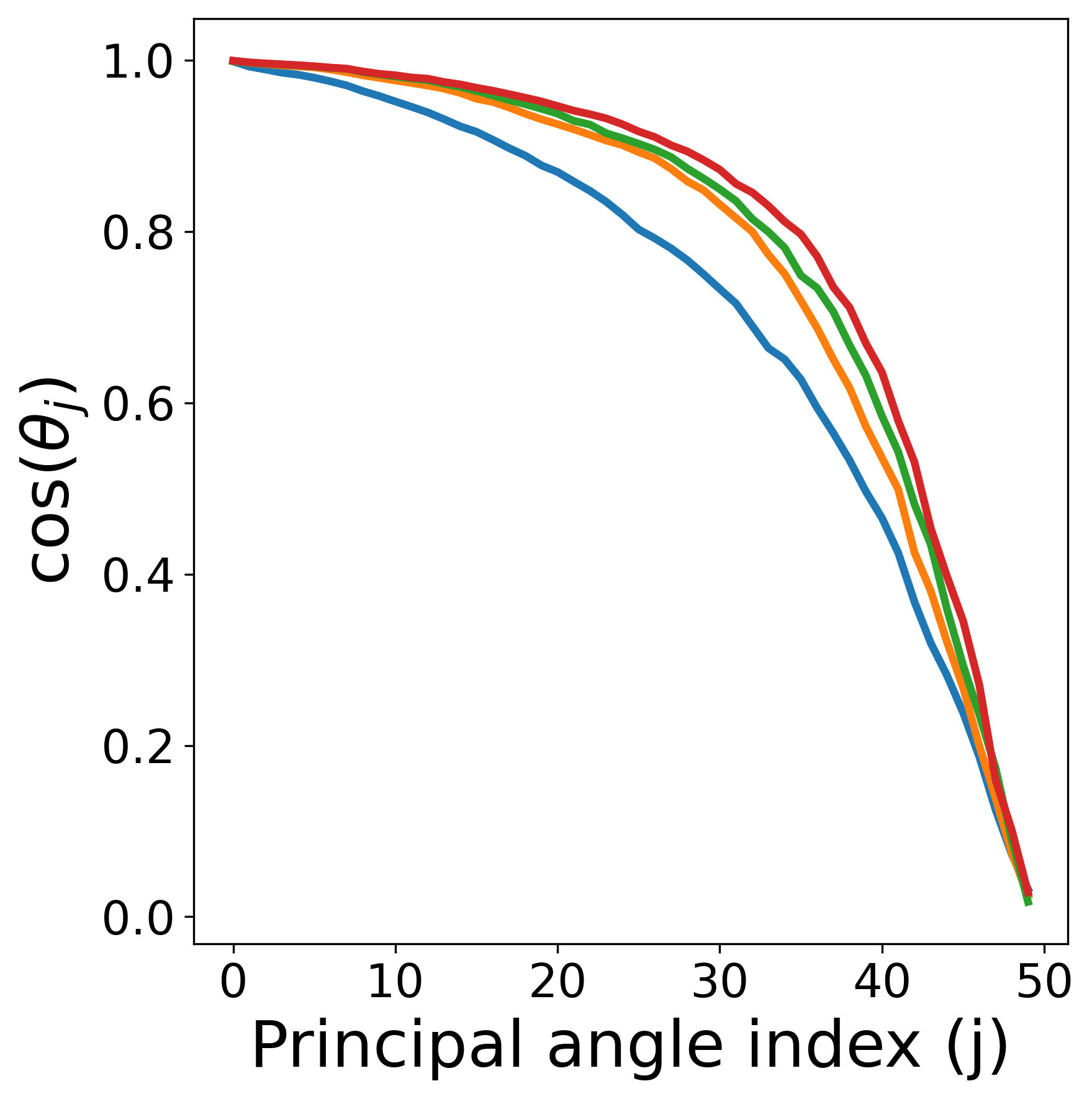}}    
    \subfloat[$k=100$]{\includegraphics[width=0.24\linewidth]{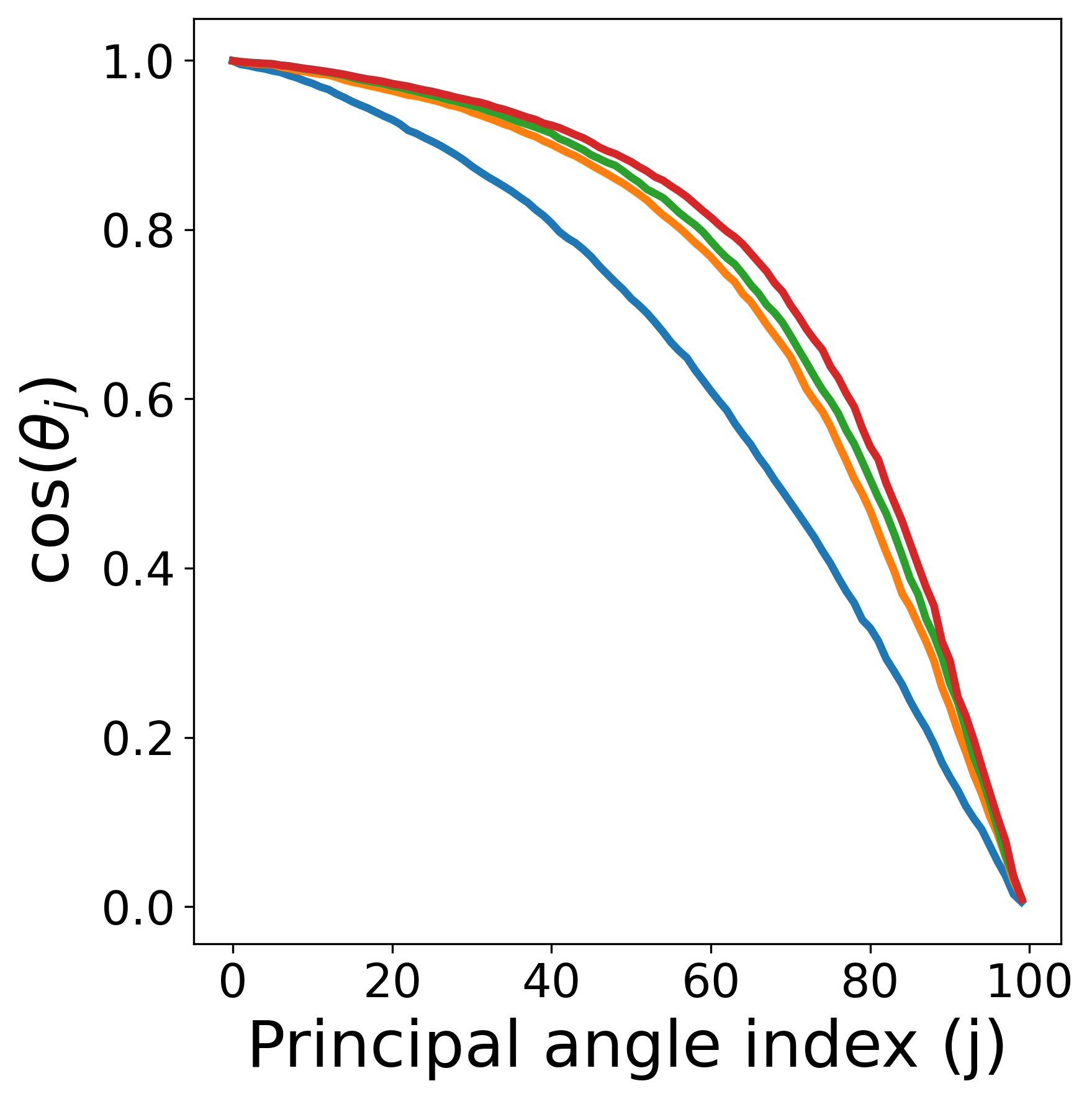}}
    \subfloat[$k=150$]{\includegraphics[width=0.24\linewidth]{CELEBA_results/principal_angles/principal_angles_k150_without_legend.png}}\\
    \subfloat{\includegraphics[width=0.2\linewidth]{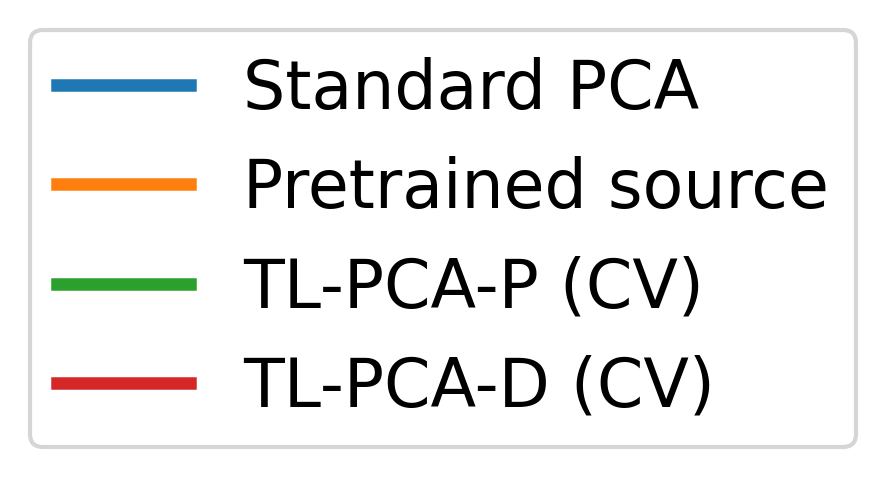}}
    \caption{Principal angles w.r.t.~ideal PCA (Tiny ImageNet to CelebA). }
    \label{appendix:fig:principal_angles_graphs_celeba}
\end{figure*}


\end{document}